\documentclass{article}

\usepackage[final]{neurips_2024}

\usepackage[utf8]{inputenc} %
\usepackage[T1]{fontenc}    %
\usepackage{hyperref}       %
\usepackage{url}            %
\usepackage{booktabs}       %
\usepackage{amsfonts}       %
\usepackage{nicefrac}       %
\usepackage{microtype}      %
\usepackage{graphicx}
\usepackage{neurips_2024}
\usepackage{natbib}
\usepackage{ulem}
\usepackage{amsmath}
\usepackage{multirow}
\usepackage{makecell}
\usepackage{wrapfig}
\usepackage{caption}
\usepackage{soul} 
\usepackage{color, xcolor} 
\definecolor{myc1}{rgb}{0.94, 0.97, 1.0}
\definecolor{myyellow}{HTML}{DBB428}
\definecolor{mypurple}{HTML}{682487}
\definecolor{mygreen}{HTML}{548235}
\definecolor{mygrey}{HTML}{7F7F7F}
\definecolor{mylightblue}{HTML}{84BA42}
\definecolor{mydarkblue}{HTML}{4485C7}

\newtheorem{lemma}{Lemma}
\newtheorem{definition}{Definition}
\newtheorem{prop}{Proposition}
\usepackage{algorithm}
\usepackage{algorithmic}
\usepackage[figuresright]{rotating}
\usepackage{amssymb} 
\title{Skill-aware Mutual Information Optimisation for Generalisation in Reinforcement Learning}

\author{%
Xuehui Yu$^{1,2}$ \quad Mhairi Dunion$^2$ \quad Xin Li$^1$ \quad Stefano V. Albrecht$^2$ \\
$^1$Harbin Institute of Technology  \quad $^2$University of Edinburgh\\
\texttt{\{yuxuehui,22s103169\}@stu.hit.edu.cn}\\
\texttt{\{mhairi.dunion,s.albrecht\}@ed.ac.uk}
}

\begin{document}
\maketitle
\begin{abstract} 
    Meta-Reinforcement Learning (Meta-RL) agents can struggle to operate across tasks with varying environmental features that require different optimal \textit{skills} (i.e., different modes of behaviour). Using context encoders based on contrastive learning to enhance the generalisability of Meta-RL agents is now widely studied but faces challenges such as the requirement for a large sample size, also referred to as the $\log$-$K$ curse. To improve RL generalisation to different tasks, we first introduce \textbf{S}kill-\textbf{a}ware \textbf{M}utual \textbf{I}nformation (SaMI), an optimisation objective that aids in distinguishing context embeddings according to skills, thereby equipping RL agents with the ability to identify and execute different skills across tasks. We then propose \textbf{S}kill-\textbf{a}ware \textbf{N}oise \textbf{C}ontrastive \textbf{E}stimation (SaNCE), a $K$-sample estimator used to optimise the SaMI objective. We provide a framework for equipping an RL agent with SaNCE in practice and conduct experimental validation on modified MuJoCo and Panda-gym benchmarks. We empirically find that RL agents that learn by maximising SaMI achieve substantially improved zero-shot generalisation to unseen tasks. Additionally, the context encoder trained with SaNCE demonstrates greater robustness to a reduction in the number of available samples, thus possessing the potential to overcome the $\log$-$K$ curse.
\end{abstract}

\section{Introduction}
\begin{minipage}[b]{0.33\linewidth}
Reinforcement Learning (RL) agents often learn policies that do not generalise across tasks in which the environmental features and optimal \textit{skills} are different \citep{tachet2018learning,garcin2024dred}. Consider a set of cube-moving tasks where an agent is required to move a cube to a goal position on a table (Figure \ref{fig:example}). These tasks become challenging if environmental features, such as table friction, vary between tasks. When facing an unknown environment, the agent
\end{minipage}
\hfill
\begin{minipage}[b]{0.65\linewidth}
\captionsetup{type=figure}
\includegraphics[width=1.0\linewidth]{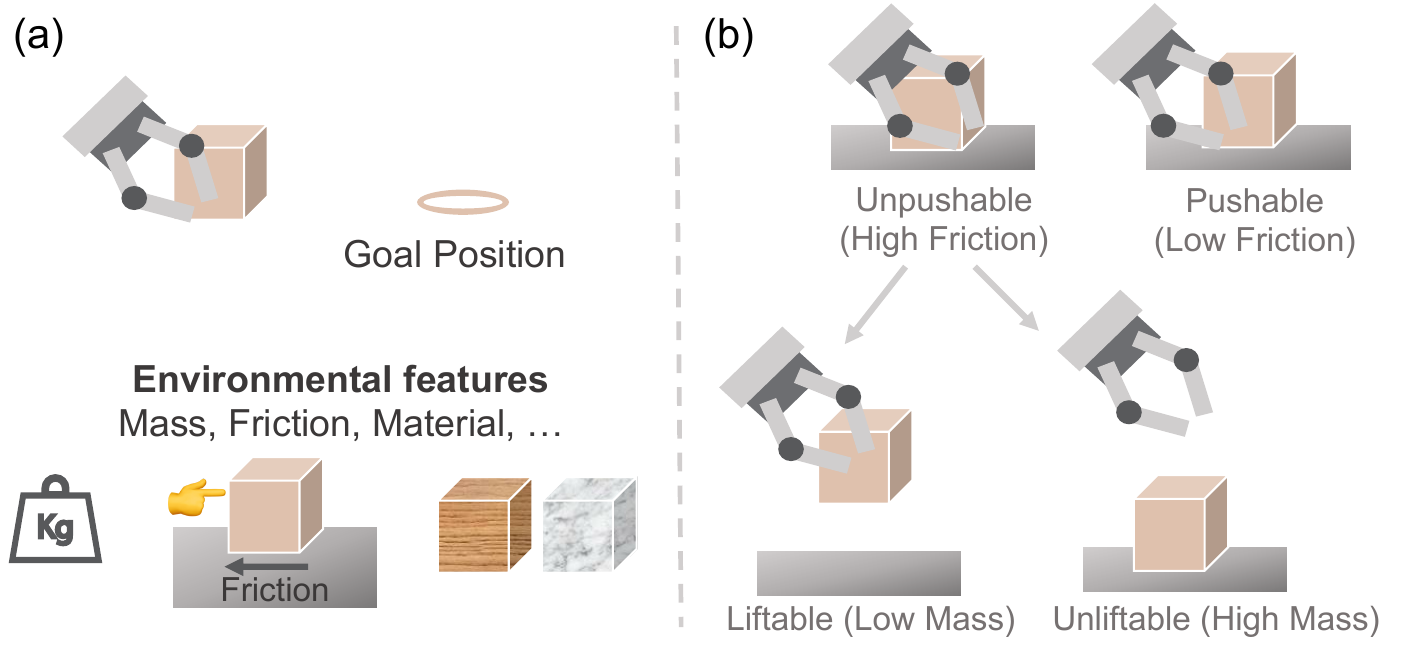}
\captionof{figure}{(a) In a cube-moving environment, tasks are defined according to different environmental features. (b) Different tasks have different transition dynamics caused by underlying environmental features, hence optimal skills are different across tasks.}
\label{fig:example}
\end{minipage}
needs to explore effectively, understand the environment, and adjust its behaviour accordingly within an episode. For instance, if the agent tries to push a cube across a table covered by a tablecloth and finds it “unpushable,” it should infer that the table friction is relatively high and adapt by lifting the cube to avoid friction, rather than continuing to push. Recent advances in Meta-Reinforcement Learning (Meta-RL) \citep{lee2020context,agarwal2021contrastive,mu2022decomposed,dunion2023temporal,dunion2023conditional,mcinroe2023hksl} enable agents to understand environmental features by inferring context embeddings from a small amount of exploration, and to train a policy conditioned on the context embedding to generalise to novel tasks.

In recent years, unsupervised contrastive learning algorithms have been shown to learn context embeddings that perform remarkably well on generalisation tasks \citep{nagabandi2018learning,lee2020context}. In particular, some Meta-RL algorithms \citep{fu2021towards,wang2021improving,li2021provably,sang2022pandr} train context encoders by maximising InfoNCE \citep{oord2019representation}, which has yielded vital insights into contrastive learning through the lens of mutual information (MI) analysis. The InfoNCE objective can be interpreted as a $K$-sample lower bound on the MI between trajectories and context embeddings. Despite significant advances, integrating contrastive learning with Meta-RL poses several unresolved challenges, of which two are particularly relevant to this research:
\textbf{(i) Existing context encoders based on contrastive learning do not distinguish tasks that require different skills}. Many prior algorithms only pull embeddings of the same tasks together and push those of different tasks apart. However, for example, a series of cube-moving tasks with high friction may only require a Pick\&Place skill (picking the cube off the table and placing it at the goal position), making further differentiation unnecessary. 
\textbf{(ii) Existing $\textit{\text{K}}$-sample MI estimators are sensitive to the sample size $\textit{\text{K}}$ (i.e., the $\text{log-\textit{K}}$ curse)} \citep{poole2019variational}. The low sample efficiency of RL \citep{franke2020sample} and the sample limitations in zero-shot generalisation make collecting a substantial quantity of samples often impractical \citep{arora2019theoretical,nozawa2021understanding}. The effectiveness of $K$-sample MI estimators breaks down with a finite sample size and leads to a significant performance drop in downstream RL tasks \citep{mnih2012fast,guo2022tight}.

To enhance RL generalisation across different tasks, we propose that the context embeddings should optimise downstream tasks and indicate whether the current skill remains optimal or requires further exploration. This also reduces the necessary sample size by focusing solely on skill-related information. In this work,
(1) we introduce \textbf{\textit{Skill-aware Mutual Information (SaMI)}}, a generalised form of MI objective between context embeddings, skills, and trajectories, designed to address issue (i). We provide a theoretical proof showing that by introducing skills as a third variable into the MI of context embeddings and trajectories, the resulting SaMI is smaller and easier to optimise.
Furthermore, (2) we propose a data-efficient $K$-sample estimator, \textbf{\textit{Skill-aware Noise Contrastive Estimation (SaNCE)}} to optimise SaMI, effectively addressing issue (ii).
Additionally, (3) we propose a practical skill-aware trajectory sampling strategy that shows how to sample positive and negative examples without relying on any prior skill distribution. In that way, Meta-RL agents autonomously acquire a set of skills applicable to many tasks, with these skills emerging solely from the SaMI learning objective and data.

We demonstrate empirically in MuJoCo \citep{todorov2012mujoco} and Panda-gym \citep{gallouedec2021pandagym} that SaMI enhances the zero-shot generalisation capabilities of two Meta-RL algorithms \citep{yu2020meta,fu2021towards} by achieving higher returns and success rates on previously unseen tasks, ranging from moderate to extreme difficulty. Visualisation of the learned context embeddings reveals distinct clusters corresponding to different skills, suggesting that the SaMI learning objective enables the context encoder to capture skill-related information from trajectories and incentivise Meta-RL agents to acquire a diverse set of skills. Moreover, SaNCE enables Meta-RL algorithms to use smaller sample spaces while achieving improved downstream control performance, indicating their potential to overcome the $\log$-$K$ curse.

\begin{figure}
\centering
\includegraphics[width=0.85\linewidth]{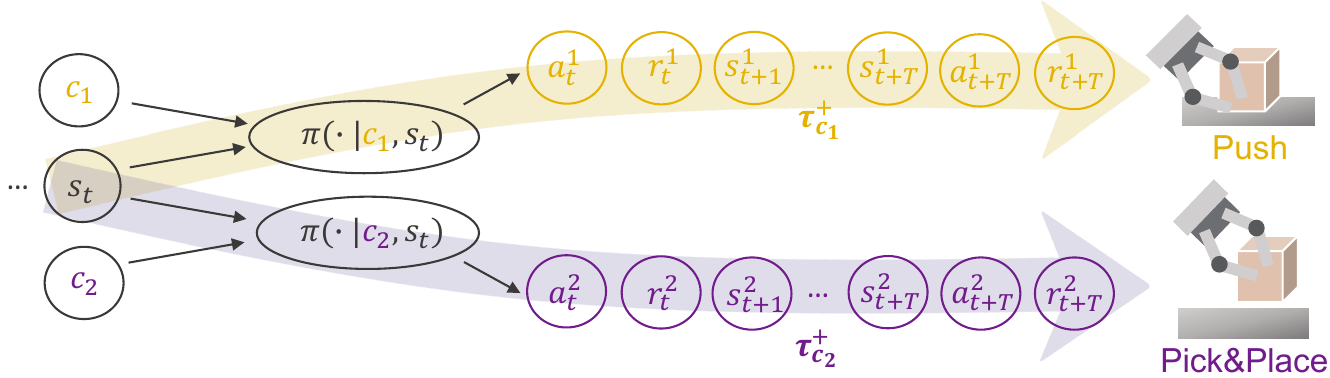}
\captionof{figure}{
A policy $\pi$ conditioned on a fixed context embedding $c$ is defined as a skill $\pi(\cdot|c)$ (shortened as $\pi_c$). The policy $\pi$ conditioned on a fixed $c$ alters the state of the environment in a consistent way, thereby exhibiting a mode of skill. The \textcolor{myyellow}{skill $\pi(\cdot|c_1)$} moves the cube on the table in trajectory \textcolor{myyellow}{$\tau^+_{c_1}$} and is referred to as the \textcolor{myyellow}{Push skill}; correspondingly, the \textcolor{mypurple}{Pick\&Place skill $\pi(\cdot|c_2)$} takes the cube off the table and places it in the goal position in the trajectory \textcolor{mypurple}{$\tau^+_{c_2}$}.}
\label{fig:mdp2}
\end{figure}

\section{Related works}
\textbf{Meta-RL.} By conditioning on an effective context embedding, Meta-RL policies can zero-shot generalise to new tasks with a small amount of exploration \cite{kirk2023survey}. Existing algorithms can be categorised into three types based on different context embeddings. In the first category, the context embedding is learned by minimising the downstream RL loss \citep{rakelly2019efficient,yu2020meta}. PEARL \citep{rakelly2019efficient} learns probabilistic context embeddings by recovering the value function. Multi-task SAC with task embeddings as inputs to policies (TESAC) \citep{hausman2018learning,yu2020meta} parameterises the learned policies through a shared embedding space, aiming to maximise the average returns across tasks. However, the update signals from the RL loss are stochastic and weak, and may not capture the similarity relations among tasks \citep{fu2021towards}. The second category involves learning context embeddings through dynamics prediction \citep{lee2020context, zhou2019environment}, which can make the context embeddings noisy, as they may model irrelevant dependencies and overlook task-specific information \citep{fu2021towards}. The third category employs contrastive learning \citep{fu2021towards, wang2021improving, li2021provably, sang2022pandr}, achieving significant improvements in context learning. However, these methods overlook the similarity of skills between different tasks, thus failing to achieve effective zero-shot generalisation by executing different skills. Our improvements build upon this third category by distinguishing context embeddings according to different optimal skills.

\textbf{Contrastive learning.} Contrastive learning has been applied to RL due to its significant momentum in representation learning in recent years, attributed to its superior effectiveness \citep{tishby2015deep,hjelm2019learning, dunion2024mvd}, ease of implementation \citep{oord2019representation}, and strong theoretical connection to MI estimation \citep{poole2019variational}. MI is often estimated using InfoNCE \citep{oord2019representation} that has gained recent attention due to its lower variance \citep{song2019understanding} and superior performance in downstream tasks. However, InfoNCE may underestimate the true MI when the sample size $K$ is finite. To address this limitation, CCM \citep{fu2021towards} uses a large number of samples for maximising InfoNCE. DOMINO \citep{mu2022decomposed} reduces the true MI by introducing an independence assumption; however, this results in biased estimates. We focus on proposing an unbiased alternative MI objective and a more data-efficient $K$-sample estimator tailored for downstream RL tasks, which, to our knowledge, have not been addressed in previous research.

\section{Preliminaries}
\label{pre}

\textbf{Reinforcement learning.} In Meta-RL, we assume an \textit{environment} is a distribution $\xi(e)$ of \textit{tasks} $e$ (e.g. uniform in our experiments). Each task $e \sim \xi(e)$ has a similar structure that corresponds to a Markov Decision Process (MDP) \citep{puterman2014markov}, defined by $\mathcal{M}_e=(\mathcal{S},\mathcal{A}, R, P_e,\gamma)$, with a state space $\mathcal{S}$, an action space $\mathcal{A}$, a reward function $R(s_t,a_t)$ where $s_t \in \mathcal{S}$ and $a_t \in \mathcal{A}$, state transition dynamics $P_e(s_{t+1}|s_t,a_t)$, and a discount factor $\gamma \in[0,1)$. In order to address the problem of zero-shot generalisation, we consider the transition dynamics $P_{e}(s_{t+1}|s_t,a_t)$ vary across tasks $e \sim \xi(e)$ according to multiple \textit{environmental features} $e=\{e^0,e^1,...,e^N\}$ that are not included in states $s$ and can be continuous random variables, such as mass and friction, or discrete random variables, such as the cube's material. For instance, in a cube-moving environment (Figure \ref{fig:example}), an agent has different tasks that are defined by different environmental features (e.g., mass and friction). The Meta-RL agent's goal is to learn a generalisable policy $\pi$ that is robust to such dynamic changes. Specifically, given a set of training tasks $e$ sampled from $\xi_{\text{train}}(e)$, we aim to learn a policy that can maximise the discounted returns, $\arg\max_\pi \mathbb{E}_{e \sim \xi_{\text{train}}(e)} [\sum_{t=0}^{\infty}\gamma^t R(s_t,a_t)|a_t \sim \pi(a_t|s_t), s_{t+1}\sim P_e(s_{t+1}|s_t,a_t)]$, and can produce accurate control for unseen test tasks sampled from $\xi_\text{test}(e)$. 

\textbf{Contrastive learning.} In Meta-RL, the context encoder $\psi(c|\tau_{c,0:t})$ first takes the trajectory $\tau_{c,0:t}= \{s_{0}, a_{0}, r_{0},...,s_{t}\}$ from the current episode as input and compresses it into a context embedding $c$ \citep{fu2021towards}. Then, the policy $\pi$, conditioned on context embedding $c$, consumes the current state $s_t$ and outputs the action $a_t$. As a key component, the
quality of context embedding $c$ can affect algorithms' performance significantly. MI is an effective measure of embedding quality \citep{goldfeld2019estimatinginformationflowdeep}, hence we focus on a context encoder that optimises the InfoNCE objective $I_{\text{InfoNCE}}(x;y)$, which is a $K$-sample estimator and lower bound of the MI $I(x;y)$ \citep{oord2019representation}. Given a query $x$ and a set $Y=\{y_1, ..., y_K\}$ of $K$ random samples containing one positive sample $y_1$ and $K-1$ negative samples from the distribution $p(y)$, $I_{\text{InfoNCE}}(x;y)$ is obtained by comparing pairs sampled from the joint distribution $x, y_1 \sim p(x, y)$ to pairs $x, y_k$ built using a set of negative examples $y_{2:K}$:
\begin{equation}
    I_{\text{InfoNCE}}(x;y|\psi,K)=\mathbb{E} \left[ \log \frac{f_{\psi}(x,y_1)}{\frac{1}{K} \sum_{k=1}^K f_{\psi}(x,y_k)} \right].
    \label{eq_4_1}
\end{equation}
InfoNCE constructs a formal lower bound on the MI, i.e., $I_{\text{InfoNCE}}(x;y|\psi, K) \leq I(x;y)$ \citep{guo2022tight,chen2021simpler}. Given two inputs $x$ and $y$, their \textit{embedding similarity} is $f_{\psi}(x,y) = e^{\psi(x)^\top\cdot \psi(y)/ \beta}$, where $\psi$ is the context encoder that projects $x$ and $y$ into the context embedding space, the dot product is used to calculate the similarity score between $\psi(x),\psi(y)$ pairs \citep{wu2018unsupervised,he2020momentum}, and $\beta$ is a temperature hyperparameter that controls the sensitivity of the product. Some previous Meta-RL methods \citep{lee2020context,mu2022decomposed} learn a context embedding $c$ by maximising $I_{\text{InfoNCE}}(c;\tau_c|\psi,K)$ between the context $c$ embedded from a trajectory in the current task, and the historical trajectories $\tau_c$ under the same environmental features setting.

\section{Skill-aware mutual information optimisation for Meta-RL}
\label{method}
\subsection{The $\boldsymbol{\log}$-$\boldsymbol{K}$ curse of $\boldsymbol{K}$-sample MI estimators} \label{sec:log_k}
In this section, we provide a theoretical analysis of the challenge inherent in learning a $K$-sample estimator for MI, commonly referred to as the $\log$-$K$ curse. Based on this theoretical analysis, we give insights to overcome this challenge. Given that we focus on the generalisation of RL, we only consider cases with a finite sample size of $K$. If a context encoder $\psi$ in Equation \eqref{eq_4_1} has sufficient training epochs, then $I_{\text{InfoNCE}}(x;y|\psi,K) \approx \log K$ \citep{mnih2012fast,guo2022tight}. Hence, the MI we can optimise is bottlenecked by the number of available samples, formally expressed as: 
\begin{minipage}[b]{0.52\linewidth}
\begin{lemma}
Learning a context encoder $\psi$ with a $K$-sample estimator and finite sample size $K$, we have \textcolor{mylightblue}{${I}_{\text{InfoNCE}}(x;y|\psi,K)$} $\leq$ \textcolor{mygrey}{$\log K$}$\leq$\textcolor{mydarkblue}{$I(x;y)$}, when $x \not \! \perp \!\!\! \perp y$ (see proof in Appendix \ref{appx:lemma_1}).
\label{lemma:lemma_1}
\end{lemma}

We do not consider the case when $x \perp \!\!\! \perp y$, i.e., $\log K \ge I(x;y) = 0$ ($\forall K \geq 1$), because a Meta-RL agent learns a context encoder by maximising MI between trajectories $\tau_c$ and context embeddings $c$, which are not independent (as shown in Figure \ref{fig:mdp2}). While an unbounded sample size $K$ for learning effective context embeddings is theoretically feasible and assumed in many studies \cite{mu2022decomposed,lee2020context}, it is often impractical in practice. Therefore, building on Lemma \ref{lemma:lemma_1}, we derive two key insights with limited samples: (1) generalising the current MI objective to be smaller than $I(x;y)$ (see Section \ref{method_4_1}); (2) developing a $K$-sample estimator tighter than $I_{\text{InfoNCE}}$ (see Section \ref{method_4_2}).
\end{minipage}
\hfill
\begin{minipage}[b]{0.46\linewidth}
\captionsetup{type=figure}
\includegraphics[width=\linewidth]{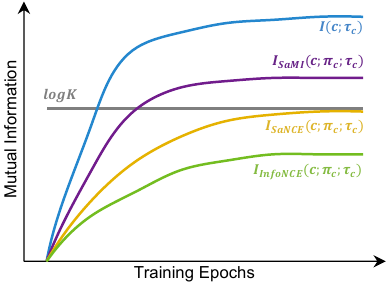}
\captionof{figure}{\textcolor{mylightblue}{$I_{\text{InfoNCE}(c;\pi_c;\tau_c)}$}, with a finite sample size of $K$, is a loose lower bound of \textcolor{mydarkblue}{$I(c;\tau_c)$} and leads to lower performance embeddings. \textcolor{mypurple}{$I_{\text{SaMI}}(c;\pi_c;\tau_c)$} is a lower ground-truth MI, and \textcolor{myyellow}{$I_{\text{SaNCE}}(c;\pi_c;\tau_c)$} is a tighter lower bound. }
\label{fig:sample_size_mi}
\end{minipage}

\subsection{Skill-aware mutual information: a smaller ground-truth MI}
\label{method_4_1}

We aim for our MI learning objective to incentivise agents to acquire a diverse set of skills, enabling them to generalise effectively across tasks. To start with, we define skills \citep{eysenbach2018diversityneedlearningskills}: 
\begin{definition}[Skills]
A policy $\pi$ conditioned on a fixed context embedding $c$ is defined as a skill $\pi(\cdot|c)$, abbreviated as $\pi_c$. If a skill $\pi_c$ is conditioned on a state $s_t$, we can sample actions $a_t \sim \pi(\cdot|c,s_t)$. After sampling actions from $\pi_c$ at consecutive timesteps, we obtain a trajectory $\tau_{c,t:t_T}=\{s_t, a_t, r_t, s_{t+1}, \ldots, s_{t+T}, a_{t+T}, r_{t+T}\}$ which demonstrates a consistent mode of behaviour.
\label{def:definition_skill}
\end{definition}
After a limited amount of exploration, an agent should be able to infer the task (i.e., environmental features $e={e^0, e^1, \ldots, e^N}$) and adapt accordingly within the current episode. The context embedding should encompass skill-related information, guiding the policy on when to explore new skills or switch between existing ones. We propose that the context encoder $\psi$ should be trained by maximising the MI between the context embedding $c$, skills $\pi_c$, and trajectories $\tau_c$. To this end, we propose a novel MI optimisation objective, \textbf{Skill-aware Mutual Information (SaMI)}, defined as:
\begin{equation}
    I_{\text{SaMI}}(c;\pi_c;\tau_c) = I(c;\tau_c) - I(c;\tau_c|\pi_c).
    \label{eq_4_1_1}
\end{equation}
SaMI is defined according to interaction information \citep{mcgill1954multivariate}, serving as a generalisation of MI for three variables $\{c,\pi_c,\tau_c\}$. Although we cannot evaluate $p(c,\pi_c,\tau_c)$ directly, we approximate it by Monte-Carlo sampling, using $K$ samples from $p(c,\pi_c,\tau_c)$. As illustrated in Figure \ref{fig:sample_size_mi}, a context encoder $\psi$ trained with the objective of maximising $I_{\text{SaMI}}(c;\pi_c;\tau_c)$ converges more quickly, as $I_{\text{SaMI}}(c;\pi_c;\tau_c) \leq I(c;\tau_c)$ (see proof in Appendix \ref{appx:lemma_2}). By focusing more on skill-related information, $I_{SaMI}$ enables agents to autonomously discover a diverse range of skills for handling multiple tasks.

\begin{figure}
\centering
\includegraphics[width=1.0\linewidth]{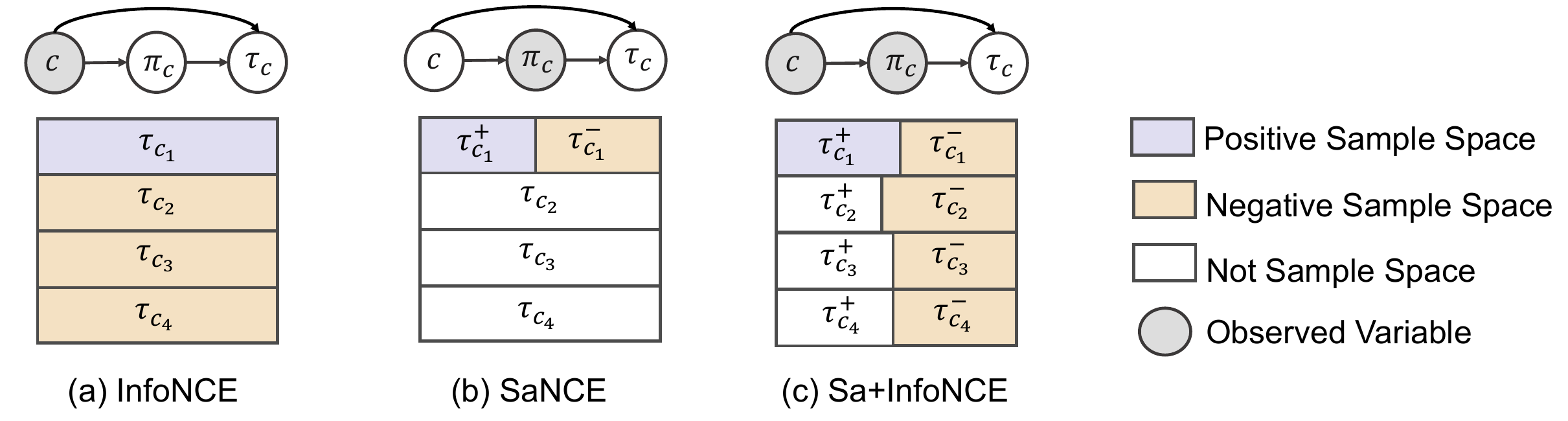}
\captionof{figure}{A comparison of sample spaces for task $e_1$. Positive samples $\tau_{c_1}$ or $\tau_{c_1}^+$ are always from current task $e_1$. For SaNCE, in a task $e_k$ with embedding $c_k$, the positive skill $\pi_{c_k}^+$ conditions on $c_k$ and generates positive trajectories $\tau_{c_k}^+$, and the negative skill $\pi_{c_k}^-$ generates negative trajectories $\tau_{c_k}^-$. The top graphs show the relationship between $c$, $\pi_c$ and $\tau_c$.}
\label{fig:fig_sample_space}
\end{figure}

\subsection{Skill-aware noise contrastive estimation: a tighter $\boldsymbol{K}$-sample estimator}
\label{method_4_2}

Despite InfoNCE's success as a $K$-sample estimator for approximating MI \citep{laskin2020curl,eysenbach2022contrastive}, its learning efficiency plunges due to limited numerical precision, which is called the $\log$-$K$ curse, i.e., $I_{\text{InfoNCE}} \leq \log K \leq I_{\text{SaMI}}$ \citep{chen2021simpler} (see proof in Appendix \ref{appx:lemma_2}). When $K \to +\infty$, we can expect $I_{\text{InfoNCE}} \approx \log K \approx I_{\text{SaMI}}$ \citep{guo2022tight}. However, increasing $K$ is too expensive, especially in complex environments with enormous negative sample space. To address this, we propose a novel $K$-sample estimator that requires a significantly smaller sample size $K \ll +\infty$. First, we define $K^*$:

\begin{definition}[$\boldsymbol{K^*}$] $K^* = |c| \cdot |\pi_c| \cdot M$ is defined as the number of trajectories in the replay buffer (i.e., the sample space), in which $|c|$ represents the number of different context embeddings $c$, $|\pi_c|$ represents the number of different skills $\pi_c$, and $M$ is a natural number. \label{def:definition_number}
\end{definition}

\begin{figure}[ht]
\centering
\includegraphics[width=1.0\linewidth]{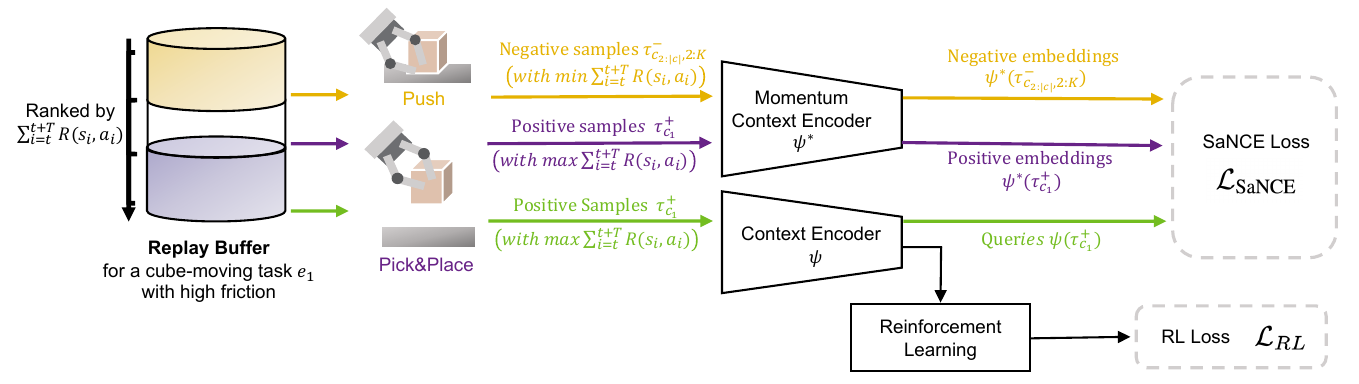}
\caption{A practical framework for using SaNCE in the meta-training phase. During meta-training, we sample trajectories from the replay buffer for off-policy training. \textcolor{mygreen}{Queries} are generated by a context encoder $\psi$, which is updated with gradients from both the SaNCE loss $\mathcal{L}_{\text{SaNCE}}$ and the RL loss $\mathcal{L}_{RL}$. \textcolor{myyellow}{Negative}/\textcolor{mypurple}{Positive} embeddings are encoded by a momentum context encoder $\psi^*$, which is driven by a momentum update with the encoder $\psi$. During meta-testing, the meta-trained context encoder $\psi$ embeds the current trajectory, and the RL policy takes the embedding as input together with the state for adaptation within an episode.}
\label{fig:framework_algorithm_2}
\end{figure}

To ensure that $I_{\text{InfoNCE}}$ is a tight bound of $I_{\text{SaMI}}$, we require that $I_{\text{InfoNCE}} \approx \log K \approx I_{\text{SaMI}}$ when $K \to K^*$. 
Under the definition of $K^*$, the replay buffer can be divided according to the different context embeddings $c$ and skills $\pi_c$ (i.e., observing context embeddings $c$ and skills $\pi_c$). In real-world robotic control tasks, the sample space size significantly increases due to multiple environmental features $e=\{e^0,e^1,...,e^N\}$. Taking the sample space of InfoNCE as an example (Figure \ref{fig:fig_sample_space}(a)), in the current task $e_1$ with context embedding $c_1$, positive samples are trajectories $\tau_{c_1}$ generated after executing the skill $\pi_{c_1}$ in task $e_1$, and negative samples are trajectories $\{\tau_{c_2},...\}$ from other tasks $\{e_2,...\}$. The permutations and combinations of $N$ environmental features lead to an exponential growth in task number $|c|$, which in turn results in an increase of sample space $K^*_{\text{InfoNCE}}=|c|\cdot|\pi|\cdot M$. 

We introduce a tight $K$-sample estimator, \textbf{Skill-aware Noise Contrastive Estimation (SaNCE)}, which is used to approximate $I_{\text{SaMI}}(c;\pi_c;\tau_c)$ with $K^*_{\text{SaNCE}} < K^*_{\text{InfoNCE}}$. For SaNCE, both positive samples $\tau_{c_1}^+$ and negative samples $\tau_{c_1}^-$ are sampled from the current tasks $e_1$, but are generated by executing positive skills $\pi_{c_1}^+$ and negative skills $\pi_{c_1}^-$, respectively. Here, a \textit{positive skill} is intuitively defined by whether it is optimal for the current task $e$, with a more formal definition provided in Section \ref{method_4_3}. For instance, in a cube-moving task under a large friction setting, the agent executes a skill $\pi_{c}^+$ after several iterations of learning, and obtains corresponding trajectories $\tau_{c}^+$ where the cubes leave the table surface. This indicates that the skill $\pi_{c}^+$ is Pick\&Place and other skills $\pi_{c}^-$ may include Push or Flip (flipping the cube to the goal position), with corresponding trajectories $\tau_{c}^-$ where the cube remains stationary or rolls on the table. Formally, we can optimise the $K$-sample lower bound $I_{\text{SaNCE}}$ to approximate $I_{\text{SaMI}}$:
\begin{align}
    & \quad I_{\text{SaNCE}}(c;\pi_c;\tau_c|\psi,K) \nonumber \\
    & = \mathbb{E}_{p(c_1,\pi_{c_1},\tau_{c_1}^+)p(\tau_{c_1,2:K}^-)} \left[ \log \left( \frac{K \cdot f_{\psi}(c_1,\pi_{c_1},\tau_{c_1}^+)}{f_{\psi}(c_1,\pi_{c_1},\tau_{c_1}^+) + \sum_{k=2}^{K} f_{\psi}(c_1,\pi_{c_1},\tau_{c_1,k}^-)} \right) \right] \nonumber  \\
    & \leq I_{\text{SaMI}}(c;\pi_c;\tau_c)
    \label{eq_4_2_2}
\end{align}
where $f_{\psi}(c_1,\pi_{c_1},\tau_{c_1}) = e^{\psi(\tau_{c_1})^\top\cdot \psi^*(\tau_{c_1})/ \beta}$. The \textcolor{mygreen}{query $c_1=\psi(\tau_{c_1})$} is generated by the context encoder $\psi$. For training stability, we use a momentum encoder $\psi^*$ to produce the \textcolor{mypurple}{positive} and \textcolor{myyellow}{negative} embeddings. SaNCE significantly reduces the required sample space size $K^*_{\text{SaNCE}}$ by sampling trajectories $\tau_c$ based on different skills $\pi_{c}$ (Figure \ref{fig:fig_sample_space}(b)) in task $e_1$, so that $K^*_{\text{SaNCE}}=|c| \cdot |\pi_{c}|\cdot M = |\pi_{c_1}|\cdot M \leq K^*_{\text{InfoNCE}}$ ($|c|=|c_1|=1$). Therefore, $I_{\text{SaNCE}}$ satisfies Lemma \ref{lemma:lemma_2}:
\begin{lemma}
With a context encoder $\psi$ and finite sample size $K$, we have $I_{\text{InfoNCE}}(c;\pi_c;\tau_c|\psi,K) \leq I_{\text{SaNCE}}(c;\pi_c;\tau_c|\psi,K) \leq \log K \leq I_{\text{SaMI}}(c;\pi_c;\tau_c) \leq I(c;\tau_c)$. (see proof in Appendix \ref{appx:lemma_2})
\label{lemma:lemma_2}
\end{lemma}
SaNCE can be used alone or combined with other optimisation objectives to train context encoders in Meta-RL algorithms. For instance, integrating SaNCE with InfoNCE diversifies the negative sample space, with \(K^*_{\text{Sa+InfoNCE}}=\left(\sum_{i=1}^{|c|}|\pi_{c_i}^-| + |\pi_{c_1}^+|\right) \cdot M\). The sample space for \(I_{\text{Sa+InfoNCE}}\) is depicted in Figure \ref{fig:fig_sample_space}(c) and further analysed in detail in Appendix \ref{appx:proof_3}.

\subsection{Skill-aware trajectory sampling strategy}
\label{method_4_3}

In this section, we propose a practical trajectory sampling method. Methods focusing on skill diversity often rely heavily on accurately defining and identifying individual skills \citep{eysenbach2018diversityneedlearningskills}. Some of these methods require a prior skill distribution, which is often inaccessible \citep{shi2022skimo}, and it is impractical to enumerate all possible skills that we hope the model to learn. Besides, we believe that distinctiveness of skills is inherently difficult to achieve — a slight difference in states can make two skills distinguishable, and not necessarily in a semantically meaningful way. Consequently, we do not directly teach any of these skills or assume any prior skill distribution. Diverse skills naturally emerge from the incentives of the SaMI learning objective in a multi-task setting, driven by the inherent need to develop generalisable skills. For example, in high-friction tasks, the agent must acquire the Pick\&Place skill to avoid large frictional forces, whereas in high-mass tasks, the agent must learn the Push skill since it cannot lift the cube. In each task, we only identify whether the skills are optimal; in this way, under a multi-task setting, the agent will acquire a set of general skills that are applicable to many tasks.

In a given task \(e\), \textcolor{mypurple}{\textit{positive skills} \(\pi_{c}^+\)} are defined as optimal skills achieving highest return \(\sum_{i=t}^{t+T}R(s_i,a_i)\), whereas \textcolor{myyellow}{\textit{negative skills} \(\pi_{c}^-\)} are those that result in lower returns. As a result, the positive sample \(\tau_c^+\) consists of trajectories generated by the skills with the highest ranked returns, while the negative samples correspond to those with the lowest returns. This straightforward approach of selecting positive samples based on the ranked highest return effectively aligns with positive skills and mitigates the challenge of hard negative examples \citep{robinson2021contrastivelearninghardnegative}. The SaNCE loss is then minimised to bring the context embeddings of the highest return trajectories closer while distancing those of negative trajectories. By the end of training, the top-ranked trajectories in the ranked replay buffer correspond to positive samples \(\tau_c^+\) with high returns, while the lower-ranked trajectories represent negative samples \(\tau_c^-\) with low returns. However, at the beginning of training, it is likely that all trajectories have low returns. Therefore, our SaNCE loss is a soft variant of the \(K\)-sample SaNCE:
\begin{equation}
    \mathcal{L}_{\text{SaNCE}} = - \max \left(||\psi(\tau_c^+),\psi(\tau_c^-)||_{L2}, 1\right) \cdot I_{\text{SaNCE}}
\end{equation}
where $||\cdot||_{L2}$ represents the Euclidean distance \citep{tabak2014geometry}. Figure \ref{fig:framework_algorithm_2} provides a practical framework of SaNCE, with a cube-moving example task $e_1$ under high friction. In task $e_1$, the positive skill $\pi_{c_1}^+$ is the \textcolor{mypurple}{Pick\&Place} skill, which is used to generate \textcolor{mygreen}{queries $\psi(\tau_{c_1}^+)$} and \textcolor{mypurple}{positive embeddings $\psi^*(\tau_{c_1}^+)$}; after executing \textcolor{myyellow}{Push} skill we get \textcolor{myyellow}{negative samples $\tau_{c_1}^-$} and \textcolor{myyellow}{negative embeddings $\psi^*(\tau_{c_1}^-)$}. 

\section{Experiments}

\begin{figure}
    \centering
    \includegraphics[width=1.0\linewidth]{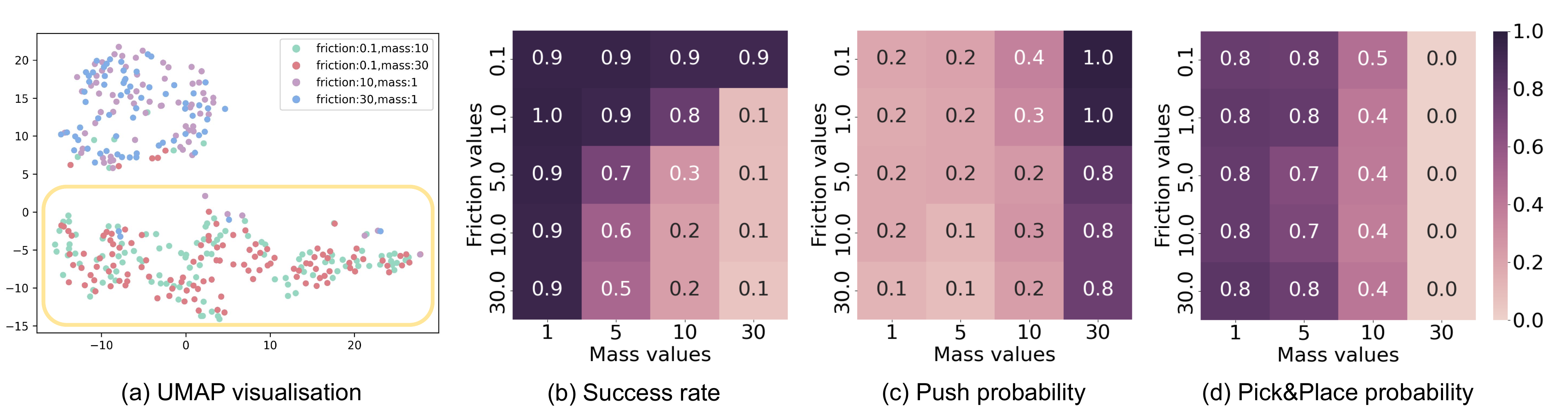}
    \caption{(a) UMAP visualisation of context embeddings for the SaCCM in the Panda-gym environment, with points in the yellow box representing the Push skill in high-mass tasks. Heatmap of (b) success rate, (c) Push skill probability, and (d) Pick\&Place skill probability for SaCCM. In large-mass scenarios, the Push skill is more likely to be executed than Pick\&Place.}
\label{fig:heatmap_skills}
\end{figure}

Our experiments aim to answer the following questions:
(1) Does optimising SaMI lead to increased returns during training and zero-shot generalisation (see Table \ref{tab:panda-gym-average-return} and \ref{tab:mujoco-average-return})?; (2) Does SaMI help the RL agents to be versatile and embody multiple skills (see Figure \ref{fig:heatmap_skills})?; (3) Can SaNCE overcome the $\log$-$K$ curse in sample-limited scenarios (see Table \ref{tab:panda-gym-average-return} and \ref{tab:mujoco-average-return}, and Section \ref{sec:discussion})?

\subsection{Experimental setup}

\textbf{Modified benchmarks with multiple environmental features.\footnote{Our modified benchmarks are open-sourced at https://github.com/uoe-agents/Skill-aware-Panda-gym \label{github_benchmark}}} We evaluate our method on two benchmarks, Panda-gym \citep{gallouedec2021pandagym} and MuJoCo \citep{todorov2012mujoco} (details in Sections \ref{section:panda_gym} and \ref{section:mujoco}). The benchmarks are modified to be influenced by multiple environmental features, which are sampled at the start of each episode during meta-training and meta-testing. During meta-training, we uniform-randomly select a combination of environmental features from a training task set. At test time, we evaluate each algorithm in unseen tasks with environmental features outside the training range. Generalisation performance is measured in two different regimes: moderate and extreme. The moderate regime draws environmental features from a closer range to the training range compared to the extreme. Our results report the mean and standard deviation of models trained over five seeds in both training and test tasks. Further details are available in Appendix \ref{appx:environment_setup}.

\textbf{Baselines.} The loss function of the context encoder in all algorithms consists of two key components: the RL loss and the contrastive loss. The RL loss $\mathcal{L}_{RL}$, which is the same across all methods, corresponds to the RL value function loss. Our primary comparison focuses on the contrastive loss, taking the form of either SaNCE, InfoNCE, or no contrastive loss. Accordingly, we select baselines based on contrastive loss: \textbf{CCM} \citep{fu2021towards}, which utilises InfoNCE, and \textbf{TESAC} \citep{yu2020meta}, which relies solely on the RL loss, allowing assessment of the context encoder without contrastive loss. Since CCM and TESAC use an RNN encoder, we also include \textbf{PEARL} \citep{rakelly2019efficient}, which utilises an MLP context encoder and follows the similar RL loss. Additionally, Appendix \ref{appx:sub-standard-comparison} includes comparisons with DOMINO \citep{mu2022decomposed} and CaDM \citep{lee2020context}, using the same environmental setup in the MuJoCo benchmark.

\textbf{Our method.\footnote{Our code, video demos and experimental data are available at https://github.com/uoe-agents/SaMI \label{github}}} We use Soft Actor-Critic (SAC) \citep{haarnoja2018soft} as the base RL algorithm, training agents for 1.6 million timesteps in each environment (details in Appendix \ref{appendix:Implementation_details}). SaNCE is a simple objective based on MI that can be used to train any context encoder. We integrate SaNCE into two Meta-RL algorithms: (1) \textbf{SaTESAC} is TESAC with SaNCE, which uses SaNCE for contrastive learning, using a $|c|$ times smaller sample space (Figure \ref{fig:fig_sample_space}(b)); (2) \textbf{SaCCM} is CCM with SaNCE, where the contrastive learning combines InfoNCE and SaNCE, as shown in Figure \ref{fig:fig_sample_space}(c). 

\begin{minipage}[b]{0.43\linewidth}
\subsection{Panda-gym} \label{section:panda_gym}
 \textbf{Task description.}  Our modified Panda-gym benchmark involves a robot arm control task using the Franka Emika Panda \citep{gallouedec2021pandagym}, where the robot moves a cube to a target position. Unlike previous works, we simultaneously modify multiple environmental features (cube mass and table friction) that characterise the transition dynamics, and the robot can flexibly execute different skills (Push and Pick\&Place) for different tasks. This environment requires high skill diversity; for instance, the agent must use Pick\&Place in high-friction tasks and Push in high-mass tasks.
\end{minipage}
\hfill
\begin{minipage}[b]{0.54\linewidth}
    \centering
    \fontsize{8}{11}\selectfont  
    \captionsetup{type=table}
    \captionof{table}{Comparison of success rate $\pm$ standard deviation with baselines in Panda-gym (over 5 seeds). \textbf{Bold text} signifies the highest average return. $*$ next to the number means that the algorithm with SaMI has statistically significant improvement over the same algorithm without SaMI. All significance claims based on paried t-tests with significance threshold of $p < 0.05$.}
    \centering
    \scalebox{0.98}{
    \begin{tabular}{cccc}
		\toprule
		\toprule
		  & Training & Test (moderate) & Test (extreme)  \cr
		\cmidrule(lr){1-4}
		PEARL & 0.42$\pm$0.19 & 0.10$\pm$0.06 & 0.11$\pm$0.05 \cr
		TESAC & 0.50$\pm$0.22 & 0.31$\pm$0.20 & 0.22$\pm$0.21\cr
		CCM & 0.80$\pm$0.19 & 0.49$\pm$0.23 &  0.29$\pm$0.28 \cr
            \cmidrule(lr){1-4}
            SaTESAC & 0.92$\pm$0.04$^*$ & 0.56$\pm$0.24$^*$ & \textbf{0.37$\pm$0.34}$^*$  \cr
		SaCCM & \textbf{0.93$\pm$0.05}$^*$ & \textbf{0.57$\pm$0.26}$^*$ &  0.36$\pm$0.35$^*$  \cr
		\bottomrule
		\bottomrule
	\end{tabular}}\vspace{0cm}
\label{tab:panda-gym-average-return}
\end{minipage}

\textbf{Results and skill analysis.} 
As shown in Table \ref{tab:panda-gym-average-return}, SaTESAC and SaCCM achieve superior generalisation performance compared to PEARL, TESAC, and CCM, with a smaller sample space. The t-test results in Table \ref{tab:panda-gym-average-return} show that SaMI significantly improves success rates across training, moderate, and extreme test sets at a 0.05 significance level. Video demos\textsuperscript{\ref{github}} show that agents equipped with SaMI acquired multiple skills (Push, Pick\&Place) to handle various tasks. When faced with an unknown task, the agents explore by attempting to lift the cube, infer the context, and adjust their skills accordingly within the episode. We visualised the context embeddings using UMAP \citep{mcinnes2020umapuniformmanifoldapproximation} (Figure \ref{fig:heatmap_skills}(a) and Appendix \ref{appx:visualisation}) and t-SNE \citep{van2008visualizing} (Appendix \ref{appx:visualisation}), plotting the final step from 100 tests per task. Skills were identified through contact points between the end effector, cube, and table (see Appendix \ref{appx:environment_setup} for more details), and heatmaps \citep{Waskom2021} were used to visualise executed skills. Figure \ref{fig:heatmap_skills} shows that SaCCM agents learned the Push skill for large cube masses (30 Kg, 10 Kg) and Pick\&Place for smaller masses, while CCM showed no clear skill grouping (Figure \ref{fig:vis_tnse_panda} in Appendix \ref{appx:visualisation_panda}). Overall, SaMI incentivises agents to autonomously learn diverse skills, enhancing generalisation across a wider range of tasks. Specifically, through the cycle of effective exploration, context inference, and adaptation, diverse skills emerge solely from the data. Further visualisation results are in Appendix \ref{appx:visualisation}.

\begin{table}
\centering
\fontsize{7.5}{8.4}\selectfont  
\caption{Comparison of average return $\pm$ standard deviation with baselines in modified MuJoCo benchmark (over 5 seeds). \textbf{Bold number} signifies the highest return. $*$ next to the number means that the algorithm with SaMI has statistically significant improvement over the same algorithm without SaMI. All significance claims based on t-tests with significance threshold of $p < 0.05$.}
	\begin{tabular}{ccccccc}
		\toprule
		\toprule
		\multirow{2}{*}{ }& \multicolumn{3}{c}{Crippled Ant} & \multicolumn{3}{c}{Crippled Half-cheetah}\cr
		\cmidrule(lr){2-4} \cmidrule(lr){5-7}
		  & Training & Test (moderate) & Test (extreme) & Training & Test (moderate) & Test (extreme)\cr
		\cmidrule(lr){1-7}
		PEARL & 1682$\pm$73 & 996$\pm$21 & 888$\pm$31 & 1998$\pm$973 & 698$\pm$548 &   746$\pm$1092 \cr
		TESAC & 2139$\pm$90 & 1952$\pm$40 &  1048$\pm$124  &  3967$\pm$955 & 874$\pm$901 &  846$\pm$849   \cr
		CCM & 2361$\pm$114 & 2047$\pm$83 &  1527$\pm$301  & 3481$\pm$488 & 821$\pm$575 &  873$\pm$914  \cr
            \cmidrule(lr){1-7}
            SaTESAC & \textbf{2638$\pm$406} & \textbf{2379$\pm$528} & \textbf{2131$\pm$132}$^*$   & 
            4328$\pm$1092 & \textbf{1143$\pm$664}$^*$ & \textbf{1540$\pm$1094}  \cr
		SaCCM & 2355$\pm$170 & 2310$\pm$314 & 2007$\pm$68$^*$ & 
  \textbf{4478$\pm$1131}$^*$ & 1007$\pm$568 &  1027$\pm$782 \cr
		\bottomrule
		\bottomrule
  
    	\multirow{2}{*}{ }& \multicolumn{3}{c}{Ant} & \multicolumn{3}{c}{Half-cheetah}  \cr
		\cmidrule(lr){2-4} \cmidrule(lr){5-7} 
		  & Training & Test (moderate) & Test (extreme) & Training & Test (moderate) & Test (extreme) \cr
		\cmidrule(lr){1-7}
		PEARL & 5153$\pm$581 & 3873$\pm$235 & 3802$\pm$409 & 5802$\pm$773 & 2190$\pm$970 &   1346$\pm$692 \cr
		TESAC & 6789$\pm$451 & 4705$\pm$279 & 4108$\pm$369 & 6298$\pm$2310 & 3173$\pm$1210 &  1159$\pm$338  \cr
		CCM & 6901$\pm$567 & 5179$\pm$902 &  4700$\pm$696  & 6955$\pm$788 & 3963$\pm$622 & 1325$\pm$269\cr
            \cmidrule(lr){1-7}
            SaTESAC & 7314$\pm$545 & 5513$\pm$648$^*$ & 4940$\pm$531$^*$  & 
            \textbf{7430$\pm$1026} & \textbf{4058$\pm$890} &  1780$\pm$102$^*$ \cr
		SaCCM & \textbf{7478$\pm$539} & \textbf{5717$\pm$488} & \textbf{5215$\pm$377} & 7154$\pm$965 & 3849$\pm$689  &  \textbf{1926$\pm$218}$^*$ \cr
  		\bottomrule
		\bottomrule

  \multirow{2}{*}{ }&  \multicolumn{3}{c}{SlimHumanoid} & \multicolumn{3}{c}{HumanoidStandup}  \cr
		\cmidrule(lr){2-4} \cmidrule(lr){5-7}
		  & Training & Test (moderate) & Test (extreme) & Training & Test (moderate) & Test (extreme)\cr
		\cmidrule(lr){1-7}
		PEARL & 6947$\pm$3541 & 3697$\pm$2674 & 2018$\pm$907 & 95456$\pm$13445 & 63242$\pm$13546 & 64224$\pm$15467\cr
		TESAC & 8437$\pm$1798 & 6989$\pm$1301 &   3760$\pm$308 & 158384$\pm$14455 & 153944$\pm$15046 &  74220$\pm$19980 \cr
		CCM & 7696$\pm$1907 & 5784$\pm$531 &  2887$\pm$1058  & 146480$\pm$33745 & 154601$\pm$16291 & 94991$\pm$15258  \cr
        \cmidrule(lr){1-7}
        SaTESAC & \textbf{10216$\pm$1620} & \textbf{7886$\pm$2203} &  6123$\pm$1403$^*$ &  178142$\pm$10081$^*$ & 168337$\pm$12123 &  133335$\pm$24607$^*$ \cr
		SaCCM & 9312$\pm$705 & 7430$\pm$1587 & \textbf{6473$\pm$2001}$^*$  & \textbf{187930$\pm$19338}$^*$ & \textbf{181033$\pm$14628}  &  \textbf{141750$\pm$27426} \cr
		\bottomrule
		\bottomrule
  
    \multirow{2}{*}{ }&  \multicolumn{3}{c}{Hopper} & \multicolumn{3}{c}{Crippled Hopper}  \cr
		\cmidrule(lr){2-4} \cmidrule(lr){5-7}
		  & Training & Test (moderate) & Test (extreme) & Training & Test (moderate) & Test (extreme)\cr
		\cmidrule(lr){1-7}
		PEARL & 934$\pm$242 & 874$\pm$366 & 799$\pm$298 & 3091$\pm$298 & 2387$\pm$656 & 456$\pm$235  \cr
		TESAC & 1492$\pm$59 & \textbf{1499$\pm$35} &  \textbf{1459$\pm$72} & \textbf{3575$\pm$192} & 3298$\pm$551 & 722$\pm$161  \cr
		CCM & 1484$\pm$54 & 1446$\pm$64 & 1452$\pm$58 &  3455$\pm$301 & \textbf{3409$\pm$239} & 1009$\pm$289  \cr
            \cmidrule(lr){1-7}
            SaTESAC & \textbf{1502$\pm$20} & 1453$\pm$39 &  1447$\pm$14  & 3391$\pm$84 & 3262$\pm$166 & 1839$\pm$130  \cr
		SaCCM & 1462$\pm$45 & 1462$\pm$14  &  1451$\pm$67 & 3449$\pm$103 & 3390$\pm$211 & \textbf{2059$\pm$221}$^*$ \cr
		\bottomrule
		\bottomrule
  
    \multirow{2}{*}{ }&  \multicolumn{3}{c}{Walker} & \multicolumn{3}{c}{Crippled Walker}  \cr
		\cmidrule(lr){2-4} \cmidrule(lr){5-7}
		  & Training & Test (moderate) & Test (extreme) & Training & Test (moderate) & Test (extreme)\cr
		\cmidrule(lr){1-7}
		PEARL & 7524$\pm$2455 & 3355$\pm$2555 & 1984$\pm$356 & 7899$\pm$2532 & 4377$\pm$2563 & 2965$\pm$1426 \cr
		TESAC & 7747$\pm$1772 & 4355$\pm$1530 &  2581$\pm$407 & 9908$\pm$1561 & 5929$\pm$1971 &  3041$\pm$912 \cr
		CCM & 8136$\pm$557 & 5476$\pm$803 &  2519$\pm$682  & 10317$\pm$1137 & 6233$\pm$1869 & 3098$\pm$821  \cr
            \cmidrule(lr){1-7}
            SaTESAC & \textbf{8675$\pm$752} & \textbf{5840$\pm$676} & \textbf{3632$\pm$404}$^*$ &  10389$\pm$1031 & \textbf{8387$\pm$1291}$^*$ &  4280$\pm$485$^*$ \cr
		SaCCM & 8361$\pm$586 & 5779$\pm$691 & 3481$\pm$332$^*$  & \textbf{10496$\pm$951} & 8235$\pm$1212  & \textbf{4824$\pm$839}$^*$ \cr
		\bottomrule
		\bottomrule
	\end{tabular}
	\label{tab:mujoco-average-return}
\end{table}

\subsection{MuJoCo} \label{section:mujoco}

\textbf{Task description.} We extended the modified MuJoCo benchmark introduced in DOMINO \citep{mu2022decomposed} and CaDM \citep{lee2020context}. It contains ten typical robotic control environments based on the MuJoCo physics engine \citep{todorov2012mujoco}. Hopper, Walker, Half-cheetah, Ant, HumanoidStandup, and SlimHumanoid are influenced by continuous environmental features (i.e., mass, damping) that affect transition dynamics. Crippled Ant, Crippled Hopper, Crippled Walker, and Crippled Half-cheetah are more challenging due to the addition of discrete environmental features (i.e., randomly crippled leg joints), requiring agents to master different skills (e.g., switching from running to crawling after a leg is crippled).

\textbf{Results and skill analysis.} 
Table \ref{tab:mujoco-average-return} shows the average return of our method and baselines on training and test tasks. SaTESAC and SaCCM achieved higher returns in most tasks, except for Ant, Half-Cheetah, and Hopper, where only a single skill was needed. For instance, the Hopper robot learned to hop forward, adapting to different mass values. When environments become complex and require diverse skills for different tasks (Crippled Ant, Crippled Hopper, Crippled Half-Cheetah, SlimHumanoid, HumanoidStandup, and Crippled Walker), SaNCE brings significant improvements. For example, when the Ant robot has 3 or 4 legs available, it learns to roll to generalise across varying mass and damping. In more challenging zero-shot settings, when only 2 legs are available, the ant robot can no longer roll and it adapts by walking using its 2 healthy legs. This aligns with the results in Table \ref{tab:mujoco-average-return}, where SaMI significantly improved performance in extreme test sets. In summary, i) SaMI helps the RL agents to be versatile and embody multiple skills; ii) SaMI leads to increased returns during training and zero-shot generalisation, especially in environments that require different skills. Our video demos\textsuperscript{\ref{github}} and visualisation results in Appendix \ref{appx:visualisation_mujoco} show different skills in all environments.

\subsection{Analysis of the $\boldsymbol{\log}$-$\boldsymbol{K}$ curse in sample-limited scenarios} \label{sec:discussion}
\begin{minipage}[b]{0.63\linewidth} 
This section analyses whether SaNCE can overcome the $\log$-$K$ curse. During training, environmental features are sampled at the start of each episode, requiring the context encoder to learn the context embedding distribution across multiple tasks. Since InfoNCE samples negative examples from all tasks and SaNCE samples from the current task, SaNCE’s negative sample space is $|c|$ times smaller than InfoNCE’s. For instance, in the SlimHumanoid environment, where both mass and damping have five values, InfoNCE’s sampling space can be 25 times larger than SaNCE’s. As shown in Tables \ref{tab:panda-gym-average-return} and \ref{tab:mujoco-average-return}, RL algorithms using SaNCE achieve better or comparable performance with significantly fewer negative samples ($K$) than InfoNCE. This suggests SaNCE effectively addresses the $\log$-$K$ curse, and the SaMI objective helps the contrastive context encoder extract critical information for downstream RL tasks. \\

The number of negative samples $K$ is influenced by two hyperparameters: \textbf{\textit{buffer size}}, which determines the negative sample space, and \textbf{\textit{contrastive batch size}}, which controls the number of samples used to train the contrastive context encoder per update. We analysed these hyperparameters further, and as shown in Figure \ref{fig:batchsize}, reductions in buffer and contrastive batch size do not significantly impact the average return for SaCCM and SaTESAC, which maintain state-of-the-art performance with small buffers and batch sizes. The results in Table \ref{tab:mujoco-average-return} correspond to a buffer size of 100,000 and a batch size of 12. Results across all environments (refer to Appendix \ref{appx:sample_space}) show that SaNCE exhibits low sensitivity to $K$, highlighting its potential to overcome the $\log$-$K$ curse.
\end{minipage}
\hfill
\begin{minipage}[b]{0.35\linewidth}
    \centering
    \includegraphics[width=1.0\linewidth]{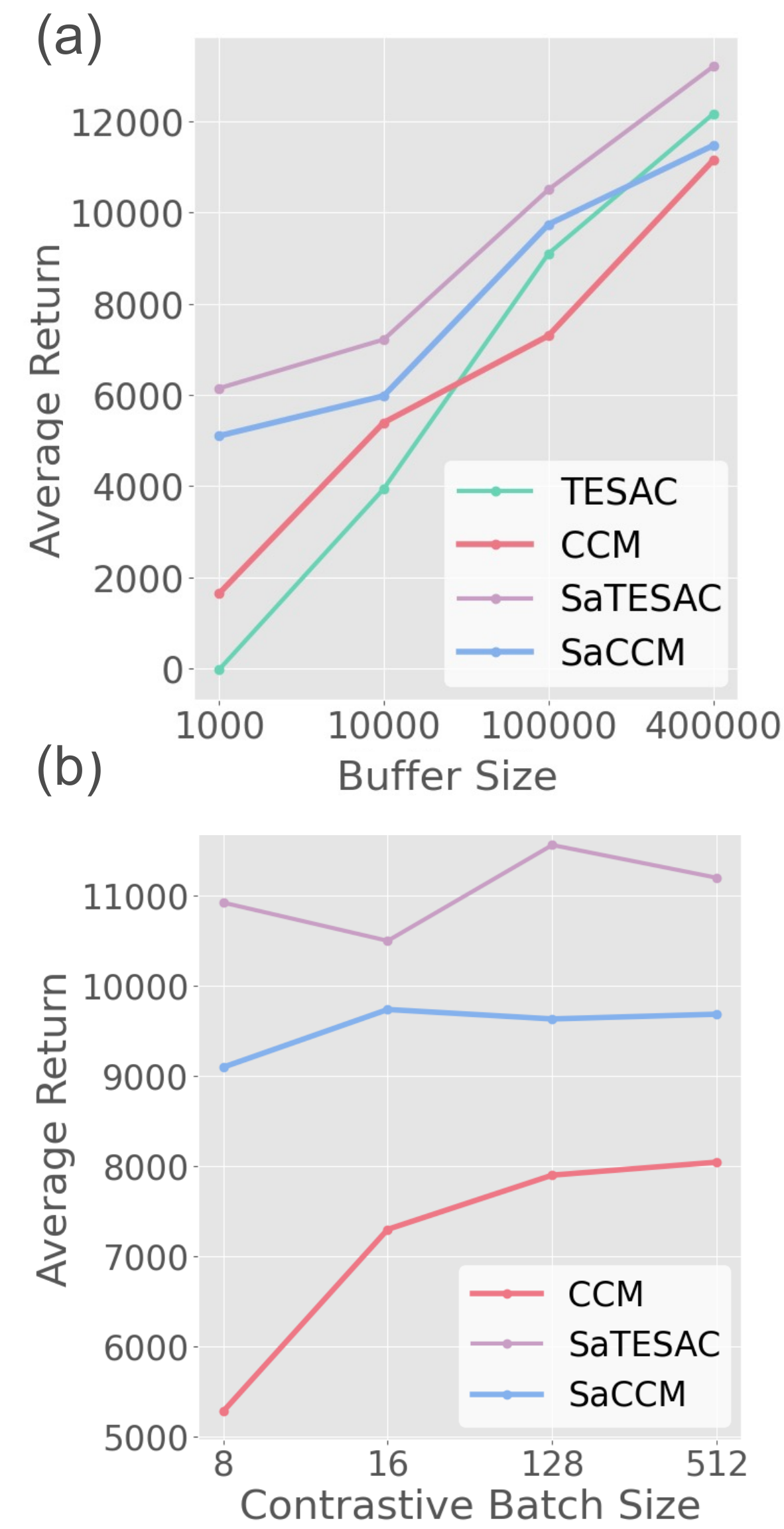}
    \captionof{figure}{Effect of (a) buffer size (TESAC, CCM, SaTESAC, SaCCM) and (b) contrastive batch size (CCM, SaTESAC, SaCCM) in the SlimHumanoid environment.}
    \label{fig:batchsize}
\end{minipage}

\section{Conclusion and future work} \label{sec:future_work}
Zero-shot generalisation has been a longstanding challenge concerning Meta-RL agents, with skill diversity and sample efficiency being key to generalising to previously unseen environments. In this paper, we proposed Skill-aware Mutual Information (SaMI) to learn context embeddings for zero-shot generalisation in downstream RL tasks, and Skill-aware Noise Contrastive Estimation (SaNCE) to optimise SaMI and overcome the $\log$-$K$ curse, along with a practical skill-aware trajectory sampling strategy. Experimental results showed that RL algorithms equipped with SaMI achieved state-of-the-art performance in MuJoCo and Panda-gym benchmarks, particularly in zero-shot generalisation within more complex environments. During the zero-shot generalisation, when faced with an unseen task, SaMI assists agents in exploring effectively, inferring context, and rapidly adapting their skills within the current episode. SaNCE’s optimisation uses a significantly smaller negative sample space than baselines, and our analysis on buffer and contrastive batch sizes demonstrated its effectiveness in addressing the $\log$-$K$ curse.

Given that environmental features are often interdependent, such as a cube's material correlating with friction and mass, SaMI does not introduce independence assumptions like DOMINO \citep{mu2022decomposed}. Therefore, future work will focus on verifying and enhancing SaMI's potential in more complex tasks where environmental features are correlated \citep{dunion2023conditional}. This will contribute to our ultimate goal: developing a generalist and versatile agent capable of working across multiple tasks and even real-world tasks in the near future.

\bibliography{mybib.bib}

\begin{thebibliography}{55}
\providecommand{\natexlab}[1]{#1}
\providecommand{\url}[1]{\texttt{#1}}
\expandafter\ifx\csname urlstyle\endcsname\relax
  \providecommand{\doi}[1]{doi: #1}\else
  \providecommand{\doi}{doi: \begingroup \urlstyle{rm}\Url}\fi

\bibitem[Agarwal et~al.(2021)Agarwal, Machado, Castro, and Bellemare]{agarwal2021contrastive}
Rishabh Agarwal, Marlos~C Machado, Pablo~Samuel Castro, and Marc~G Bellemare.
\newblock Contrastive behavioral similarity embeddings for generalization in reinforcement learning.
\newblock \emph{arXiv preprint arXiv:2101.05265}, 2021.

\bibitem[Arora et~al.(2019)Arora, Khandeparkar, Khodak, Plevrakis, and Saunshi]{arora2019theoretical}
Sanjeev Arora, Hrishikesh Khandeparkar, Mikhail Khodak, Orestis Plevrakis, and Nikunj Saunshi.
\newblock A theoretical analysis of contrastive unsupervised representation learning.
\newblock \emph{arXiv preprint arXiv:1902.09229}, 2019.

\bibitem[Bachman et~al.(2019)Bachman, Hjelm, and Buchwalter]{bachman2019learning}
Philip Bachman, R~Devon Hjelm, and William Buchwalter.
\newblock Learning representations by maximizing mutual information across views.
\newblock \emph{Advances in neural information processing systems}, 32, 2019.

\bibitem[Chen et~al.(2021)Chen, Gan, Li, Guo, Chen, Gao, Chung, Xu, Zeng, Lu, et~al.]{chen2021simpler}
Junya Chen, Zhe Gan, Xuan Li, Qing Guo, Liqun Chen, Shuyang Gao, Tagyoung Chung, Yi~Xu, Belinda Zeng, Wenlian Lu, et~al.
\newblock Simpler, faster, stronger: Breaking the log-k curse on contrastive learners with flatnce.
\newblock \emph{arXiv preprint arXiv:2107.01152}, 2021.

\bibitem[Clavera et~al.(2019{\natexlab{a}})Clavera, Nagabandi, Liu, Fearing, Abbeel, Levine, and Finn]{nagabandi2018learning}
Ignasi Clavera, Anusha Nagabandi, Simin Liu, Ronald~S. Fearing, Pieter Abbeel, Sergey Levine, and Chelsea Finn.
\newblock Learning to adapt in dynamic, real-world environments through meta-reinforcement learning.
\newblock In \emph{International Conference on Learning Representations}, 2019{\natexlab{a}}.

\bibitem[Clavera et~al.(2019{\natexlab{b}})Clavera, Nagabandi, Liu, Fearing, Abbeel, Levine, and Finn]{nagabandi2019learning}
Ignasi Clavera, Anusha Nagabandi, Simin Liu, Ronald~S. Fearing, Pieter Abbeel, Sergey Levine, and Chelsea Finn.
\newblock Learning to adapt in dynamic, real-world environments through meta-reinforcement learning.
\newblock In \emph{International Conference on Learning Representations}, 2019{\natexlab{b}}.

\bibitem[des Combes et~al.(2018)des Combes, Bachman, and van Seijen]{tachet2018learning}
Remi~Tachet des Combes, Philip Bachman, and Harm van Seijen.
\newblock Learning invariances for policy generalization, 2018.
\newblock URL \url{https://openreview.net/forum?id=BJHRaK1PG}.

\bibitem[Dunion and Albrecht(2024)]{dunion2024mvd}
Mhairi Dunion and Stefano~V Albrecht.
\newblock Multi-view disentanglement for reinforcement learning with multiple cameras.
\newblock In \emph{Reinforcement Learning Conference}, 2024.

\bibitem[Dunion et~al.(2023{\natexlab{a}})Dunion, McInroe, Luck, Hanna, and Albrecht]{dunion2023conditional}
Mhairi Dunion, Trevor McInroe, Kevin~Sebastian Luck, Josiah~P. Hanna, and Stefano~V Albrecht.
\newblock Conditional mutual information for disentangled representations in reinforcement learning.
\newblock In \emph{Thirty-seventh Conference on Neural Information Processing Systems}, 2023{\natexlab{a}}.

\bibitem[Dunion et~al.(2023{\natexlab{b}})Dunion, McInroe, Luck, Hanna, and Albrecht]{dunion2023temporal}
Mhairi Dunion, Trevor McInroe, Kevin~Sebastian Luck, Josiah~P. Hanna, and Stefano~V Albrecht.
\newblock Temporal disentanglement of representations for improved generalisation in reinforcement learning.
\newblock In \emph{The Eleventh International Conference on Learning Representations}, 2023{\natexlab{b}}.

\bibitem[Eysenbach et~al.(2018)Eysenbach, Gupta, Ibarz, and Levine]{eysenbach2018diversityneedlearningskills}
Benjamin Eysenbach, Abhishek Gupta, Julian Ibarz, and Sergey Levine.
\newblock Diversity is all you need: Learning skills without a reward function, 2018.
\newblock URL \url{https://arxiv.org/abs/1802.06070}.

\bibitem[Eysenbach et~al.(2022)Eysenbach, Zhang, Levine, and Salakhutdinov]{eysenbach2022contrastive}
Benjamin Eysenbach, Tianjun Zhang, Sergey Levine, and Russ~R Salakhutdinov.
\newblock Contrastive learning as goal-conditioned reinforcement learning.
\newblock \emph{Advances in Neural Information Processing Systems}, 35:\penalty0 35603--35620, 2022.

\bibitem[Franke et~al.(2021)Franke, Koehler, Biedenkapp, and Hutter]{franke2020sample}
J{\"o}rg~K.H. Franke, Gregor Koehler, Andr{\'e} Biedenkapp, and Frank Hutter.
\newblock Sample-efficient automated deep reinforcement learning.
\newblock In \emph{International Conference on Learning Representations}, 2021.

\bibitem[Fu et~al.(2021)Fu, Tang, Hao, Chen, Feng, Li, and Liu]{fu2021towards}
Haotian Fu, Hongyao Tang, Jianye Hao, Chen Chen, Xidong Feng, Dong Li, and Wulong Liu.
\newblock Towards effective context for meta-reinforcement learning: an approach based on contrastive learning.
\newblock In \emph{Proceedings of the AAAI Conference on Artificial Intelligence}, pages 7457--7465, 2021.

\bibitem[Gallou{\'e}dec et~al.(2021)Gallou{\'e}dec, Cazin, Dellandr{\'e}a, and Chen]{gallouedec2021pandagym}
Quentin Gallou{\'e}dec, Nicolas Cazin, Emmanuel Dellandr{\'e}a, and Liming Chen.
\newblock {panda-gym: Open-Source Goal-Conditioned Environments for Robotic Learning}.
\newblock \emph{4th Robot Learning Workshop: Self-Supervised and Lifelong Learning at NeurIPS}, 2021.

\bibitem[Garcin et~al.(2024)Garcin, Doran, Guo, Lucas, and Albrecht]{garcin2024dred}
Samuel Garcin, James Doran, Shangmin Guo, Christopher~G. Lucas, and Stefano~V. Albrecht.
\newblock {DRED}: Zero-shot transfer in reinforcement learning via data-regularised environment design.
\newblock In \emph{International Conference on Machine Learning (ICML)}, 2024.

\bibitem[Goldfeld et~al.(2019)Goldfeld, van~den Berg, Greenewald, Melnyk, Nguyen, Kingsbury, and Polyanskiy]{goldfeld2019estimatinginformationflowdeep}
Ziv Goldfeld, Ewout van~den Berg, Kristjan Greenewald, Igor Melnyk, Nam Nguyen, Brian Kingsbury, and Yury Polyanskiy.
\newblock Estimating information flow in deep neural networks, 2019.
\newblock URL \url{https://arxiv.org/abs/1810.05728}.

\bibitem[Guo et~al.(2022)Guo, Chen, Wang, Yang, Deng, Huang, Carin, Li, and Tao]{guo2022tight}
Qing Guo, Junya Chen, Dong Wang, Yuewei Yang, Xinwei Deng, Jing Huang, Larry Carin, Fan Li, and Chenyang Tao.
\newblock Tight mutual information estimation with contrastive fenchel-legendre optimization.
\newblock \emph{Advances in Neural Information Processing Systems}, 35:\penalty0 28319--28334, 2022.

\bibitem[Haarnoja et~al.(2018)Haarnoja, Zhou, Abbeel, and Levine]{haarnoja2018soft}
Tuomas Haarnoja, Aurick Zhou, Pieter Abbeel, and Sergey Levine.
\newblock Soft actor-critic: Off-policy maximum entropy deep reinforcement learning with a stochastic actor.
\newblock In \emph{International conference on machine learning}, pages 1861--1870. PMLR, 2018.

\bibitem[Hausman et~al.(2018)Hausman, Springenberg, Wang, Heess, and Riedmiller]{hausman2018learning}
Karol Hausman, Jost~Tobias Springenberg, Ziyu Wang, Nicolas Heess, and Martin Riedmiller.
\newblock Learning an embedding space for transferable robot skills.
\newblock In \emph{International Conference on Learning Representations}, 2018.

\bibitem[He et~al.(2020)He, Fan, Wu, Xie, and Girshick]{he2020momentum}
Kaiming He, Haoqi Fan, Yuxin Wu, Saining Xie, and Ross Girshick.
\newblock Momentum contrast for unsupervised visual representation learning.
\newblock In \emph{Proceedings of the IEEE/CVF conference on computer vision and pattern recognition}, pages 9729--9738, 2020.

\bibitem[Hjelm et~al.(2019)Hjelm, Fedorov, Lavoie-Marchildon, Grewal, Bachman, Trischler, and Bengio]{hjelm2019learning}
R~Devon Hjelm, Alex Fedorov, Samuel Lavoie-Marchildon, Karan Grewal, Phil Bachman, Adam Trischler, and Yoshua Bengio.
\newblock Learning deep representations by mutual information estimation and maximization.
\newblock In \emph{International Conference on Learning Representations}, 2019.

\bibitem[Jaderberg et~al.(2016)Jaderberg, Mnih, Czarnecki, Schaul, Leibo, Silver, and Kavukcuoglu]{jaderberg2016reinforcement}
Max Jaderberg, Volodymyr Mnih, Wojciech~Marian Czarnecki, Tom Schaul, Joel~Z Leibo, David Silver, and Koray Kavukcuoglu.
\newblock Reinforcement learning with unsupervised auxiliary tasks, 2016.

\bibitem[Kirk et~al.(2023)Kirk, Zhang, Grefenstette, and Rockt{\"a}schel]{kirk2023survey}
Robert Kirk, Amy Zhang, Edward Grefenstette, and Tim Rockt{\"a}schel.
\newblock A survey of zero-shot generalisation in deep reinforcement learning.
\newblock \emph{Journal of Artificial Intelligence Research}, 76:\penalty0 201--264, 2023.

\bibitem[Laskin et~al.(2020)Laskin, Srinivas, and Abbeel]{laskin2020curl}
Michael Laskin, Aravind Srinivas, and Pieter Abbeel.
\newblock Curl: Contrastive unsupervised representations for reinforcement learning.
\newblock In \emph{International conference on machine learning}, pages 5639--5650. PMLR, 2020.

\bibitem[Lee et~al.(2020)Lee, Seo, Lee, Lee, and Shin]{lee2020context}
Kimin Lee, Younggyo Seo, Seunghyun Lee, Honglak Lee, and Jinwoo Shin.
\newblock Context-aware dynamics model for generalization in model-based reinforcement learning.
\newblock In \emph{International Conference on Machine Learning}, pages 5757--5766. PMLR, 2020.

\bibitem[Li et~al.(2021)Li, Huang, Chen, Luo, Luo, and Huang]{li2021provably}
Lanqing Li, Yuanhao Huang, Mingzhe Chen, Siteng Luo, Dijun Luo, and Junzhou Huang.
\newblock Provably improved context-based offline meta-rl with attention and contrastive learning.
\newblock \emph{arXiv preprint arXiv:2102.10774}, 2021.

\bibitem[Li(2022)]{li2022functional}
Qiang Li.
\newblock Functional connectivity inference from fmri data using multivariate information measures.
\newblock \emph{Neural Networks}, 146:\penalty0 85--97, 2022.

\bibitem[McGill(1954)]{mcgill1954multivariate}
William McGill.
\newblock Multivariate information transmission.
\newblock \emph{Transactions of the IRE Professional Group on Information Theory}, 4\penalty0 (4):\penalty0 93--111, 1954.

\bibitem[McInnes et~al.(2020)McInnes, Healy, and Melville]{mcinnes2020umapuniformmanifoldapproximation}
Leland McInnes, John Healy, and James Melville.
\newblock Umap: Uniform manifold approximation and projection for dimension reduction, 2020.
\newblock URL \url{https://arxiv.org/abs/1802.03426}.

\bibitem[McInroe et~al.(2024)McInroe, Schäfer, and Albrecht]{mcinroe2023hksl}
Trevor McInroe, Lukas Schäfer, and Stefano~V. Albrecht.
\newblock Multi-horizon representations with hierarchical forward models for reinforcement learning.
\newblock \emph{Transactions on Machine Learning Research (TMLR)}, 2024.

\bibitem[Mnih and Teh(2012)]{mnih2012fast}
Andriy Mnih and Yee~Whye Teh.
\newblock A fast and simple algorithm for training neural probabilistic language models.
\newblock \emph{arXiv preprint arXiv:1206.6426}, 2012.

\bibitem[Mu et~al.(2022)Mu, Zhuang, Ni, Wang, Chen, HAO, and Luo]{mu2022decomposed}
Yao Mu, Yuzheng Zhuang, Fei Ni, Bin Wang, Jianyu Chen, Jianye HAO, and Ping Luo.
\newblock {DOMINO}: Decomposed mutual information optimization for generalized context in meta-reinforcement learning.
\newblock In Alice~H. Oh, Alekh Agarwal, Danielle Belgrave, and Kyunghyun Cho, editors, \emph{Advances in Neural Information Processing Systems}, 2022.

\bibitem[Neuberg(2003)]{neuberg2003causality}
Leland~Gerson Neuberg.
\newblock Causality: models, reasoning, and inference, by judea pearl, cambridge university press, 2000.
\newblock \emph{Econometric Theory}, 19\penalty0 (4):\penalty0 675--685, 2003.

\bibitem[Nozawa and Sato(2021)]{nozawa2021understanding}
Kento Nozawa and Issei Sato.
\newblock Understanding negative samples in instance discriminative self-supervised representation learning.
\newblock \emph{Advances in Neural Information Processing Systems}, 34:\penalty0 5784--5797, 2021.

\bibitem[Oord et~al.(2019)Oord, Li, and Vinyals]{oord2019representation}
Aaron van~den Oord, Yazhe Li, and Oriol Vinyals.
\newblock Representation learning with contrastive predictive coding.
\newblock \emph{arXiv preprint arXiv:1807.03748}, 2019.

\bibitem[Poole et~al.(2019)Poole, Ozair, Van Den~Oord, Alemi, and Tucker]{poole2019variational}
Ben Poole, Sherjil Ozair, Aaron Van Den~Oord, Alex Alemi, and George Tucker.
\newblock On variational bounds of mutual information.
\newblock In \emph{International Conference on Machine Learning}, pages 5171--5180. PMLR, 2019.

\bibitem[Puterman(2014)]{puterman2014markov}
Martin~L Puterman.
\newblock \emph{Markov decision processes: discrete stochastic dynamic programming}.
\newblock John Wiley \& Sons, 2014.

\bibitem[Raffin et~al.(2021)Raffin, Hill, Gleave, Kanervisto, Ernestus, and Dormann]{stable-baselines3}
Antonin Raffin, Ashley Hill, Adam Gleave, Anssi Kanervisto, Maximilian Ernestus, and Noah Dormann.
\newblock Stable-baselines3: Reliable reinforcement learning implementations.
\newblock \emph{Journal of Machine Learning Research}, 22\penalty0 (268):\penalty0 1--8, 2021.
\newblock URL \url{http://jmlr.org/papers/v22/20-1364.html}.

\bibitem[Rakelly et~al.(2019)Rakelly, Zhou, Finn, Levine, and Quillen]{rakelly2019efficient}
Kate Rakelly, Aurick Zhou, Chelsea Finn, Sergey Levine, and Deirdre Quillen.
\newblock Efficient off-policy meta-reinforcement learning via probabilistic context variables.
\newblock In \emph{International conference on machine learning}, pages 5331--5340. PMLR, 2019.

\bibitem[Rice and Rice(2007)]{rice2007mathematical}
John~A Rice and John~A Rice.
\newblock \emph{Mathematical statistics and data analysis}, volume 371.
\newblock Thomson/Brooks/Cole Belmont, CA, 2007.

\bibitem[Robinson et~al.(2021)Robinson, Chuang, Sra, and Jegelka]{robinson2021contrastivelearninghardnegative}
Joshua Robinson, Ching-Yao Chuang, Suvrit Sra, and Stefanie Jegelka.
\newblock Contrastive learning with hard negative samples, 2021.
\newblock URL \url{https://arxiv.org/abs/2010.04592}.

\bibitem[Sang et~al.(2022)Sang, Tang, Ma, Hao, Zheng, Meng, Li, and Wang]{sang2022pandr}
Tong Sang, Hongyao Tang, Yi~Ma, Jianye Hao, Yan Zheng, Zhaopeng Meng, Boyan Li, and Zhen Wang.
\newblock Pandr: Fast adaptation to new environments from offline experiences via decoupling policy and environment representations, 2022.

\bibitem[Seo et~al.(2020)Seo, Lee, Clavera~Gilaberte, Kurutach, Shin, and Abbeel]{seo2020trajectory}
Younggyo Seo, Kimin Lee, Ignasi Clavera~Gilaberte, Thanard Kurutach, Jinwoo Shin, and Pieter Abbeel.
\newblock Trajectory-wise multiple choice learning for dynamics generalization in reinforcement learning.
\newblock \emph{Advances in Neural Information Processing Systems}, 33:\penalty0 12968--12979, 2020.

\bibitem[Shi et~al.(2022)Shi, Lim, and Lee]{shi2022skimo}
Lucy~Xiaoyang Shi, Joseph~J. Lim, and Youngwoon Lee.
\newblock Skill-based model-based reinforcement learning.
\newblock In \emph{Conference on Robot Learning}, 2022.

\bibitem[Song and Ermon(2020)]{song2019understanding}
Jiaming Song and Stefano Ermon.
\newblock Understanding the limitations of variational mutual information estimators.
\newblock In \emph{International Conference on Learning Representations}, 2020.

\bibitem[Tabak(2014)]{tabak2014geometry}
John Tabak.
\newblock \emph{Geometry: the language of space and form}.
\newblock Infobase Publishing, 2014.

\bibitem[Tishby and Zaslavsky(2015)]{tishby2015deep}
Naftali Tishby and Noga Zaslavsky.
\newblock Deep learning and the information bottleneck principle.
\newblock In \emph{2015 ieee information theory workshop (itw)}, pages 1--5. IEEE, 2015.

\bibitem[Todorov et~al.(2012)Todorov, Erez, and Tassa]{todorov2012mujoco}
Emanuel Todorov, Tom Erez, and Yuval Tassa.
\newblock Mujoco: A physics engine for model-based control.
\newblock In \emph{2012 IEEE/RSJ international conference on intelligent robots and systems}, pages 5026--5033. IEEE, 2012.

\bibitem[Van~der Maaten and Hinton(2008)]{van2008visualizing}
Laurens Van~der Maaten and Geoffrey Hinton.
\newblock Visualizing data using t-sne.
\newblock \emph{Journal of machine learning research}, 9\penalty0 (11), 2008.

\bibitem[Wang et~al.(2021)Wang, Xu, Keutzer, Gao, and Wu]{wang2021improving}
Bernie Wang, Simon Xu, Kurt Keutzer, Yang Gao, and Bichen Wu.
\newblock Improving context-based meta-reinforcement learning with self-supervised trajectory contrastive learning, 2021.

\bibitem[Waskom(2021)]{Waskom2021}
Michael~L. Waskom.
\newblock seaborn: statistical data visualization.
\newblock \emph{Journal of Open Source Software}, 6\penalty0 (60):\penalty0 3021, 2021.
\newblock \doi{10.21105/joss.03021}.
\newblock URL \url{https://doi.org/10.21105/joss.03021}.

\bibitem[Wu et~al.(2018)Wu, Xiong, Yu, and Lin]{wu2018unsupervised}
Zhirong Wu, Yuanjun Xiong, Stella~X Yu, and Dahua Lin.
\newblock Unsupervised feature learning via non-parametric instance discrimination.
\newblock In \emph{Proceedings of the IEEE conference on computer vision and pattern recognition}, pages 3733--3742, 2018.

\bibitem[Yu et~al.(2020)Yu, Quillen, He, Julian, Hausman, Finn, and Levine]{yu2020meta}
Tianhe Yu, Deirdre Quillen, Zhanpeng He, Ryan Julian, Karol Hausman, Chelsea Finn, and Sergey Levine.
\newblock Meta-world: A benchmark and evaluation for multi-task and meta reinforcement learning.
\newblock In \emph{Conference on robot learning}, pages 1094--1100. PMLR, 2020.

\bibitem[Zhou et~al.(2019)Zhou, Pinto, and Gupta]{zhou2019environment}
Wenxuan Zhou, Lerrel Pinto, and Abhinav Gupta.
\newblock Environment probing interaction policies.
\newblock In \emph{International Conference on Learning Representations}, 2019.

\end{thebibliography}

\clearpage
\appendix

\section{Proof of Lemma \ref{lemma:lemma_1}} \label{appx:lemma_1}

Given a query $x$ and a set $Y=\{y_1, \dots, y_K\}$ of $K$ random samples, containing one positive sample $y_1$ and $K-1$ negative samples drawn from the distribution $p(y)$, a $K$-sample InfoNCE estimator is obtained by comparing pairs sampled from the joint distribution $(x, y_1) \sim p(x, y)$ with pairs $(x, y_k)$, constructed using the set of negative examples $y_{2:K}$. InfoNCE compares the positive pairs $(x, y_1)$ with the negative pairs $(x, y_k)$, where $y_k \sim y_{2:K}$, as follows:
\begin{align}
I_{\text{InfoNCE}}(x;y|\psi,K) & = \mathbb{E}_{p(x,y_1)p(y_{2:K})} \left[ \log \left( \frac{f_{\psi}(x,y_1)}{ \frac{1}{K}\sum_{k=1}^K f_{\psi}(x,y_k)} \right) \right]
\label{eq:infonce_definition} 
\end{align}

\textbf{Step 1.} Let us prove that the $K$-sample InfoNCE estimator is upper-bounded by $\log K$. According to \cite{mu2022decomposed}, $\frac{f_{\psi}(x,y_1)}{\sum_{k=1}^K f_{\psi}(x,y_k)} = \frac{f_{\psi}(x,y_1)}{f_{\psi}(x,y_1) + \sum_{k=2}^K f_{\psi}(x,y_k)} \leq 1$. So we have:
\begin{align}
I_{\text{InfoNCE}}(x;y|\psi,K) & = \mathbb{E}_{p(x,y_1)p(y_{2:K})} \left[ \log \left( \frac{f_{\psi}(x,y_1)}{\frac{1}{K}\sum_{k=1}^K f_{\psi}(x,y_k)} \right) \right] \nonumber \\
& = \mathbb{E}_{p(x,y)} \left[ \mathbb{E}_{p(y_{2:K})} \log \left( \frac{ K \cdot f_{\psi}(x,y_1)}{\sum_{k=1}^K f_{\psi}(x,y_k)} \right) \right] \nonumber \\
& \leq \log K \label{eq:infonce_upper_bound_logk} 
\end{align}
Hence, we have $I_{\text{InfoNCE}}(x;y|\psi,K) \leq \log K$.

\textbf{Step 2.} We have the $I(x;y) \geq I_{\text{InfoNCE}}(x;y|\psi,K)$ according to:
\begin{prop} \citep{poole2019variational}
A $K$-sample estimator is an asymptotically tight lower bound on the MI, i.e.,
$$I(x;y) \geq I_{\text{InfoNCE}}(x;y|\psi,K),\lim_{x \to +\infty} I_{\text{InfoNCE}}(x;y|\psi,K) \to I(x;y)$$
\label{pro:tight_lower_bound}
\end{prop}
\textit{Proof}. See \cite{poole2019variational} for a neat proof of how the multi-sample estimator (e.g., InfoNCE) lower bounds MI.

\textbf{Step 3.} In this research, the context encoder $\psi$ in $f_{\psi}(x,y)$ is implemented using an RNN to approximate $\frac{p(y|x)}{p(y)}$ \citep{oord2019representation}. With a sufficiently powerful deep learning model for $\psi$ and a finite sample size $K$, such that $I(x;y) \geq \log K$, we can reasonably expect that $I_{\text{InfoNCE}} \approx \log K$ after a few training epochs. Therefore, during training, when $K \ll +\infty$, we always have $I(x;y) \geq \log K$.

\textit{Proof}. See \cite{chen2021simpler} for more detailed proof.

\textbf{Step 4.} Let us prove that the $K$-sample InfoNCE bound is asymptotically tight. The specific choice of the context encoder $\psi$ influences the tightness of the $K$-sample NCE bound. InfoNCE \citep{oord2019representation} sets $f_{\psi}(x, y) \propto \frac{p(y|x)}{p(y)}$ to model a density ratio that preserves the MI between $x$ and $y$, where $\propto$ stands for 'proportional to' (i.e., up to a multiplicative constant). Substituting $f_{\psi}(x, y) = f_{\psi}^*(x, y) = \frac{p(y|x)}{p(y)}$ into InfoNCE, we obtain:
\begin{align}
    I_{\text{InfoNCE}}(x;y|\psi,K) & = \mathbb{E} \left[ \log \left( \frac{f_{\psi}^*(x,y_1)}{\sum_{k=1}^K f_{\psi}^*(x,y_k)} \right) \right] + \log K \nonumber \\
    & = - \mathbb{E} \left[ \log \left( 1 + \frac{p(y)}{p(y|x)}\sum_{k=2}^K \frac{p(y_k|x)}{p(y_k)} \right) \right] + \log K \nonumber \\
    & \approx - \mathbb{E} \left[ \log \left( 1 + \frac{p(y)}{p(y|x)}(K-1) \mathbb{E}_{y_k\sim p(y)} \frac{p(y_k|x)}{p(y_k)} \right) \right] + \log K \nonumber \\
    & = - \mathbb{E} \left[ \log \left( 1 + \frac{p(y_1)}{p(y_1|x)}(K-1) \right) \right] + \log K \nonumber \\
    & \approx - \mathbb{E} \left[ \log \frac{p(y)}{p(y|x)} \right] - \log (K-1) + \log K \nonumber \\
    & = I(x;y) - \log (K-1) + \log K \label{eq:upper_bound_logK} 
\end{align}

Now taking $K \to +\infty$, the last two terms cancel out.

\textbf{Putting it together.} Combining $I(x;y) \geq \log K$ with Proposition \ref{pro:tight_lower_bound} and Equation \eqref{eq:infonce_upper_bound_logk}, we have Lemma \ref{lemma:lemma_1}:
\begin{equation}
    {I}_{\text{InfoNCE}}(x;y|\psi,K) \leq \log K \leq I(x;y).
    \label{eq:infonce_logk_i}
\end{equation}
Moreover, according to Equation \eqref{eq:upper_bound_logK}, as the sample size $K \to +\infty$, the $K$-sample InfoNCE bound becomes sharp and approaches the true MI $I(x;y)$, i.e., $I_{\text{InfoNCE}}(x;y|\psi,K) \approx \log K \approx I(x;y)$.

\section{Proof for Lemma \ref{lemma:lemma_2}} \label{appx:lemma_2}

\textbf{Step 1.} According to Lemma \ref{lemma:lemma_1}, we have $\textcolor{mylightblue}{I_{\text{InfoNCE}}(c;\pi_c;\tau_c|\psi,K)} \leq \textcolor{mygrey}{\log K} \leq \textcolor{mypurple}{I_{\text{SaMI}}(c;\pi_c;\tau_c)}$ (shown in Figure \ref{fig:sample_size_mi}). 

\textbf{Step 2.} Let us prove that SaNCE is a $K$-sample SaNCE estimator and is upper bounded by $\textcolor{mygrey}{\log K}$. Since $\frac{f_{\psi}(c,\pi_{c},\tau_{c}^+)}{f_{\psi}(c,\pi_{c},\tau_{c}^+) + \sum_{k=2}^K f_{\psi}(c,\pi_{c},\tau_{c,k}^-)} \leq 1$ \citep{mu2022decomposed}, we have:
\begin{align}
& \quad \textcolor{myyellow}{I_{\text{SaNCE}}(c;\pi_c;\tau_c|\psi,K)} \nonumber  \\
& = \mathbb{E}_{p(c_1,\pi_{c_1},\tau_{c_1}^+)p(\tau_{c_1,2:K}^-)} \left[ \log \left( \frac{K \cdot f_{\psi}(c_1,\pi_{c_1},\tau_{c_1}^+)}{f_{\psi}(c_1,\pi_{c_1},\tau_{c_1}^+) + \sum_{k=2}^{K} f_{\psi}(c_1,\pi_{c_1},\tau_{c_1,k}^-)} \right) \right] \nonumber  \\
& = \mathbb{E}_{p(c_1,\pi_{c_1})} \left[ \mathbb{E}_{p(\tau_{c_1,2:K}^-)} \log \left( \frac{ K \cdot f_{\psi}(c_1,\pi_{c_1},\tau_{c_1}^+)}{f_{\psi}(c_1,\pi_{c_1},\tau_{c_1}^+) + \sum_{k=2}^{K} f_{\psi}(c_1,\pi_{c_1},\tau_{c_1,k}^-)} \right) \right] \nonumber \\
& \leq \textcolor{mygrey}{\log K}
\label{eq:sance_upper_bound_logk} 
\end{align}
Thus, we obtain $\textcolor{myyellow}{I_{\text{SaNCE}}(c;\pi_c;\tau_c|\psi,K)} \leq \textcolor{mygrey}{\log K}$, similar to Equation \eqref{eq:infonce_upper_bound_logk}.

\textbf{Step 3.} With the definition of $K^*$, we can prove that $\textcolor{mylightblue}{I_{\text{InfoNCE}}(c;\pi_c;\tau_c|\psi,K)} \leq \textcolor{myyellow}{I_{\text{SaNCE}}(c;\pi_c;\tau_c|\psi,K)}$ with the same sample size $K$. In task $e_1$, SaNCE obtains positive and negative samples from the current task $e_1$. Since the variable $c = c_1$ is constant, we have:
\begin{align}
& \quad \textcolor{myyellow}{I_{\text{SaNCE}}(c;\pi_c;\tau_c|\psi,K)} \nonumber  \\
& = \mathbb{E}_{p(c_1,\pi_{c_1},\tau_{c_1}^+)p(\tau_{c_1,2:K}^-)} \left[ \log \left( \frac{K \cdot f_{\psi}(c_1,\pi_{c_1},\tau_{c_1}^+)}{f_{\psi}(c_1,\pi_{c_1},\tau_{c_1}^+) + \sum_{k=2}^{K} f_{\psi}(c_1,\pi_{c_1},\tau_{c_1,k}^-)} \right) \right] \nonumber  \\
& \leq \mathbb{E}_{p(c_1,\pi_{c_1},\tau_{c_1}^+)p(\tau_{c_1,2:K^*_{\text{SaNCE}}}^-)} \left[ \log \left( \frac{K^*_{\text{SaNCE}} \cdot f_{\psi}(c_1,\pi_{c_1},\tau_{c_1}^+)}{f_{\psi}(c_1,\pi_{c_1},\tau_{c_1}^+) + \sum_{k=2}^{K^*_{\text{SaNCE}}} f_{\psi}(c_1,\pi_{c_1},\tau_{c_1,k}^-)} \right) \right] \nonumber \\
& = \mathbb{E}_{p(\pi_{c_1},\tau_{c_1}^+)p(\tau_{c_1,2:K^*_{\text{SaNCE}}}^-)} \left[ \log \left( \frac{K^*_{\text{SaNCE}} \cdot f_{\psi}(\pi_{c_1},\tau_{c_1}^+)}{f_{\psi}(\pi_{c_1},\tau_{c_1}^+) + \sum_{k=2}^{K^*_{\text{SaNCE}}} f_{\psi}(\pi_{c_1},\tau_{{c_1},k}^-)} \right) \right] \text{($c_1$ is constant.)} \nonumber  \\
& \approx \textcolor{mygrey}{\log K^*_{\text{SaNCE}}}
\label{eq:sance_desired_k} 
\end{align}
The required sample size is $K^*_{\text{SaNCE}} = |c_1| \cdot |\pi| \cdot M = |\pi| \cdot M$. As $K \to K^*_{\text{SaNCE}}$, we have $\textcolor{myyellow}{I_{\text{SaNCE}}(c;\pi_c;\tau_c|\psi,K)} \approx \textcolor{mypurple}{I_{\text{SaMI}}(c;\pi_c;\tau_c)}$. Correspondingly, for InfoNCE, in the current task $e_1$ with context embedding $c_1$, positive samples are trajectories $\tau_1$ generated after executing the skill $\pi_1$ in task $e_1$, while negative samples are trajectories $\{\tau_{c_2}^-,...\}$ from other tasks $\{e_2,...\}$. Under the definition of $K^*$, we have:
\begin{align}
& \quad \textcolor{mylightblue}{I_{\text{InfoNCE}}(c;\pi_c;\tau_c|\psi,K)} \nonumber \\
& = \mathbb{E}_{p(c,\pi_c,\tau_{c_1})p({\tau_{c_{2:|c|},2:K}})} \left[ \log \left( \frac{K \cdot f_{\psi}(c,\pi_c,\tau_{c_1})}{f_{\psi}(c,\pi_c,\tau_{c_1}) + \sum_{k=2}^K f_{\psi}(c,\pi_c,\tau_{c_{2:|c|},k})} \right) \right] \nonumber \\
& \leq \mathbb{E}_{p(c,\pi_c,\tau_{c_1})p(\tau_{c_{2:|c|},2:K^*_{\text{InfoNCE}}})} \left[ \log \left( \frac{K^*_{\text{InfoNCE}} \cdot f_{\psi}(c,\pi_c,\tau_{c_1})}{f_{\psi}(c,\pi_c,\tau_{c_1}) + \sum_{k=2}^{K^*_{\text{InfoNCE}}} f_{\psi}(c,\pi_c,\tau_{c_{2:|c|},k})} \right) \right] \nonumber \\
& \approx \textcolor{mygrey}{\log K^*_{\text{InfoNCE}}}, \label{eq:infonce_desired_k} 
\end{align}
where $K^*_{\text{InfoNCE}}=|c|\cdot|\pi|\cdot M \approx |c| \cdot K^*_{\text{SaNCE}}$. In real-world robotic control tasks, the sample space size increases significantly due to multiple environmental features $e = \{e^0, e^1, \dots, e^N\}$. The number of different tasks $|c|$ grows exponentially due to the permutations and combinations of the $N$ environmental features. When the current task $e_1$ has context embedding $c_1$, the $c_{2:|c|}$ refer to the context embeddings for the other tasks. As $K \to K^*_{\text{InfoNCE}}$, we have $\textcolor{mylightblue}{I_{\text{InfoNCE}}(c;\pi_c;\tau_c|\psi,K)} \approx \textcolor{mypurple}{I_{\text{SaMI}}(c;\pi_c;\tau_c)}$. Thus, during the training process, $\textcolor{mylightblue}{I_{\text{InfoNCE}}(c;\pi_c;\tau_c|\psi,K)} \leq \textcolor{myyellow}{I_{\text{SaNCE}}(c;\pi_c;\tau_c|\psi,K)}$ with the same sample size $K$.

\begin{figure}[ht]
\centering
\includegraphics[width=\linewidth]{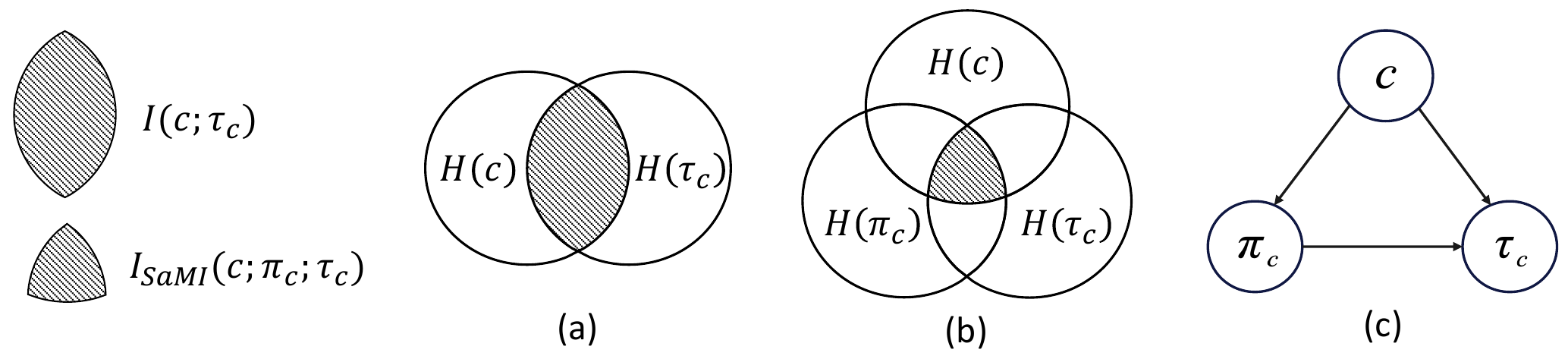}
\caption{Venn diagrams illustrating (a) mutual information \(I(c;\tau_c)\), (b) interaction information \(I_{\text{SaMI}}(c;\pi_c;\tau_c)\), and (c) the MDP graph of the context embedding \(c\), skill \(\pi_c\), and trajectory \(\tau_c\), which represents a common-cause structure \citep{neuberg2003causality}.}
\label{fig:mi_ii_causal_graph}
\end{figure}

\textbf{Step 4.} According to the definition of SaMI in Equation \eqref{eq_4_1_1}, we have \(\textcolor{mypurple}{I_{\text{SaMI}}(c;\pi_c;\tau_c)} \leq \textcolor{mydarkblue}{I(c;\tau_c)}\), as illustrated using Venn diagrams in Figure \ref{fig:mi_ii_causal_graph} (a) and (b). SaMI is formulated based on interaction information \citep{mcgill1954multivariate}, which aims to capture the relationships among multivariate variables by quantifying the amount of information (redundancy or synergy) shared among three variables. Interaction information has been less extensively studied, partly due to its challenging interpretation from both information theory and neuroscience perspectives, as it can be either positive or negative \citep{li2022functional}.

In the Meta-RL setting, the MDP (causal) graph structure, shown in Figure \ref{fig:mi_ii_causal_graph} (c), illustrates that the context embedding \(c\), skill \(\pi_c\), and trajectory \(\tau_c\) form a common-cause structure, where \(c\) acts as a shared cause influencing both the skill \(\pi_c\) and the trajectory \(\tau_c\). Consequently, \(I(\pi_c, \tau_c \mid c) < I(\pi_c, \tau_c)\). Therefore, \(I_{\text{SaMI}}(c;\pi_c;\tau_c) > 0\) is guaranteed, making it interpretable in real-world robotic control tasks.

\textbf{Putting it together.} Thus, we establish Lemma \ref{lemma:lemma_2}: we always have $\textcolor{mylightblue}{I_{\text{InfoNCE}}(c;\pi_c;\tau_c|\psi,K)} \leq \textcolor{myyellow}{I_{\text{SaNCE}}(c;\pi_c;\tau_c|\psi,K)} \leq \textcolor{mygrey}{\log K} \leq \textcolor{mypurple}{I_{\text{SaMI}}(c;\pi_c;\tau_c)} \leq \textcolor{mydarkblue}{I(c;\tau_c)}$, as shown in Figure \ref{fig:sample_size_mi}, while learning a skill-aware context encoder $\psi$ with the SaNCE estimator. Since $K^*_{\text{SaNCE}} \ll K^*_{\text{InfoNCE}}$, $\textcolor{myyellow}{I_{\text{SaNCE}}(c;\pi_c;\tau_c|\psi,K)}$ serves as a much tighter lower bound for the true \textcolor{mypurple}{$I_{\text{SaMI}}(c;\pi_c;\tau_c)$} than $\textcolor{mylightblue}{I_{\text{InfoNCE}}(c;\pi_c;\tau_c|\psi,K)}$.

\section{Sample size of $I_{\text{Sa+InfoNCE}}$} \label{appx:proof_3}

In this section, we illustrate the sample size of $I_{\text{Sa+InfoNCE}}(c;\pi_c;\tau_c|\psi,K)$. Sa+InfoNCE incorporates SaNCE into InfoNCE, using positive samples $\tau_{c_1}^+$ from task $e_1$ after executing skill $\pi_{c_1}^+$, and negative samples are trajectories $\tau_{c_{1:K}}^-$ from executing skills $\pi_{c_{1:K}}^-$ in tasks $e_{1:K}$, respectively. Therefore, this approach is equivalent to first observing the variable $c$ and then observing the variable $\pi_c$, i.e., sampling from the distribution $p(\pi_c,\tau_c|c)p(c)$. We have:
\begin{align}
& I_{\text{Sa+InfoNCE}}(c;\pi_c;\tau_c|\psi,K)  \nonumber \\
& = \mathbb{E}_{p(c)p(\pi_{c_1}^+,\tau_{c_1}^+|c)p\left(\left(\pi_{2:|c|}^-,\tau_{2:|c|}^-\right)_{2:K}\right)} \left[ \log \left( \frac{K \cdot f_{\psi}(c,\pi_{c_1}^+,\tau_{c_1}^+)}{f_{\psi}(c,\pi_{c_1}^+,\tau_{c_1}^+) + \sum_{k=2}^K f_{\psi}(c,\pi_{2:|c|,k}^-,\tau_{2:|c|,k}^-)} \right) \right] \nonumber \\
& \leq \mathbb{E}_{p(c)p(\pi_{c_1}^+,\tau_{c_1}^+|c)p\left(\left(\pi_{2:|c|}^-,\tau_{2:|c|}^-\right)_{2:K^*_{\text{Sa+InfoNCE}}}\right)} \left[ \log \left( \frac{K^*_{\text{Sa+InfoNCE}} \cdot f_{\psi}(c,\pi_{c_1}^+,\tau_{c_1}^+)}{f_{\psi}(c,\pi_{c_1}^+,\tau_{c_1}^+) + \sum_{k=2}^{K^*_{\text{Sa+InfoNCE}}} f_{\psi}(c,\pi_{2:|c|,k}^-,\tau_{2:|c|,k}^-)} \right) \right] \nonumber \\
& \approx \textcolor{mygrey}{\log K^*_{\text{Sa+InfoNCE}}}
\label{eq:sa_infonce_desired_k} 
\end{align}
It should be noted that such a combination increases the size of the negative sample space, i.e., $K^*_{\text{Sa+InfoNCE}} = \left(\sum_{i=1}^{|c|}|\pi_{c_i}^-| + |\pi_{c_1}^+|\right) \cdot M \geq K^*_{\text{SaNCE}}$. The misaligned bars in Figure \ref{fig:fig_sample_space} (c) illustrate that negative sample spaces may vary across tasks. This variation arises because we define negative samples as trajectories with low returns, making the size of the negative sample space influenced by sampling randomness. With the same number $K$ of samples, $I_{\text{Sa+InfoNCE}}$ is less precise and looser than $I_{\text{SaNCE}}$. Therefore, a trade-off between sample diversity and the precision of the $K$-sample estimator is required.

\section{Environmental setup} \label{appx:environment_setup}

\begin{figure}[ht]
\centering
\includegraphics[width=0.7\linewidth]{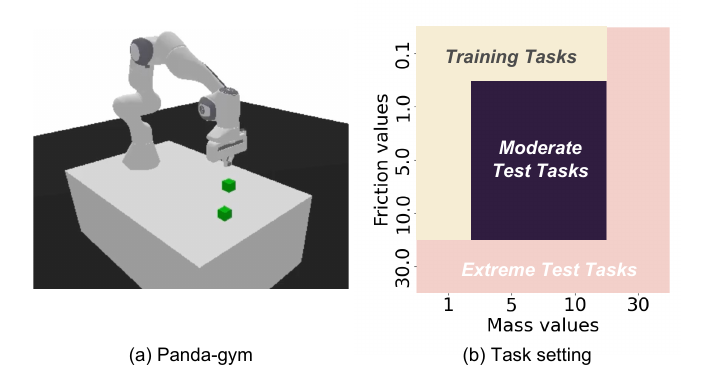}
\caption{(a) Modified Panda-gym benchmarks, (b) the training tasks, moderate test tasks, and extreme test tasks. The moderate test task setting involves combinatorial interpolation, while the extreme test task setting includes unseen ranges of environmental features and represents an extrapolation.}
\label{fig:benchmarks_panda}
\end{figure}

\subsection{Modified Panda-gym}
We modified the original Pick\&Place task in Panda-gym \citep{gallouedec2021pandagym} by setting the $z$ dimension (i.e., the desired height) of the cube's goal position to $0$ \footnote{If $z$ is not equal to 0, the Pick\&Place skill is always required to solve the tasks.} and maintaining the freedom of the grippers \footnote{In the original Push task, the grippers are blocked to ensure the agent can only push cubes. However, this restriction prevents the agent from learning Pick\&Place skills, leading to "unpushable" failure in Figure \ref{fig:example} (b).}, allowing the agent to explore whether it should push or grasp the cube. \textbf{\textit{Skills}} in this benchmark are defined as:
\begin{itemize}
    \item \textit{Pick\&Place skill}: This skill specifically refers to the agent using the gripper to grasp the cube, lift it off the table, and place it in the goal position. We determine the Pick\&Place skill by detecting no contact points between the table and the cube, two contact points between the robot’s end effector and the cube, and the cube’s height being greater than half its width.
    \item \textit{Push skill}: This skill involves the agent moving the cube on the table to the goal position, either by dragging or sliding it. We confirm the Push skill by detecting that the cube’s height equals half its width.
    \item \textit{Other skills}: Any behaviour modes other than Pick\&Place and Push are classified as other skills. 
\end{itemize}

Some elements in the RL framework are defined as follows:

\textbf{\textit{State space}}: We use feature vectors that contain the cube’s position (3 dimensions), cube rotation (3 dimensions), cube velocity (3 dimensions), cube angular velocity (3 dimensions), end-effector position (3 dimensions), end-effector velocity (3 dimensions), gripper width (1 dimension), desired goal (3 dimensions), and achieved goal (3 dimensions). Environmental features are not included in states.

\textbf{\textit{Action space}}: The action space has 4 dimensions; the first three dimensions represent changes in the end-effector’s position, and the last dimension represents the change in the gripper’s width.

During training, we randomly select a combination of environmental features from a training set by sampling combinations from the following sets: $\text{mass} = 1.0$ and $\text{friction} \in \{0.1, 1.0, 5.0, 10.0\}$; $\text{mass} \in \{1.0, 5.0, 10.0\}$ and $\text{friction} = 0.1$. At test time, we evaluate each algorithm on all tasks from the moderate test setting, where $\text{mass} \in \{5.0, 10.0\}$ and $\text{friction} \in \{1.0, 5.0, 10.0\}$ (shown in Figure \ref{fig:benchmarks_panda}(b)), and on all tasks from the extreme test setting: $\text{mass} = 30.0$ and $\text{friction} \in \{0.1, 1.0, 5.0, 10.0, 30.0\}$; $\text{mass} \in \{1.0, 5.0, 10.0, 30.0\}$ and $\text{friction} = 30.0$ (shown in Figure \ref{fig:benchmarks_panda}(b)).

\subsection{Modified MuJoCo}

\begin{figure}[b]
    \centering
    \captionsetup{type=figure}
    \includegraphics[width=0.7\linewidth]{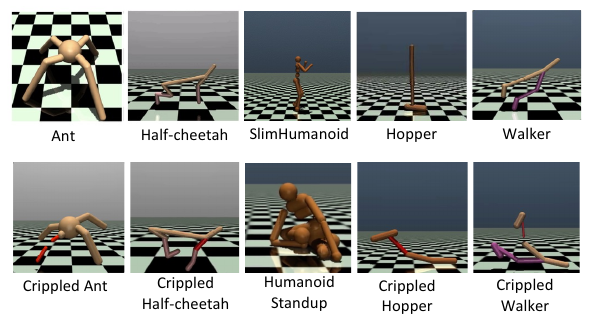}
    \caption{Ten environments in modified MuJoCo benchmark.}
\label{fig:benchmarks_mujoco}
\end{figure}
We extended the modified MuJoCo benchmark introduced in DOMINO \citep{mu2022decomposed} and CaDM \citep{lee2020context}. In our extension, four new environments were added (Walker, Crippled Hopper, Crippled Walker, HumanoidStandup), compared to the original benchmark. Additionally, in our experiments, we used a different task set design (Table \ref{tab:Training_sizes_mujoco}) than those used in the DOMINO and CaDM papers. For Hopper, Walker, Half-Cheetah, Ant, HumanoidStandup, and SlimHumanoid, we used the MuJoCo physics engine environments and implemented settings from \cite{nagabandi2019learning} and \cite{seo2020trajectory}, scaling the mass of every rigid link by a scale factor $m$ and the damping of every joint by a scale factor $d$. For Crippled Ant, Crippled Hopper, Crippled Walker, and Crippled Half-Cheetah, we used the implementation available from \cite{seo2020trajectory}, scaled the mass of each rigid link by a factor of $m$, scaled the damping of each joint by a factor of $d$, and randomly selected joints to be uncontrollable (i.e., masking the corresponding actions with 0). Generalisation performance is measured in two different regimes: moderate and extreme, where the moderate regime draws environmental features from a range closer to the training range compared to the extreme regime. Our settings for training, extreme, and moderate test tasks are provided in Table \ref{tab:Training_sizes_mujoco}.

\begin{sidewaystable}
	\centering 
	\fontsize{10}{14}\selectfont  
	\caption{Environmental features used for MuJoCo benchmark.}
 \scalebox{0.8}{
	\begin{tabular}{ccccc}
		\toprule
		\toprule
		& Training & Test (Moderate) & Test (Extrame) & Episode Length \cr
		\cmidrule(lr){1-5}
		Half-cheetah & \makecell[c]{$m\in \{0.75, 0.85, 1.0, 1.15, 1.25\}$ \\ $d \in \{0.75, 0.85, 1.0, 1.15, 1.25\}$}
                &  \makecell[c]{$m\in \{0.40, 0.50, 1.50, 1.60\}$ \\ $d \in \{0.40, 0.50, 1.50, 1.60\}$} 
                & \makecell[c]{$m\in \{0.20, 0.40, 1.60, 1.80, 4.00\}$ \\ $d \in \{0.20, 0.40, 1.60, 1.80, 4.00\}$} & 1000 \cr
            \cmidrule(lr){1-5} 
          Ant      &  \makecell[c]{$m\in \{0.40, 0.50, 1.50, 1.60\}$ \\ $d \in \{0.40, 0.50, 1.50, 1.60\}$} 
                & \makecell[c]{$m\in \{0.20, 0.40, 1.60, 1.80\}$ \\ $d \in \{0.20, 0.40, 1.60, 1.80\}$} & \makecell[c]{$m\in \{0.20, 0.40, 1.60, 1.80, 4.00\}$ \\ $d \in \{0.20, 0.40, 1.60, 1.80, 4.00\}$} & 1000 \cr
            \cmidrule(lr){1-5}
		Hopper & \makecell[c]{$m\in \{0.75, 1.0, 1.25\}$ \\ $d \in \{0.75, 1.0, 1.25\}$}
                &  \makecell[c]{$m\in \{0.40, 0.50, 1.50, 1.60\}$ \\ $d \in \{0.40, 0.50, 1.50, 1.60\}$} 
                & \makecell[c]{$m\in \{0.20, 0.40, 1.60, 1.80, 4.00\}$ \\ $d \in \{0.20, 0.40, 1.60, 1.80, 4.00\}$} & 1000 \cr
            \cmidrule(lr){1-5}

        Crippled Hopper & \makecell[c]{$m\in \{0.75, 1.0, 1.25\}$ \\ $d \in \{0.75, 1.0, 1.25\}$}
                &  \makecell[c]{$m\in \{0.40, 0.50, 1.50, 1.60\}$ \\ $d \in \{0.40, 0.50, 1.50, 1.60\}$} 
                & \makecell[c]{$m\in \{0.20, 0.40, 1.60, 1.80, 4.0\}$ \\ $d \in \{0.20, 0.40, 1.60, 1.80, 4.0\}$ \\ Crippled Legs $R_1 \in \{0,1,2\}$} & 1000 \cr
            \cmidrule(lr){1-5}
            
		SlimHumanoid & \makecell[c]{$m\in \{0.80, 0.90, 1.0, 1.15, 1.25\}$ \\ $d \in \{0.80, 0.90, 1.0, 1.15, 1.25\}$}
                &  \makecell[c]{$m\in \{0.60, 0.70, 1.50, 1.60\}$ \\ $d \in \{0.60, 0.70, 1.50, 1.60\}$} 
                & \makecell[c]{$m\in \{0.40, 0.50, 1.70, 1.80\}$ \\ $d \in \{0.40, 0.50, 1.70, 1.80\}$} & 1000 \cr
            \cmidrule(lr){1-5}
            
        HumanoidStandup & \makecell[c]{$m\in \{0.80, 0.90, 1.0, 1.15, 1.25\}$ \\ $d \in \{0.80, 0.90, 1.0, 1.15, 1.25\}$}
                &  \makecell[c]{$m\in \{0.60, 0.70, 1.50, 1.60\}$ \\ $d \in \{0.60, 0.70, 1.50, 1.60\}$} 
                & \makecell[c]{$m\in \{0.40, 0.50, 1.70, 1.80, 4.00\}$ \\ $d \in \{0.40, 0.50, 1.70, 1.80, 4.00\}$} & 1000 \cr
            \cmidrule(lr){1-5}
            
        Walker & \makecell[c]{$m\in \{0.75, 1.0, 1.25\}$ \\ $d \in \{0.75, 1.0, 1.25\}$}
                &  \makecell[c]{$m\in \{0.40, 0.50, 1.50, 1.60\}$ \\ $d \in \{0.40, 0.50, 1.50, 1.60\}$} 
                & \makecell[c]{$m\in \{0.20, 0.40, 1.60, 1.80, 4.00\}$ \\ $d \in \{0.20, 0.40, 1.60, 1.80, 4.00\}$} & 2000 \cr
            \cmidrule(lr){1-5}
            
       Crippled Walker & \makecell[c]{$m\in \{0.75, 1.0, 1.25\}$ \\ $d \in \{0.75, 1.0, 1.25\}$ \\ Crippled Joints (right leg) $= \{0,1,2\}$}
                &  \makecell[c]{$m\in \{0.40, 0.50, 1.50, 1.60\}$ \\ $d \in \{0.40, 0.50, 1.50, 1.60\}$ \\ Crippled Joints (right leg) $\in$ $\{0,1,2\}$} 
                & \makecell[c]{$m\in \{0.20, 0.40, 1.60, 1.80, 4.00\}$ \\ $d \in \{0.20, 0.40, 1.60, 1.80, 4.00\}$ \\ Crippled Joints (left leg) $\in$ $\{3,4,5\}$} & 2000 \cr
            \cmidrule(lr){1-5}
            
  	Crippled Ant & \makecell[c]{$m\in \{0.75, 0.85, 1.0, 1.15, 1.25\}$ \\ $d \in \{0.75, 0.85, 1.0, 1.15, 1.25\}$ \\ Crippled Legs $R_1 = \varnothing $ or $R_1 \in \{0,1,2,3\}$}
                &  \makecell[c]{$m\in \{0.40, 0.50, 1.50, 1.60\}$ \\ $d \in \{0.40, 0.50, 1.50, 1.60\}$  \\ Crippled Legs $R_1 \in \{0,1,2,3\}$}  
                & \makecell[c]{$m\in \{0.20, 0.40, 1.60, 1.80\}$ \\ $d \in \{0.20, 0.40, 1.60, 1.80\}$ \\ Crippled Legs $\{R_1,R_2\}\in \{0,1,2,3,4,5\}$ $(R_1\neq R_2)$ } & 2000 \cr
            \cmidrule(lr){1-5}
        
    	Crippled Half-cheetah & \makecell[c]{$m\in \{0.75, 0.85, 1.0, 1.15, 1.25\}$ \\ $d \in \{0.75, 0.85, 1.0, 1.15, 1.25\}$ \\ Crippled Joints (front leg) $R_1 \in$ $\{3,4,5\}$}
                &  \makecell[c]{$m\in \{0.40, 0.50, 1.50, 1.60\}$ \\ $d \in \{0.40, 0.50, 1.50, 1.60\}$  \\ Crippled Joints (back leg) $R_1 \in \{0,1,2\}$ } 
                & \makecell[c]{$m\in \{0.20, 0.40, 1.60, 1.80\}$ \\ $d \in \{0.20, 0.40, 1.60, 1.80\}$ \\ Crippled Joints $\{R_1,R_2\}\in \{0,1,2,3,4,5\}$ $(R_1\neq R_2)$ } & 2000 \cr
		\bottomrule
		\bottomrule
	\end{tabular}}\vspace{0cm}
	\label{tab:Training_sizes_mujoco}
\end{sidewaystable}

\subsection{Implementation details} \label{appendix:Implementation_details}

In this section, we provide the implementation details for SaMI. We present the pseudo-code for using SaNCE during meta-training and meta-testing in Algorithms \ref{algo:meta_training} and \ref{algo:meta_testing}. Our codebase is built on top of the publicly released implementation of Stable Baselines3 by \cite{stable-baselines3} and the implementation of InfoNCE by \cite{oord2019representation}. A public, open-source implementation of SaMI is available at https://github.com/uoe-agents/SaMI.

\textbf{Base algorithm.} We use SAC \citep{haarnoja2018soft} for the downstream evaluation of the learned context embedding. SAC is an off-policy actor-critic method that leverages the maximum entropy framework for soft policy iteration. At each iteration, SAC performs soft policy evaluation and improvement steps. We use the same SAC implementation across all baselines and other methods. During the meta-training phase, we trained agents for 1.6 million timesteps in each environment on the Panda-gym and MuJoCo benchmarks. For meta-testing, we evaluated 100 episodes in each environment, with tasks randomly sampled from the moderate and extreme task sets.

\begin{table}
	\centering
	\fontsize{8}{11}\selectfont  
	\caption{Hyperparameters used in the Panda-gym and MuJoCo benchmarks. Most hyperparameter values remain unchanged across tasks, except for the contrastive batch size and the SaNCE loss coefficient.}
	\begin{tabular}{cc}
		\toprule
		 Hyperparameter & Value  \cr
		\cmidrule(lr){1-2}
            Replay buffer size & 100,000  \cr
            Contrastive batch size  & MuJoCo 12, Panda-gym 256 \cr
            SaNCE loss coefficient $\alpha$ & MuJoCo 1.0, Panda-gym 0.01 \cr
            Context embedding dimension & 6 \cr
            Hidden state dimension & 128 \cr
            Learning rate (actor, critic and encoder) & 1e-3 \cr
            Training frequency (actor, critic and encoder) & 128 \cr
            Gradient steps & 16 \cr
            Momentum context encoder $\psi^*$ soft-update rate & 0.05 \cr
            SAC target soft-update rate & critic 0.01, actor 0.05 \cr
            SAC batch size & 256 \cr
            Discount factor & 0.99 \cr
            Optimizer & Adam \cr
		\bottomrule
	\end{tabular}\vspace{0cm}
	\label{tab:hyper-parameter}
\end{table}

\textbf{Encoder architecture.} In our method, the context encoder $\psi$ is modelled as a Long Short-Term Memory (LSTM) network that produces a 128-dimensional hidden state vector, which is subsequently processed through a single-layer feed-forward network to generate a 6-dimensional context embedding. We aim for the agent to complete the three steps of "explore effectively, infer, adapt" within an episode. Therefore, we initialise the hidden state and cell state of the LSTM to zero at the start of each episode. Both the actor and critic use the same context encoder to embed trajectories. For contrastive learning, SaNCE utilises a momentum encoder $\psi^*$ to generate positive and negative context embeddings \citep{laskin2020curl,he2020momentum}. Formally, denoting the parameters of $\psi$ as $\theta_{\psi}$ and those of $\psi^*$ as $\theta_{\psi^*}$, we update $\theta_{\psi^*}$ as follows:
\begin{equation}
    \theta_{\psi^*} \leftarrow m \cdot \theta_{\psi} + (1-m) \cdot \theta_{\psi^*}.
    \label{eq:soft_update}
\end{equation}
Here $m \in [0,1)$ is a soft-update rate. Only the parameters $\theta_{\psi}$ are updated by back-propagation. The momentum update in Equation \eqref{eq:soft_update} makes $\theta_{\psi^*}$ evolve more smoothly by having them slowly track the $\theta_{\psi}$ with $m \ll 1$ (e.g., $m=0.05$ in this research). This means that the target values are constrained to change slowly, greatly improving the stability of learning.

\textbf{Hyperparameters.} A full list of hyperparameters is displayed in Table \ref{tab:hyper-parameter}.

\textbf{Hardware.} For each experiment run we use a single NVIDIA Volta V100 GPU with 32GB memory and a single CPU.

\begin{algorithm}
    \caption{SaNCE Meta-training}
    \begin{algorithmic}[1]
        \REQUIRE Batch of training tasks $\{e_n\}_{n=1,...,N}$ from $\xi_{train}(e)$, soft-update rate $m$;
        \STATE Initialize RL replay buffer $\mathcal{B}_{RL}$, encoder replay buffer $\mathcal{B}_{enc}$;
        \STATE Initialize parameters $\psi$ for context encoder, $\psi^*$ for momentum context encoder and $\phi$ for the off-policy SAC;
        \WHILE {not done}
            \FOR{each task $e_n$}
                \FOR{Roll-out time steps}
                \FOR{time step $t$ < maximum episode length $T$}
                    \STATE Update context embedding $c_n \sim \psi(c_n|\tau_{c_n,0:t})$
                    \STATE Roll-out policy $\pi_{c_n}(a_t|s_t,c_n)$ and accumulate transition $(s_t,a_t,r_t,s_{t+1})$;
                \ENDFOR
                \STATE Add trajectory $\tau_{c_n} = \{s_0, a_0, r_0, s_1, r_1, ..., s_{T}, a_{T},r_{T}\}$ to replay buffer $\mathcal{B}_{RL}^{n}$ and $\mathcal{B}_{enc}^{n}$;
                \ENDFOR
            \ENDFOR
            \FOR{each training step}
                \FOR{each task $e_n$}
                \STATE Sample RL batch $\{\tau_{c_n}\} \sim \mathcal{B}_{RL}^n$;
                \STATE Sample a positive sample $\tau_{c_n}^+$ for generating query with highest return, positive samples $\{\tau_{c_n}^-\}$ and negative samples $\{\tau_{c_n}^-\}$ for encoding positive and negative embeddings;
                \STATE Update $\phi$ with RL loss $\mathcal{L}_{\text{RL}}$;
                \STATE Update $\psi$ with SaNCE loss $\mathcal{L}_{\text{SaNCE}}$ and RL loss $\mathcal{L}_{\text{RL}}$;
                \STATE $\theta_{\psi^*} \leftarrow m \cdot \theta_{\psi} + (1-m) \cdot \theta_{\psi^*}$;
                \ENDFOR
            \ENDFOR
        \ENDWHILE
    \end{algorithmic}
    \label{algo:meta_training}
\end{algorithm}

\begin{algorithm}
    \caption{SaNCE Meta-testing}
    \begin{algorithmic}[1]
        \REQUIRE Batch of training tasks $\{e_n\}_{n=1,...,N}$ from $\xi_{test}(e)$;
        \WHILE {not done}
            \FOR{each task $e_n$}
                \FOR{each episode}
                \FOR{time step $t$ < maximum episode length $T$}
                    \STATE Update context embedding $c_n \sim \psi(c_n|\tau_{c_n,0:t})$
                    \STATE Roll-out policy $\pi_{c_n}(a_t|s_t,c_n)$ and accumulate transition $(s_t,a_t,r_t,s_{t+1})$;
                \ENDFOR
                \ENDFOR
            \ENDFOR
        \ENDWHILE
    \end{algorithmic}
    \label{algo:meta_testing}
\end{algorithm}

\clearpage
\section{Additional results}

\subsection{Balance contrastive and RL updates: loss coefficient $\alpha$}
\begin{figure}[ht]
    \centering
    \includegraphics[width=1.0\linewidth]{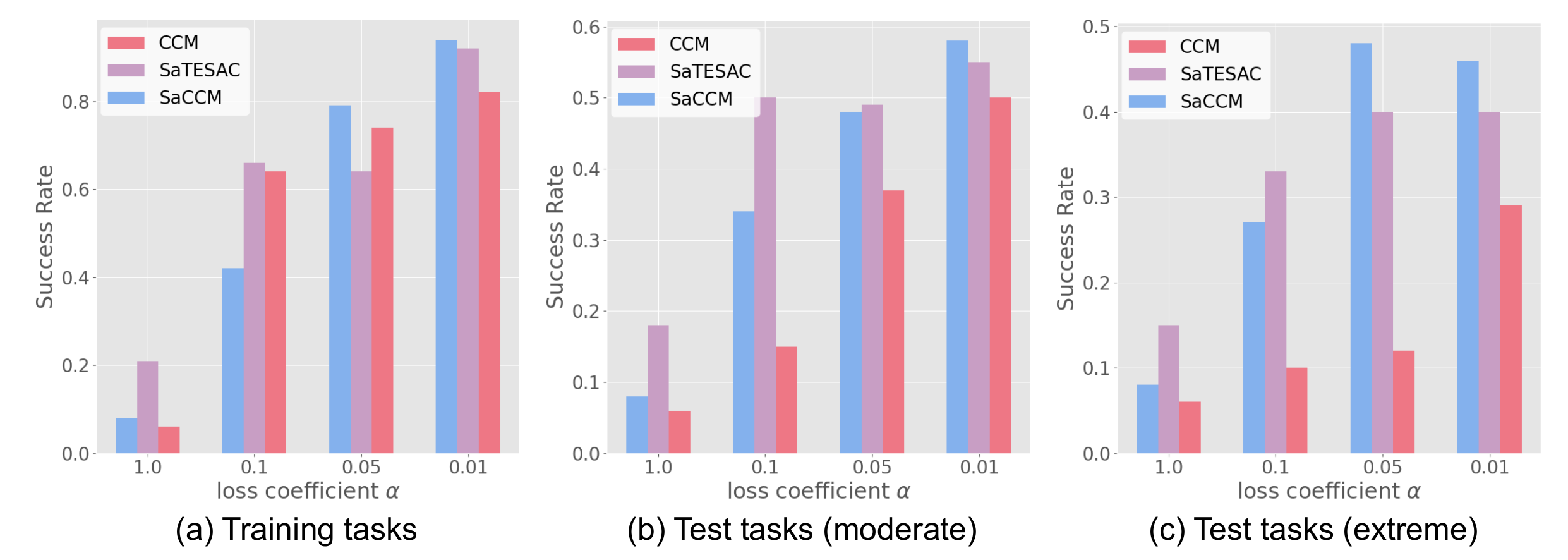}
    \captionof{figure}{Loss coefficient $\alpha$ analysis of Panda-gym benchmark in training and test (moderate and extreme) tasks.}
    \label{fig:loss_coef_panda}
\end{figure}

\begin{figure}[ht]
    \centering
    \includegraphics[width=1.0\linewidth]{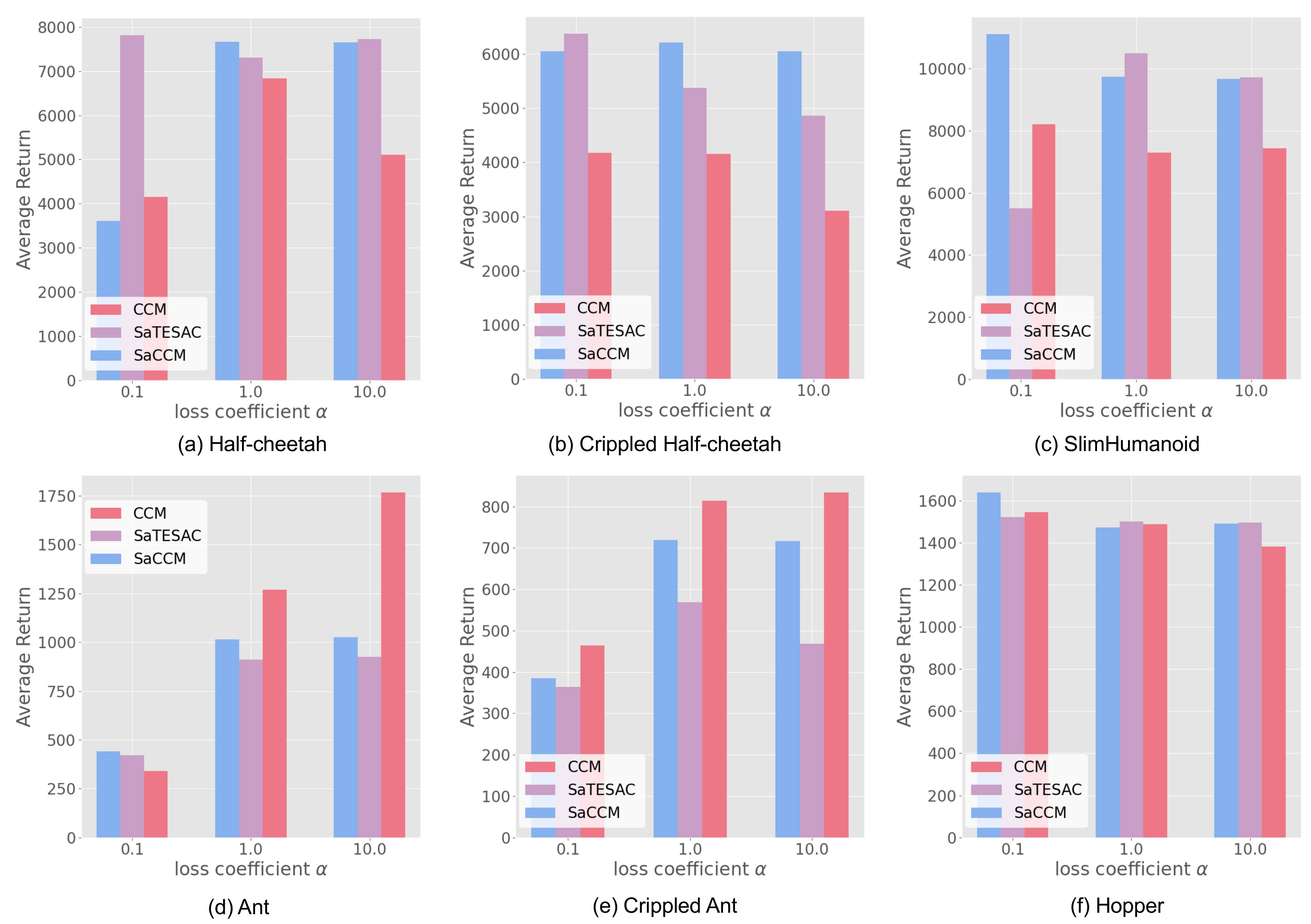}
    \captionof{figure}{Loss coefficient $\alpha$ analysis of MuJoCo benchmark in training tasks.}
    \label{fig:loss_coef_mujoco}
\end{figure}

While past work has optimised hyperparameters to balance the contrastive loss coefficient $\alpha$ relative to the RL objective \citep{jaderberg2016reinforcement,bachman2019learning}, we use both the contrastive and RL objectives with equal weight, setting $\alpha = 1.0$ for the MuJoCo benchmark and $\alpha = 0.01$ for the Panda-gym benchmark. Additionally, we analyse the effect of the loss coefficient $\alpha$ for CCM, SaTESAC, and SaCCM in the MuJoCo (Figure \ref{fig:loss_coef_mujoco}) and Panda-gym (Figure \ref{fig:loss_coef_panda}) benchmarks.

\clearpage
\subsection{Result of $\log$-$K$ curse analysis} \label{appx:sample_space}
\subsubsection{Buffer size}
\label{appx:sample_space_buffer_size}
\begin{figure}[ht]
\centering
\includegraphics[width=0.8\linewidth]{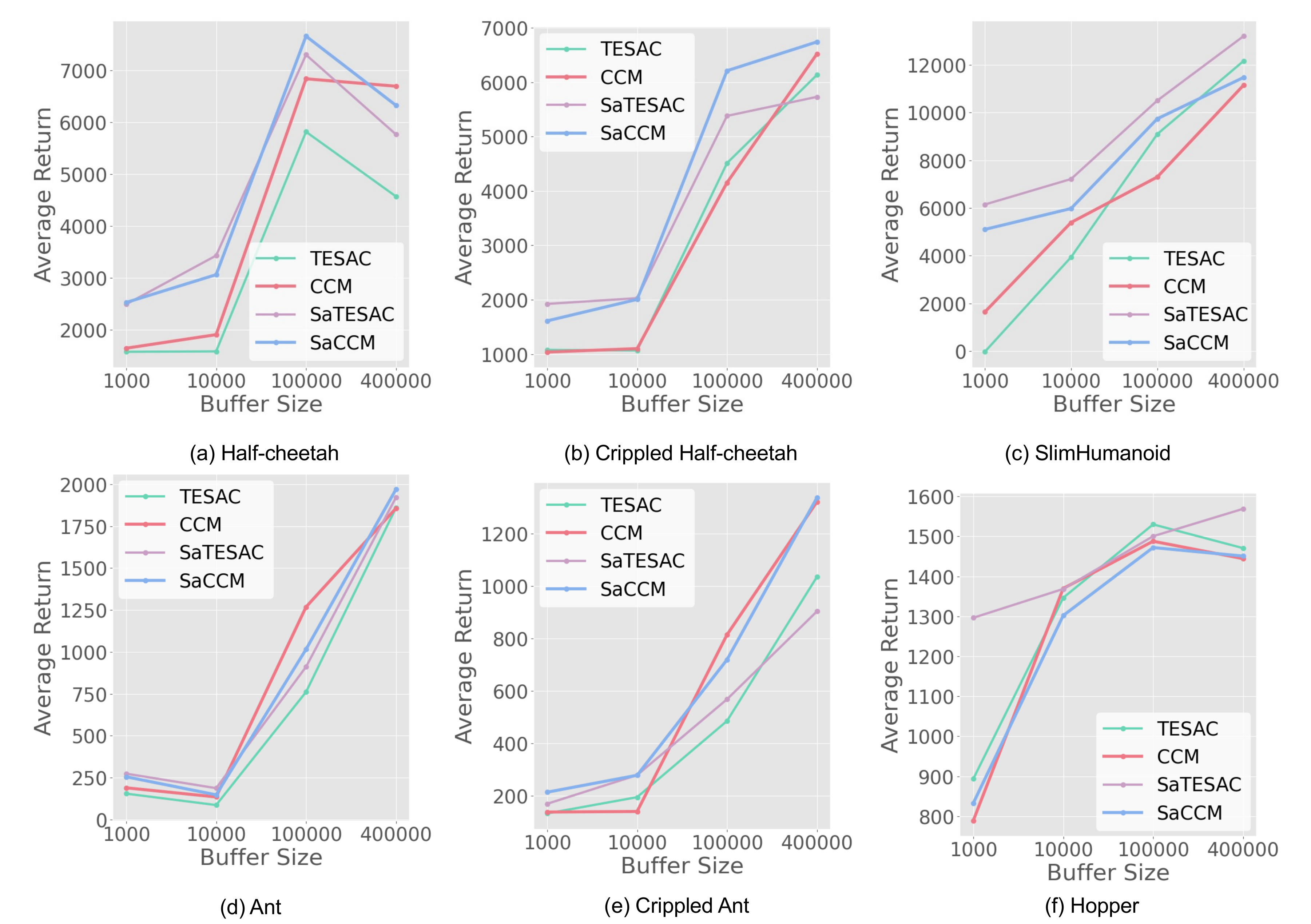}
\caption{Comparison of different buffer sizes in the MuJoCo benchmark on training tasks (averaged over 5 seeds). Buffer sizes are 400,000, 100,000, 10,000, and 1,000.}
\end{figure}

\subsubsection{Contrastive batch size}
\label{appx:sample_space_batch_size}
\begin{figure}[ht]
\centering
\includegraphics[width=0.8\linewidth]{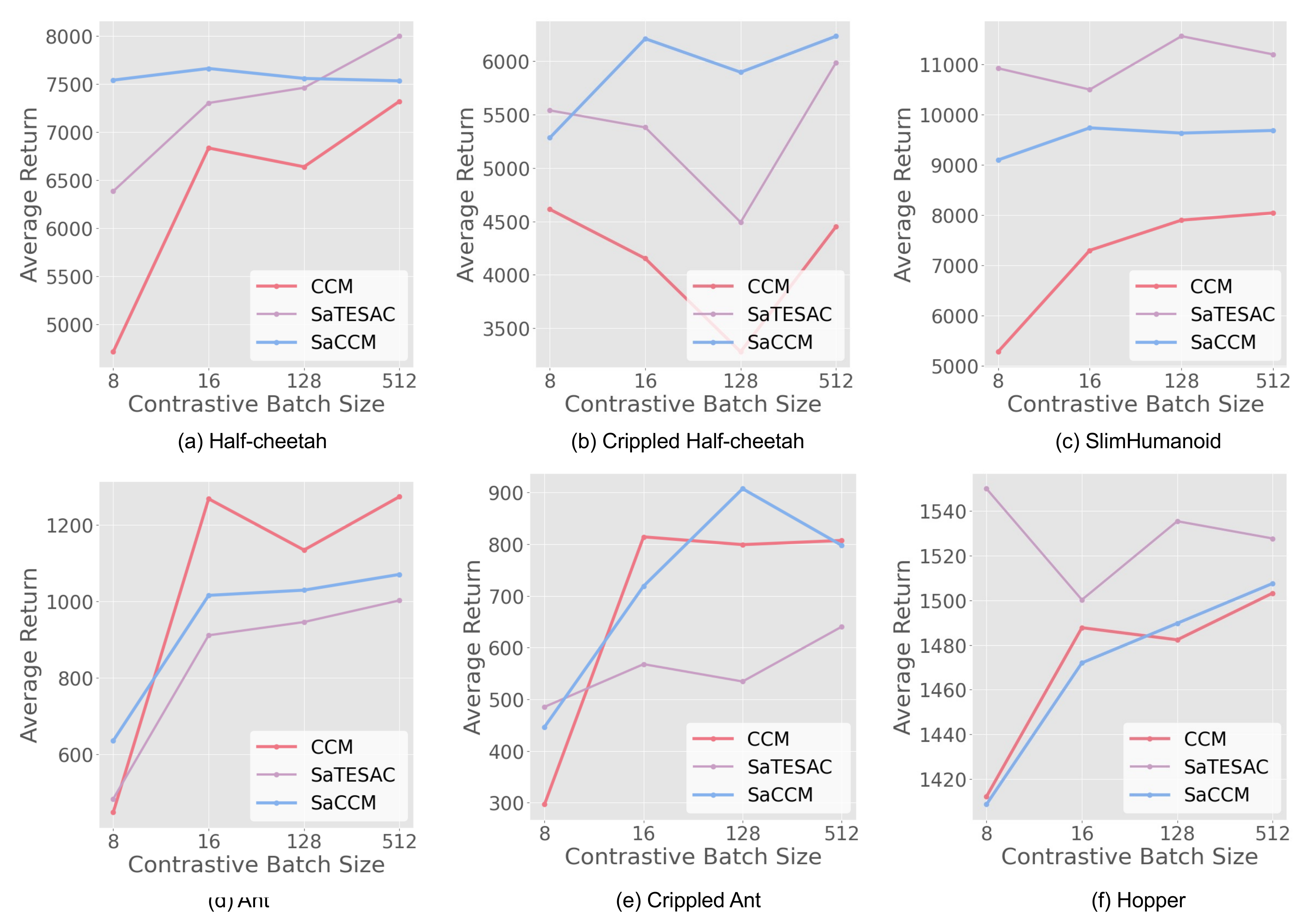}
\caption{Comparison of different contrastive batch sizes in the MuJoCo benchmark on training tasks (averaged over 5 seeds). Contrastive batch sizes are 512, 128, 16, and 8.}
\end{figure}

\clearpage
\section{Further skill analysis} \label{appx:visualisation}
\subsection{Panda-gym} \label{appx:visualisation_panda}
\subsubsection{Visualisation of context embedding}
\begin{figure}[ht]
    \centering
    \includegraphics[width=0.65\linewidth]{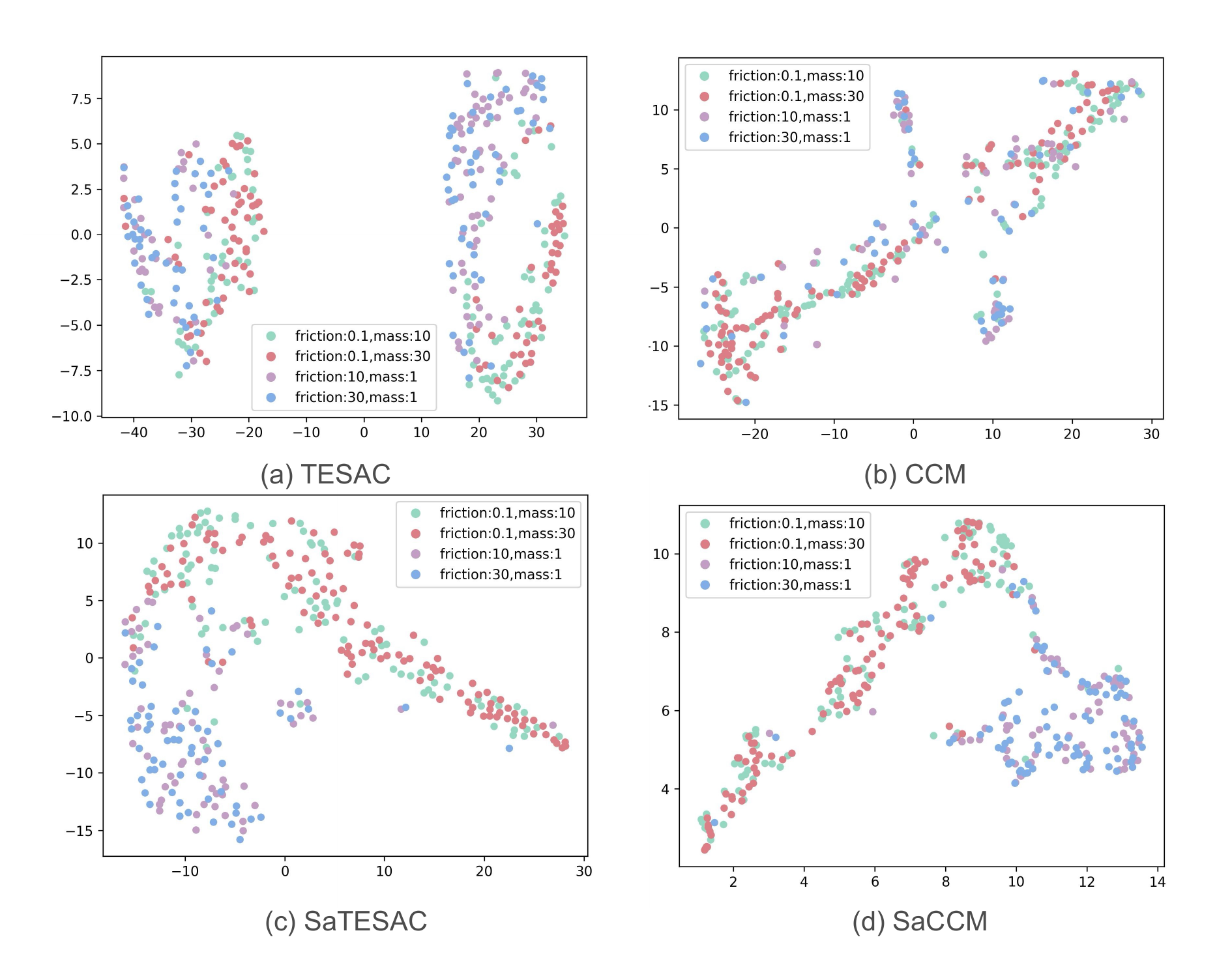}
    \captionof{figure}{t-SNE visualisation of context embeddings extracted from trajectories collected in the Panda-gym environments.}
    \label{fig:vis_tnse_panda}
\end{figure}

We visualise the context embedding using UMAP \citep{mcinnes2020umapuniformmanifoldapproximation} (Figure \ref{fig:vis_umap_panda}) and t-SNE \citep{van2008visualizing} (Figure \ref{fig:vis_tnse_panda}). When the mass of the cube is high (30 kg and 10 kg), the agent learned the Push skill (indicated by the yellow bounding box in Figure \ref{tab:panda-gym-average-return}(a)), whereas with lower masses, the agent learned the Pick\&Place skill. However, as shown in Figure \ref{fig:vis_tnse_panda}(b), CCM did not display clear skill grouping. This indicates that SaMI extracts high-quality skill-related information from the trajectories and enables agents to autonomously discover a diverse range of skills for handling multiple tasks.
\begin{figure}[ht]
    \centering
    \includegraphics[width=0.65\linewidth]{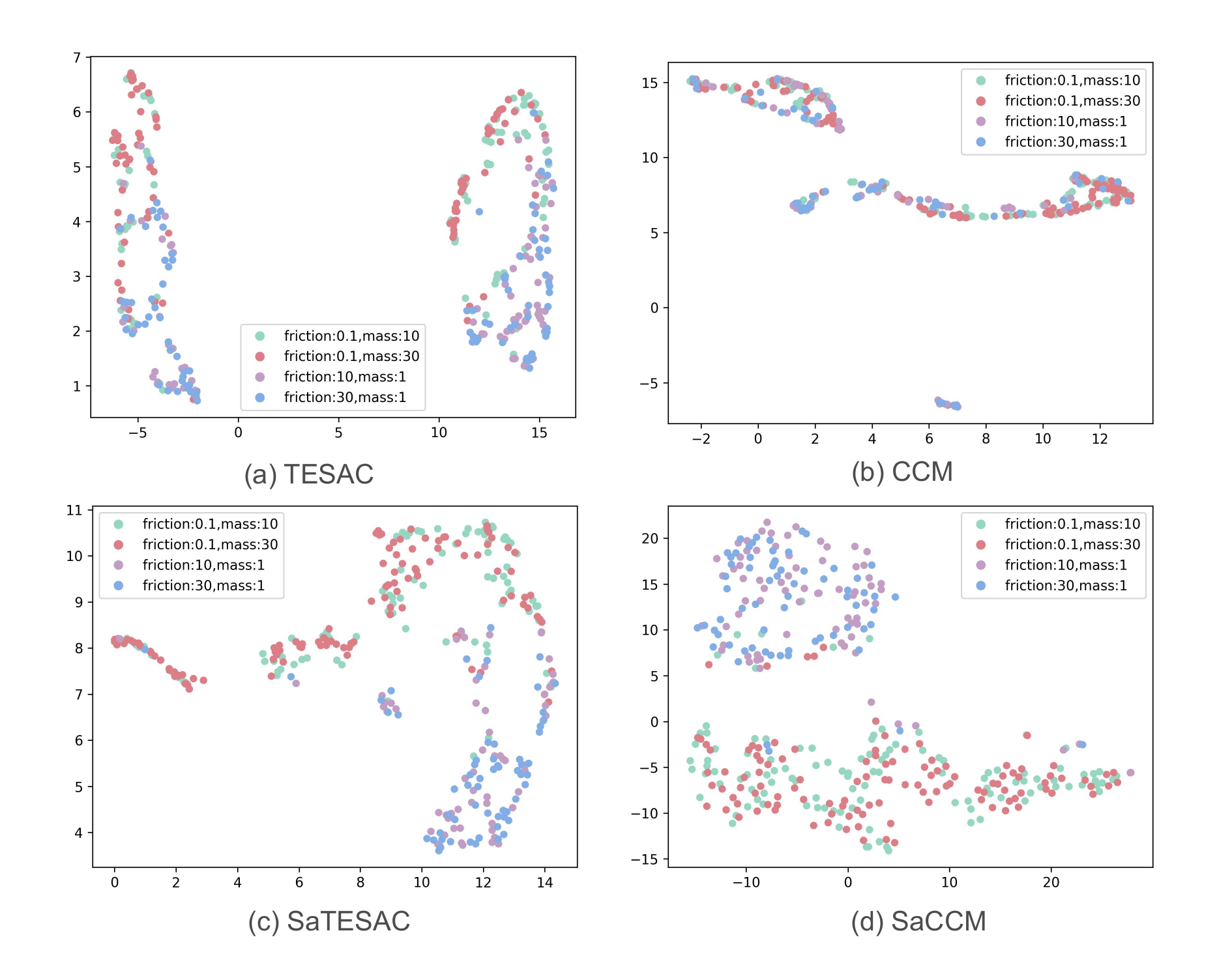}
    \captionof{figure}{UMAP visualisation of context embeddings extracted from trajectories collected in the Panda-gym environments.}
    \label{fig:vis_u ma p_panda}
\end{figure}

\clearpage
\subsubsection{Heatmap of Panda-gym benchmark} \label{appx:heatmap}
\begin{figure}[ht]
    \centering
    \includegraphics[width=0.9\linewidth]{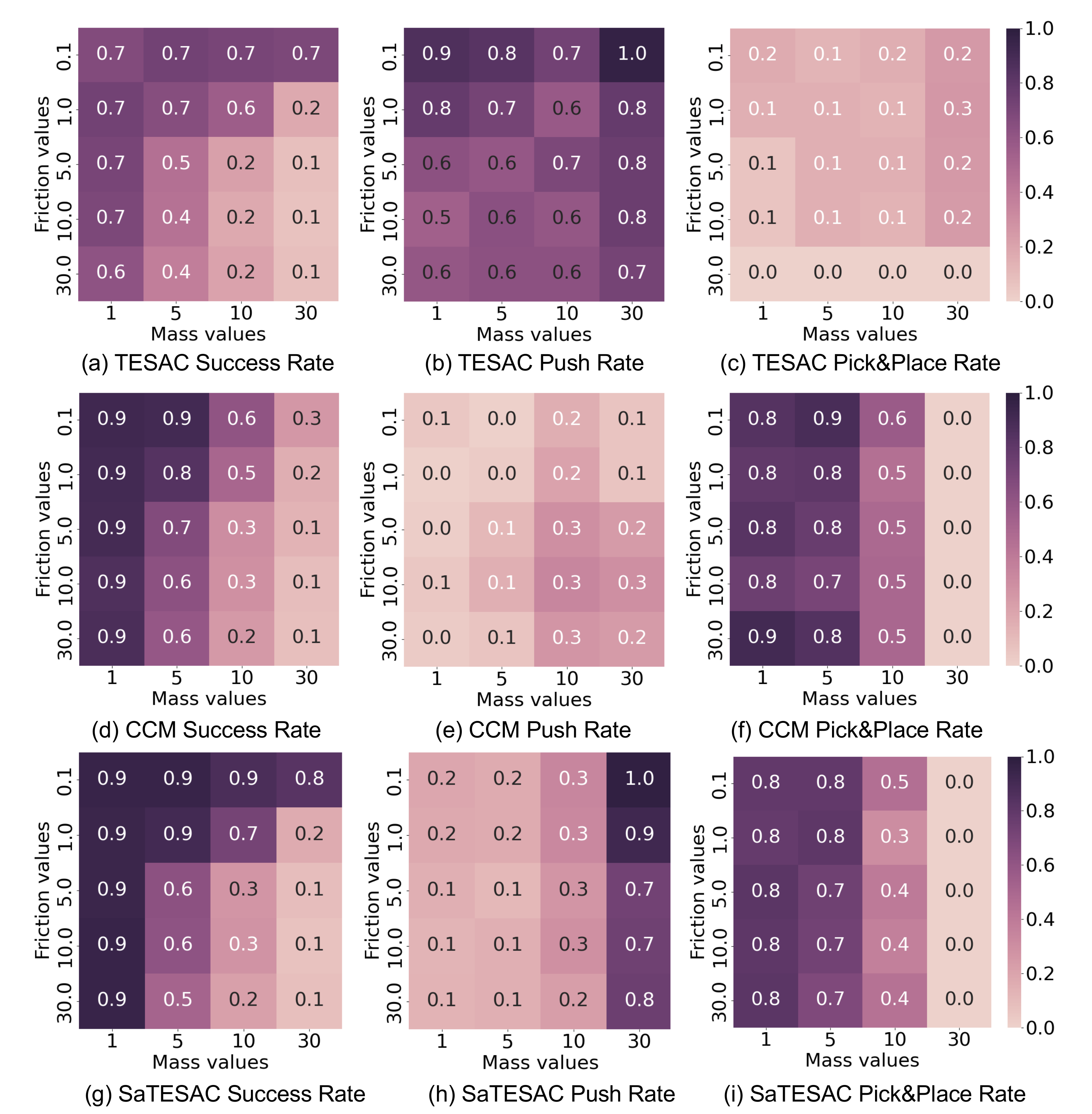}
    \captionof{figure}{Heatmap of success rates and learned skills of SaCCM. For each grid, the $(i, j)^{\text{th}}$ location shows the probability of the skills executed over 100 evaluations with $(\text{mass}=i, \text{friction}=j)$. }
    \label{fig:heatmap_all_methods}
\end{figure}

This section presents the heatmap results and further analysis of TESAC, CCM, and SaTESAC on the Panda-gym benchmark. From the heatmap results of SaTESAC and SaCCM (Figure \ref{fig:heatmap_skills}), we observe that, with higher cube masses (30 and 10 kg), the agent executed the Push skill (indicated by the clustered points within the yellow bounding box in Figure \ref{tab:panda-gym-average-return}(a)). At lower masses, the agent executed the Pick\&Place skill.

In contrast, as shown in Figures \ref{fig:heatmap_all_methods}(e-f), CCM predominantly learned the Pick\&Place skill, resulting in a drop in success rates for tasks with $mass=30$, as the agent could not lift the cube off the table using the Pick\&Place skill, as depicted in Figure \ref{fig:heatmap_all_methods}(d). The visualisation of the context embedding (Figure \ref{fig:vis_tnse_panda}) did not reveal clear skill grouping across different tasks.

Finally, TESAC primarily mastered the Push skill. The Push skill is relatively simpler to learn than the Pick\&Place skill, as it does not require the agent to manipulate its fingers to pick up cubes. Consequently, TESAC’s success rate notably decreased in environments with higher friction.

\clearpage
\subsection{MuJoCo}  \label{appx:visualisation_mujoco}

SaMI enables RL agents to be versatile and embody multiple skills. Additionally, video demos (available at https://github.com/uoe-agents/SaMI) provide a better demonstration of the different skills.

\begin{figure}[ht]
    \centering
    \includegraphics[width=1.0\linewidth]{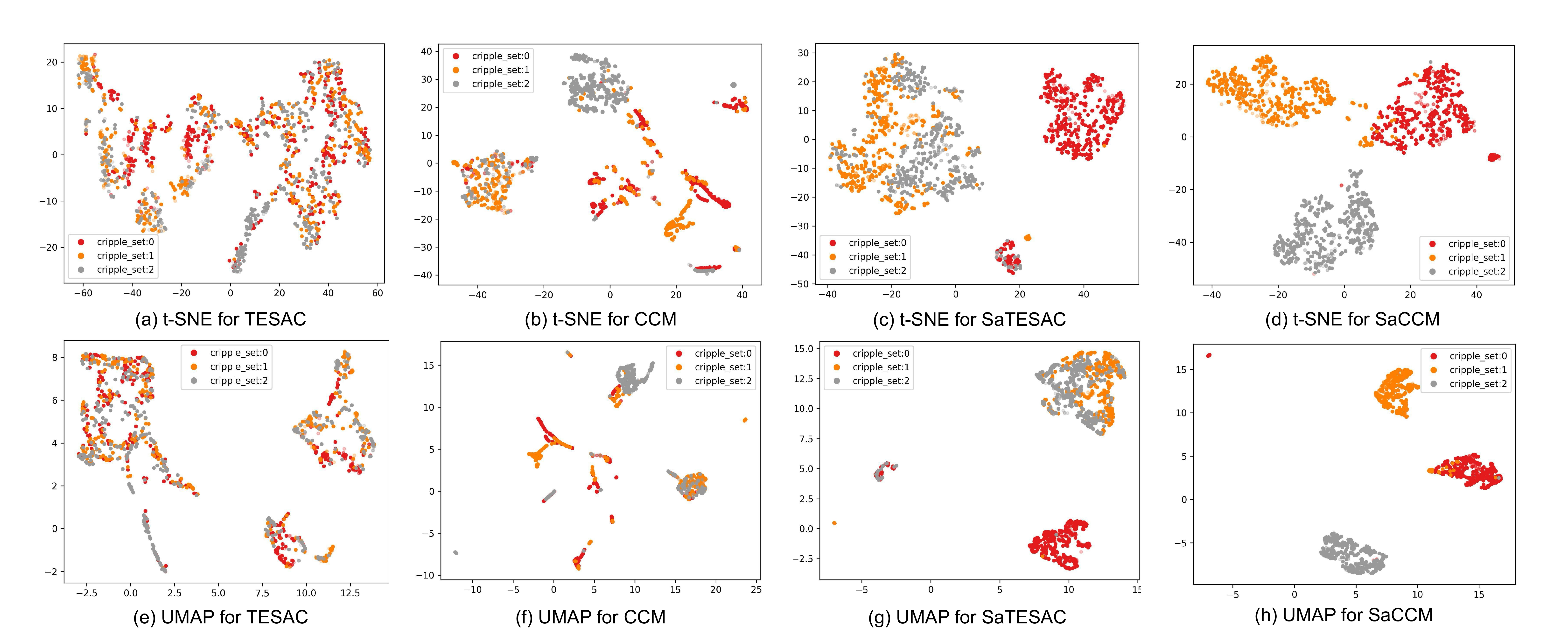}
    \captionof{figure}{t-SNE and UMAP visualisations of context embeddings extracted from trajectories collected in the Crippled Half-Cheetah environment. "cripple\_set" refers to the index of the crippled joint. The figure shows the context embeddings of tasks in three moderate test settings. Combined with the video demos\textsuperscript{\ref{github}} for skill analysis, the Crippled Half-Cheetah robot executed three distinct forward running skills.}
\label{fig:context_embedding_CrippleHalfCheetahEnv}
\end{figure}
\begin{figure}[ht]
    \centering
    \includegraphics[width=1.0\linewidth]{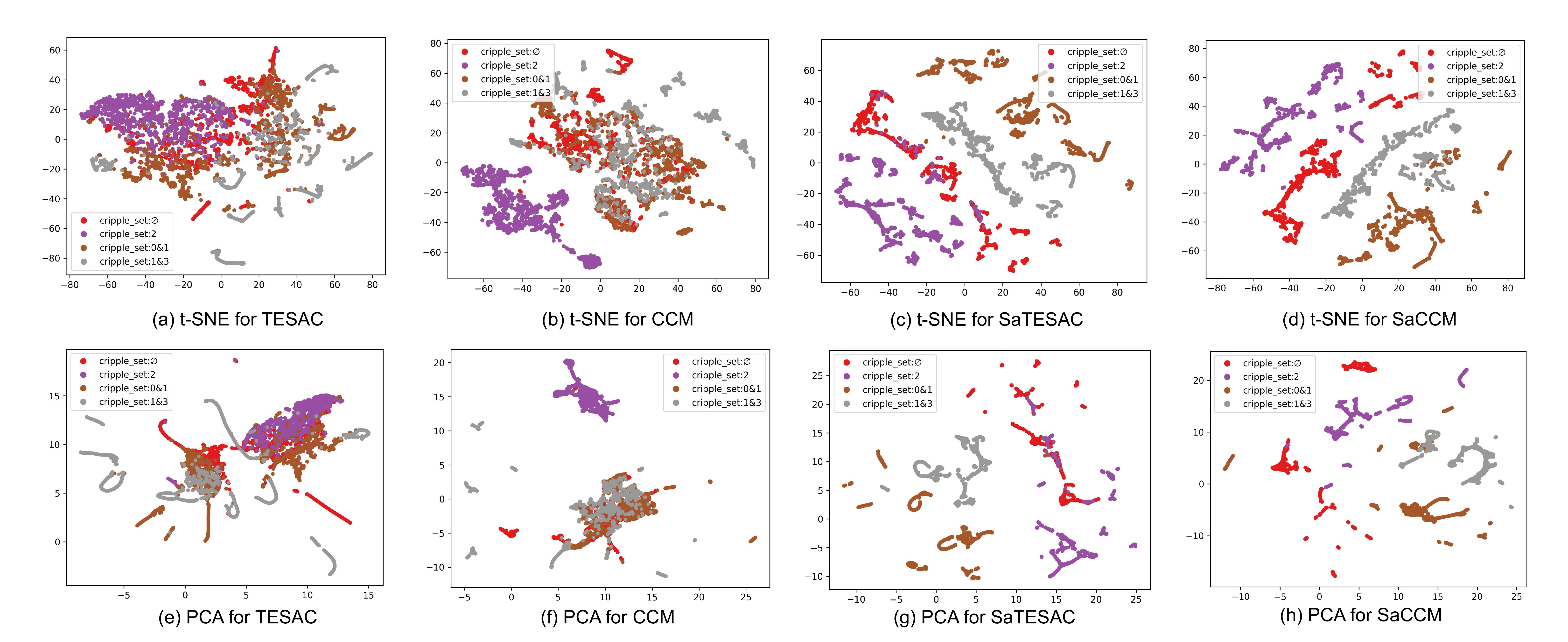}
    \captionof{figure}{t-SNE and UMAP visualisations of context embeddings extracted from trajectories collected in the Crippled Ant environment. "cripple\_set" refers to the index of the crippled joint. When 3 or 4 legs are available, the Ant robot (trained with SaCCM) rolls to adapt to varying mass and damping. However, with only 2 adjacent legs during zero-shot generalisation, it switches to walking. If 2 opposite legs are available, the Ant can still roll but eventually tips over. Please refer to the video demos\textsuperscript{\ref{github}}.}
\label{fig:context_embedding_CrippleAntEnv}
\end{figure}
\begin{figure}[ht]
    \centering
    \includegraphics[width=1.0\linewidth]{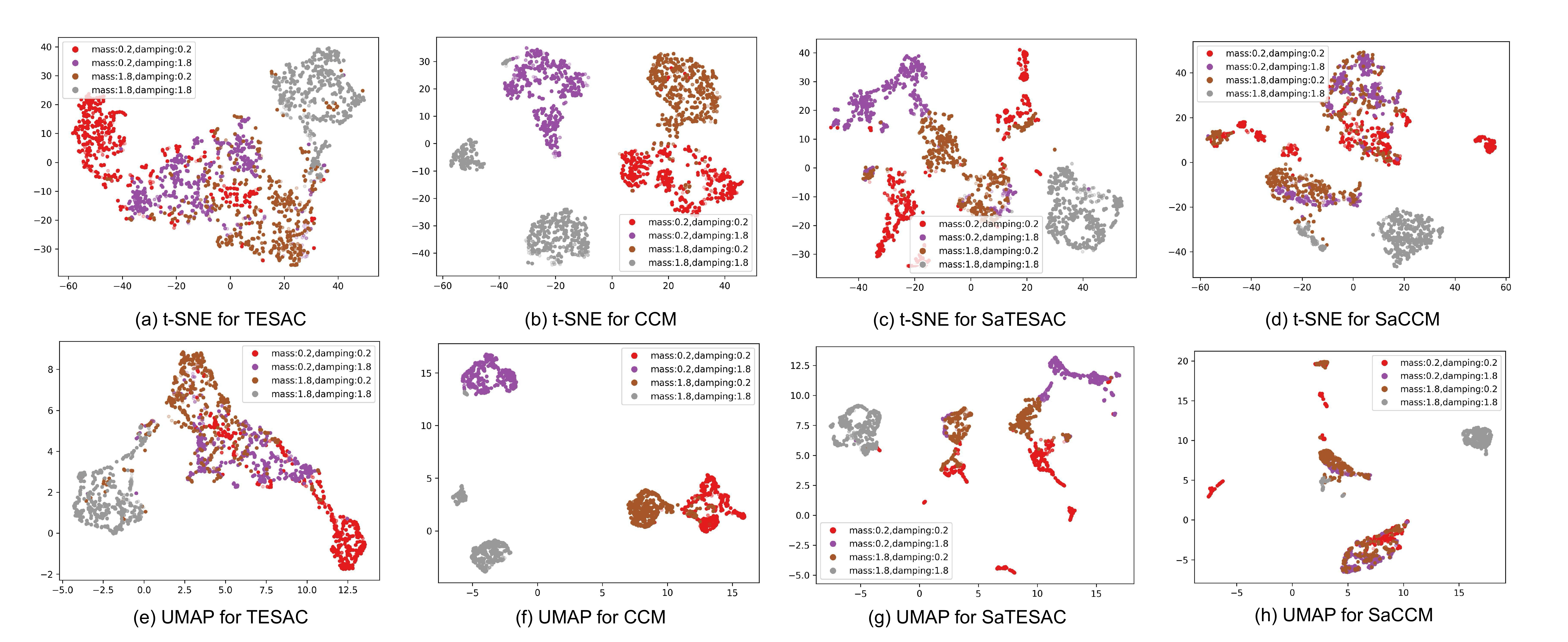}
    \captionof{figure}{t-SNE and UMAP visualisations of context embeddings extracted from trajectories collected in the Half-Cheetah environment. During zero-shot generalisation, SaCCM demonstrates different skills for running forwards at various speeds, as well as skills for doing flips and faceplanting.}
\label{fig:context_embedding_Half-cheetah}
\end{figure}
\begin{figure}[ht]
    \centering
    \includegraphics[width=1.0\linewidth]{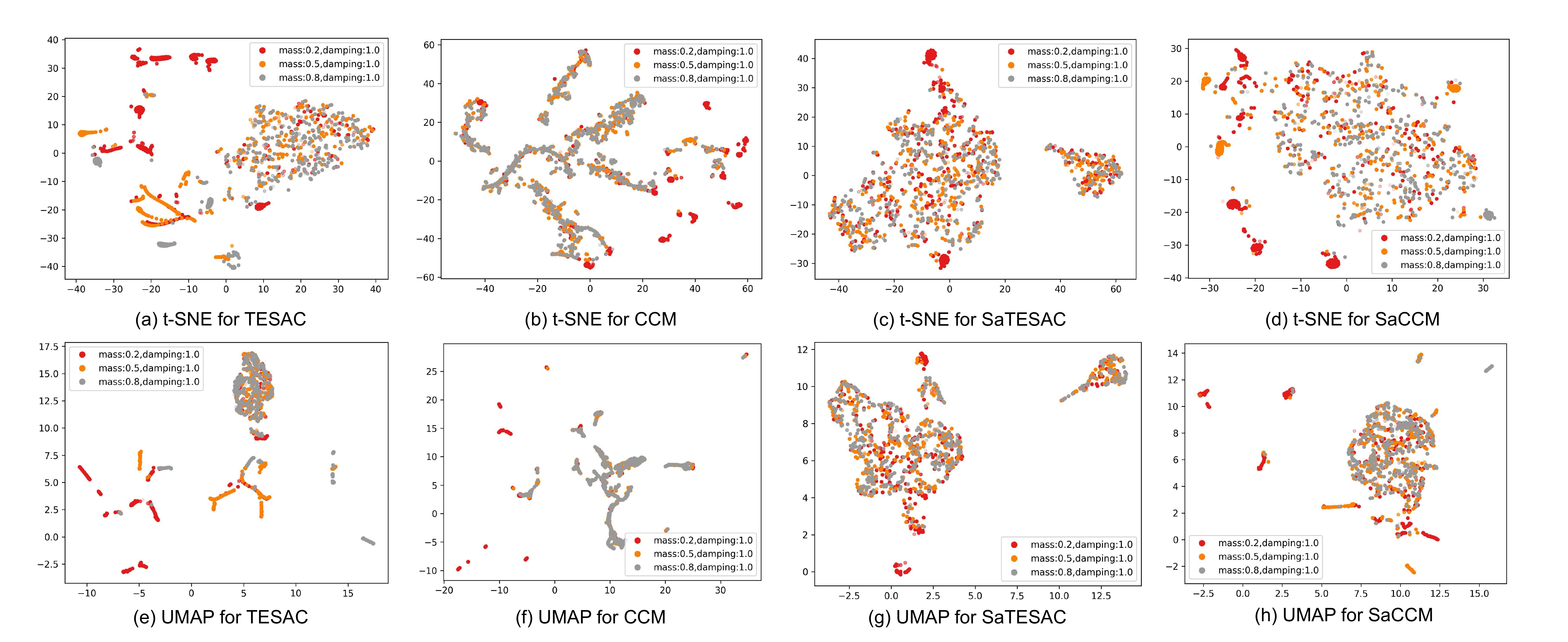}
    \captionof{figure}{t-SNE and UMAP visualisations of context embeddings extracted from trajectories collected in Ant environment. The Ant robot learned a single skill, rolling, to adapt to different mass and damping values.}
\label{fig:context_embedding_Ant}
\end{figure}
\begin{figure}[ht]
    \centering
    \includegraphics[width=1.0\linewidth]{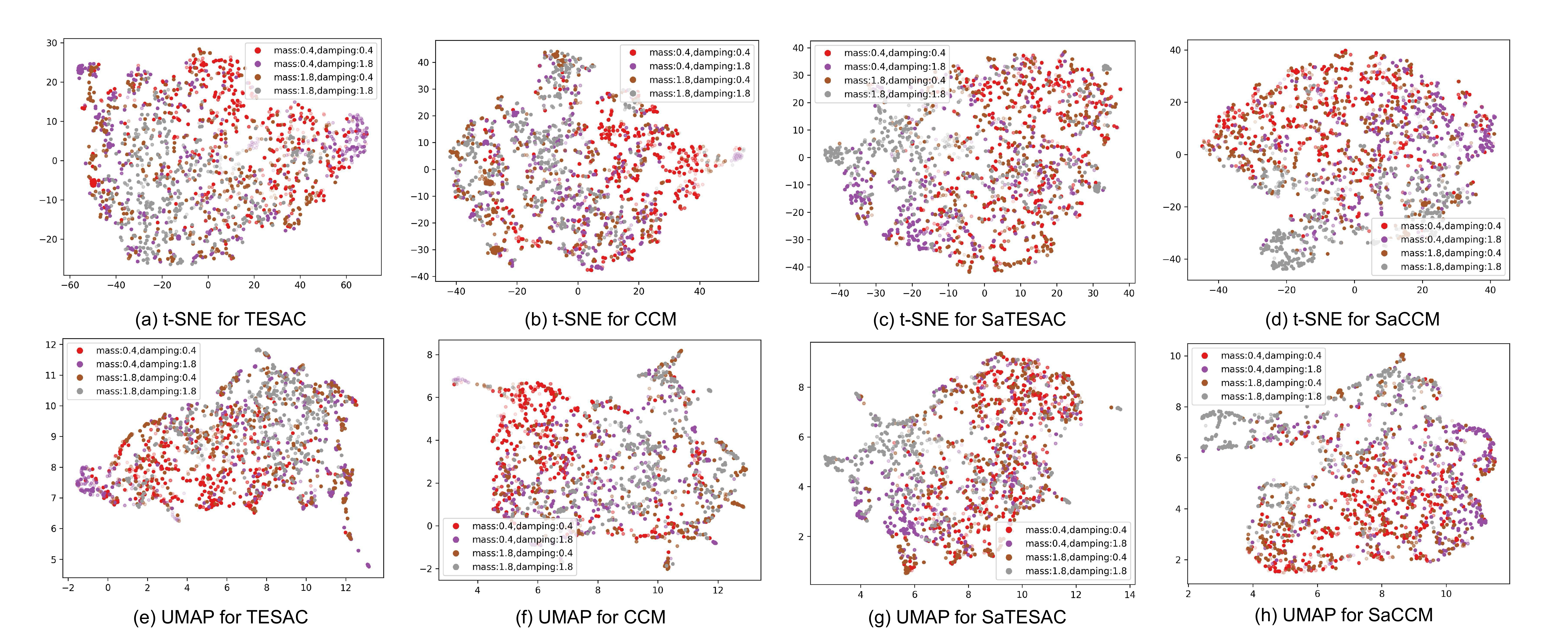}
    \captionof{figure}{t-SNE and UMAP visualisations of context embeddings extracted from trajectories collected in SlimHumanoid environment. The Humanoid Robot crawls on the ground using one elbow. When the damping is relatively high (damping=1.8), the Humanoid Robot can crawl forward stably, but when the damping is low (damping=0.4), it tends to roll.}
\label{fig:context_embedding_SlimHumanoid}
\end{figure}
\begin{figure}[ht]
    \centering
    \includegraphics[width=1.0\linewidth]{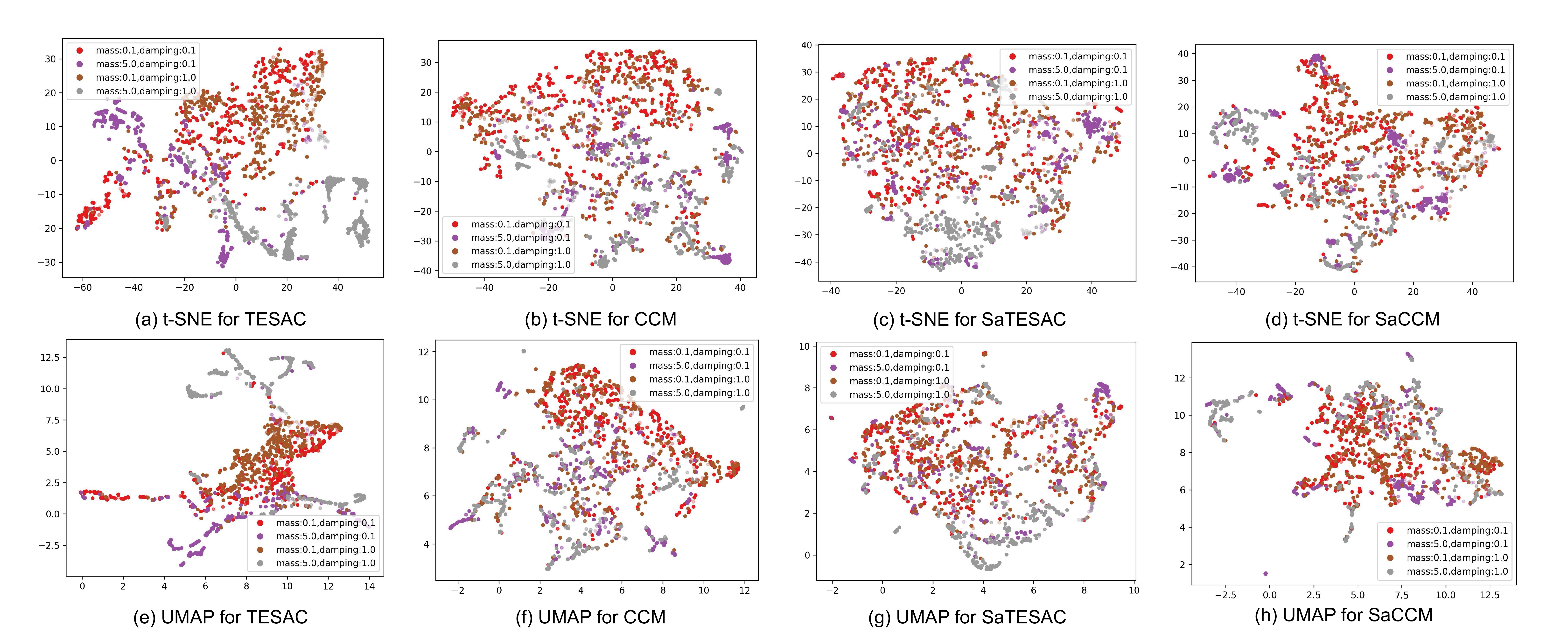}
    \captionof{figure}{t-SNE and UMAP visualisations of context embeddings extracted from trajectories collected in HumanoidStandup environment. SaCCM and SaTESAC learned a sitting posture that makes it easier to stand up, allowing it to generalise well when mass and damping change.}
\label{fig:context_embedding_HumanoidStandup}
\end{figure}
\begin{figure}[ht]
    \centering
    \includegraphics[width=1.0\linewidth]{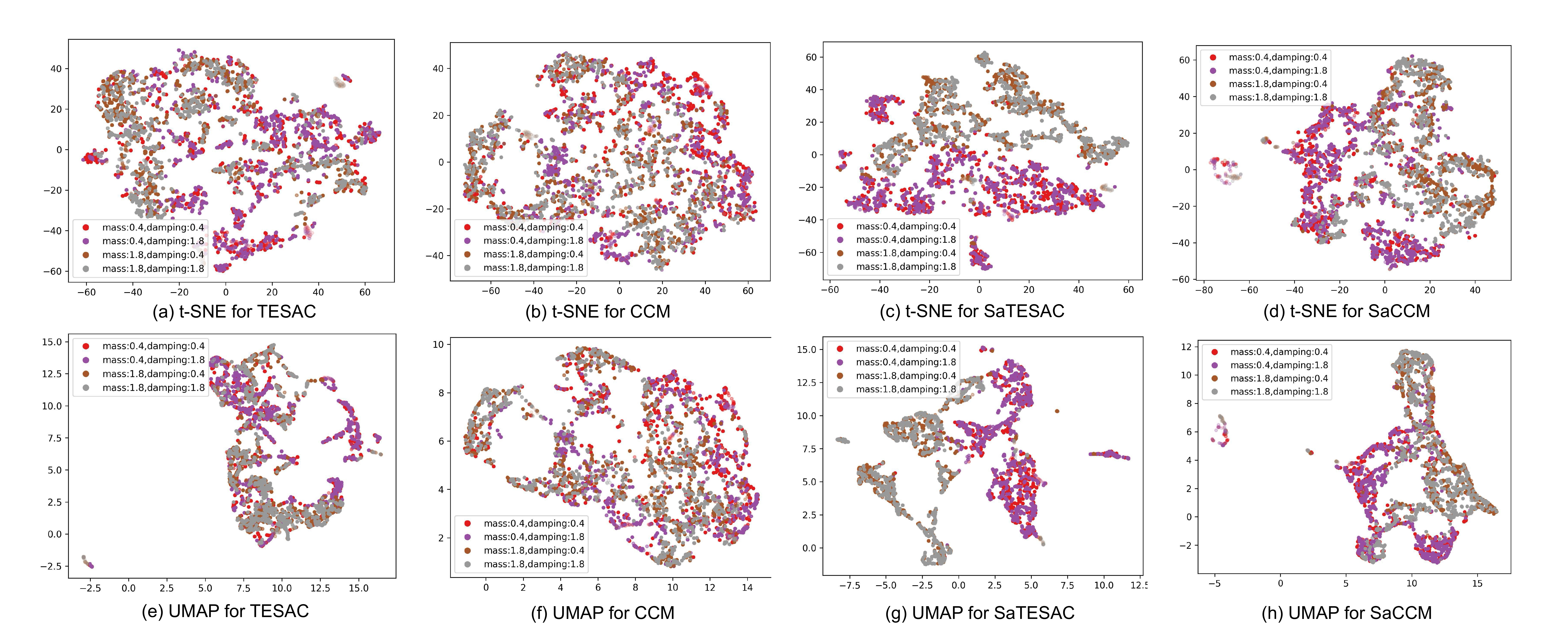}
    \captionof{figure}{t-SNE and UMAP visualisations of context embeddings extracted from trajectories collected in Hopper environment. Combined with the video demos \textsuperscript{\ref{github}} for skill analysis, the plots for SaCCM and SaTESAC show two skills: 1) when the mass is low, the Hopper hops in an upright posture; 2) when the mass is higher, the Hopper hops forward on the floor.}
\label{fig:context_embedding_Hopper}
\end{figure}
\begin{figure}[ht]
    \centering
    \includegraphics[width=1.0\linewidth]{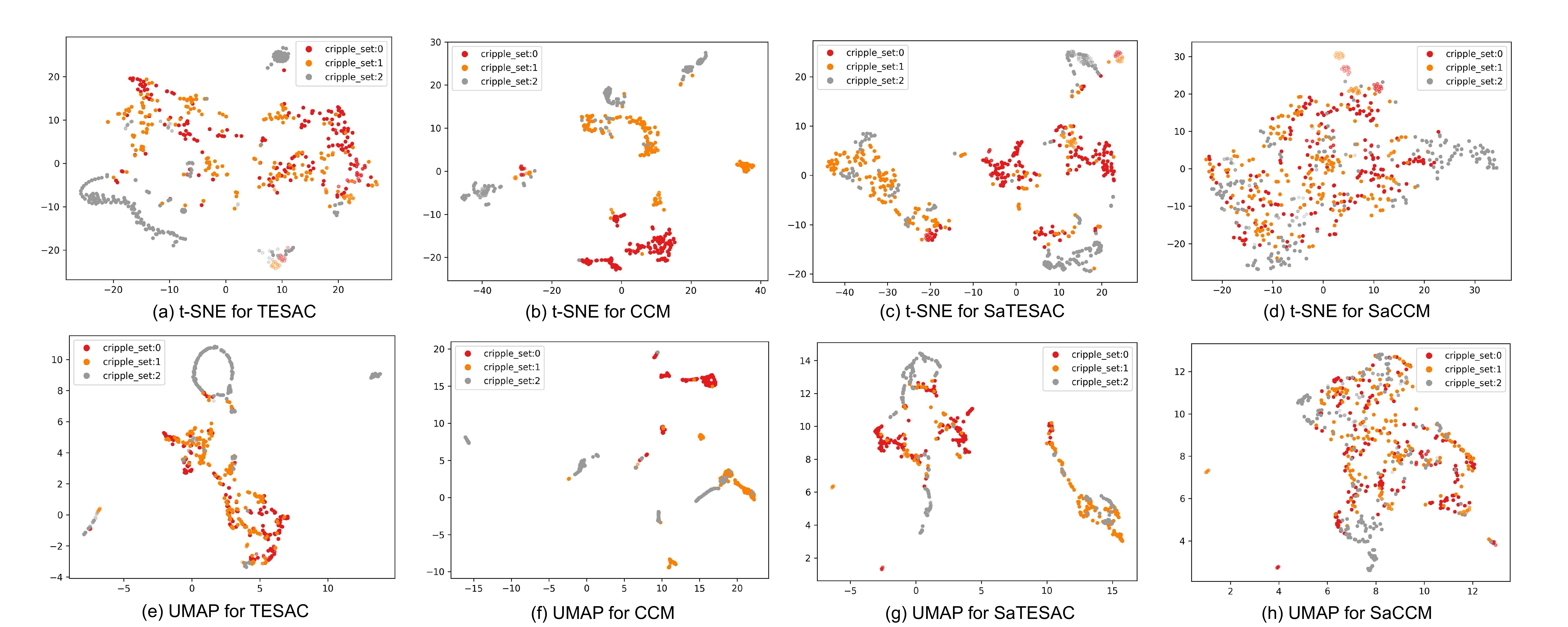}
    \captionof{figure}{t-SNE and UMAP visualisations of context embeddings extracted from trajectories collected in Crippled Hopper environment. "cripple\_set" refers to the index of the crippled joint. SaCCM and SaTESAC learned to take a big hop forward at the beginning (i.e., effective exploration) and then switch to different skills based on environmental feedback.}
\label{fig:context_embedding_CrippledHopper}
\end{figure}

\begin{figure}[ht]
    \centering
    \includegraphics[width=1.0\linewidth]{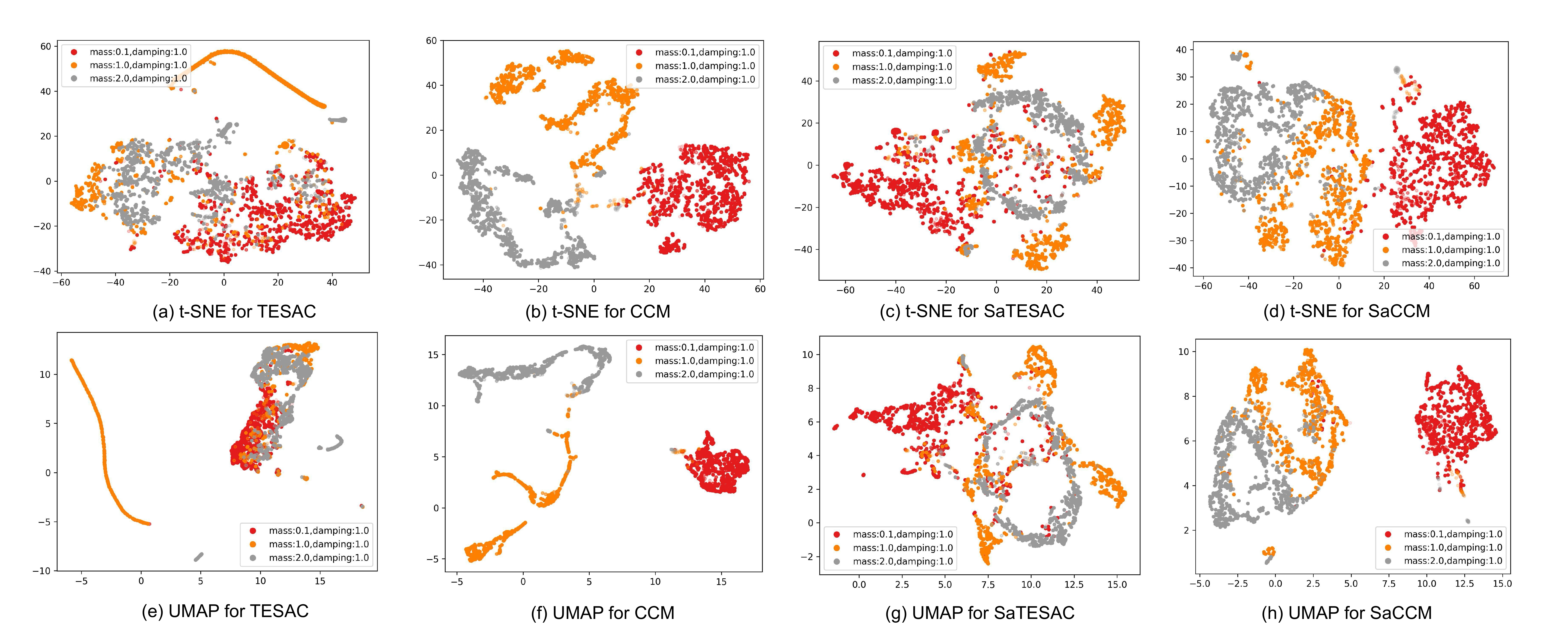}
    \captionof{figure}{t-SNE and UMAP visualisations of context embeddings extracted from trajectories collected in Walker environment. The Walker learned a single skill, hopping forward on the floor through both right and left legs.}
\label{fig:context_embedding_Walker}
\end{figure}
\begin{figure}[ht]
    \centering
    \includegraphics[width=1.0\linewidth]{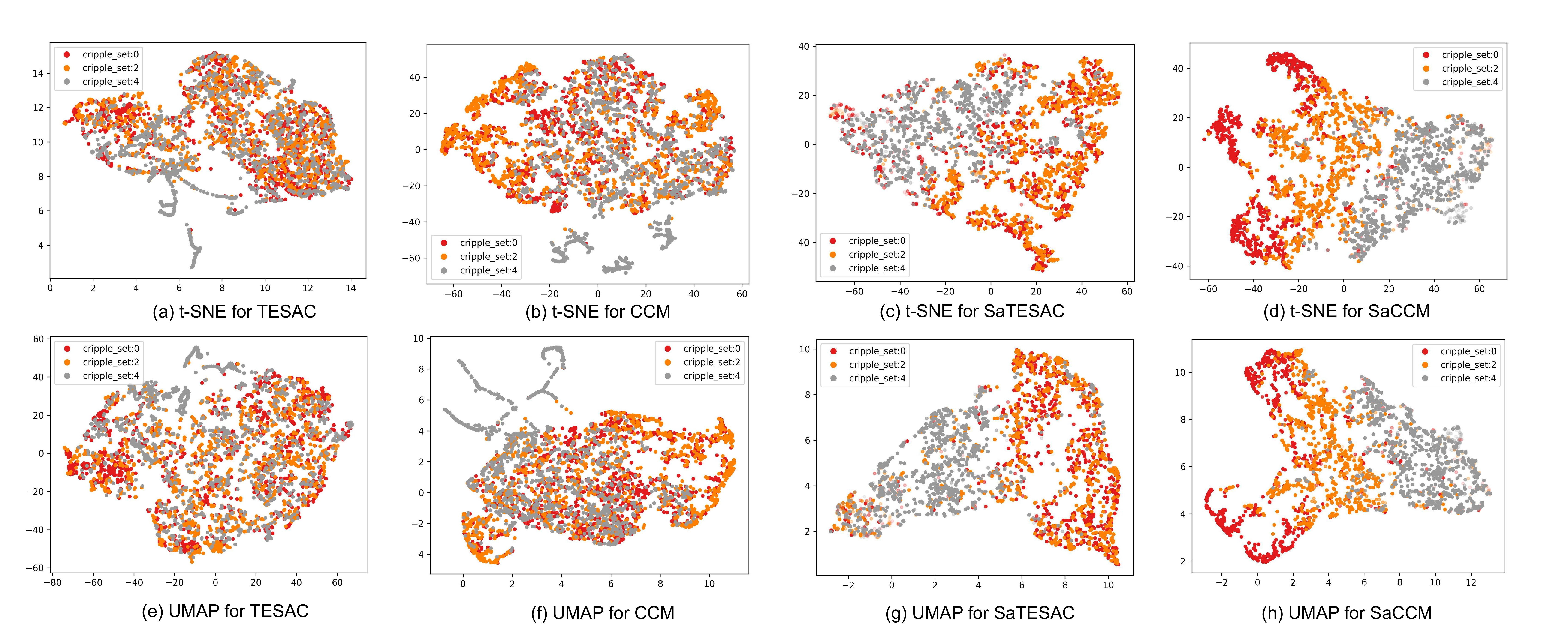}
    \captionof{figure}{t-SNE and UMAP visualisations of context embeddings extracted from trajectories collected in Crippled Walker environment. "cripple\_set" refers to the index of the crippled joint. The Crippled Walker (trained with SaTESAC and SaCCM) learned to hop forward using the right leg (cripple\_set:0 and cripple\_set:2) in the training and moderate test tasks and switched to hopping forward using the left leg (cripple\_set:4) in the extreme test tasks.}
\label{fig:context_embedding_CrippledWalker}
\end{figure}

\begin{table}
	\centering 
	\fontsize{8}{11}\selectfont  
	\caption{Environmental features used for MuJoCo benchmark from DOMINO and CaDM.}
 \scalebox{0.8}{
	\begin{tabular}{ccccc}
		\toprule
		\toprule
		& Training & Test (Moderate) & Test (Extrame) & Episode Length \cr
		\cmidrule(lr){1-5}
		Half-cheetah & \makecell[c]{$m\in \{0.75, 0.85, 1.0, 1.15, 1.25\}$ \\ $d \in \{0.75, 0.85, 1.0, 1.15, 1.25\}$}
                &  \makecell[c]{$m\in \{0.40, 0.50, 1.50, 1.60\}$ \\ $d \in \{0.40, 0.50, 1.50, 1.60\}$} 
                & \makecell[c]{$m\in \{0.20, 0.40, 1.60, 1.80\}$ \\ $d \in \{0.20, 0.40, 1.60, 1.80\}$} & 1000 \cr
            \cmidrule(lr){1-5} 
		Ant & \makecell[c]{$m\in \{0.75, 0.85, 1.0, 1.15, 1.25\}$ \\ $d \in \{0.75, 0.85, 1.0, 1.15, 1.25\}$}
                &  \makecell[c]{$m\in \{0.40, 0.50, 1.50, 1.60\}$ \\ $d \in \{0.40, 0.50, 1.50, 1.60\}$} 
                & \makecell[c]{$m\in \{0.20, 0.40, 1.60, 1.80\}$ \\ $d \in \{0.20, 0.40, 1.60, 1.80\}$} & 1000 \cr
            \cmidrule(lr){1-5}
            
		SlimHumanoid & \makecell[c]{$m\in \{0.80, 0.90, 1.0, 1.15, 1.25\}$ \\ $d \in \{0.80, 0.90, 1.0, 1.15, 1.25\}$}
                &  \makecell[c]{$m\in \{0.60, 0.70, 1.50, 1.60\}$ \\ $d \in \{0.60, 0.70, 1.50, 1.60\}$} 
                & \makecell[c]{$m\in \{0.40, 0.50, 1.70, 1.80\}$ \\ $d \in \{0.40, 0.50, 1.70, 1.80\}$} & 500 \cr
            \cmidrule(lr){1-5}
            
    	Crippled Half-cheetah & \makecell[c]{$m\in \{0.75, 0.85, 1.0, 1.15, 1.25\}$ \\ $d \in \{0.75, 0.85, 1.0, 1.15, 1.25\}$ \\ Crippled Joints: $\{0,1,2,3\}$}
                &  \makecell[c]{$m\in \{0.40, 0.50, 1.50, 1.60\}$ \\ $d \in \{0.40, 0.50, 1.50, 1.60\}$  \\ Crippled Joints: $\{4,5\}$ } 
                & \makecell[c]{$m\in \{0.20, 0.40, 1.60, 1.80\}$ \\ $d \in \{0.20, 0.40, 1.60, 1.80\}$ \\ Crippled Joints: $\{4,5\}$ } & 1000 \cr
		\bottomrule
		\bottomrule
	\end{tabular}}\vspace{0cm}
	\label{tab:Training_sizes_mujoco_2}
\end{table}

\begin{table}
	\centering
	\fontsize{8}{11}\selectfont  
	\caption{Comparison of average return $\pm$ standard deviation with baselines in MuJoCo benchmark (over 5 seeds). The \textbf{bold text} signifies the highest average return. The numerical results for PPO+DOMINO are copied from \cite{mu2022decomposed}; the numerical results for PPO+CaDM, Vanilla+CaDM, and PE-TS+CaDM are copied from \cite{lee2020context}.}
 \scalebox{0.9}{
	\begin{tabular}{ccccccc}
		\toprule
		\toprule  
    	\multirow{2}{*}{ }& \multicolumn{3}{c}{Ant} & \multicolumn{3}{c}{Half-cheetah} \cr
		\cmidrule(lr){2-4} \cmidrule(lr){5-7}
		  & Training & Test (moderate) & Test (extreme) & Training & Test (moderate) & Test (extreme) \cr
		\cmidrule(lr){1-7}
            PPO+DOMINO & 227$\pm$86 & 216$\pm$52 &  & 2472$\pm$803 & 1034$\pm$476 &  \cr
		PPO+CaDM & 268.6$\pm$77.0 & 228.8$\pm$48.4 & 199.2$\pm$52.1 & 2652.0$\pm$1133.6 & 1224.2$\pm$630.0 &   1021.1$\pm$676.6 \cr
		Vanilla+CaDM & 1851.0$\pm$113.7 & 1315.7$\pm$45.5 & 821.4$\pm$113.5 & 3536.5$\pm$641.7 & 1556.1$\pm$260.6 &  1264.5$\pm$228.7  \cr
		PE-TS+CaDM & \textbf{2848.4$\pm$61.9} & \textbf{2121.0$\pm$60.4} &  \textbf{1200.7$\pm$21.8} & \textbf{8264.0$\pm$1374.0} & \textbf{7087.2$\pm$1495.6} & \textbf{4661.8$\pm$783.9}  \cr
            \cmidrule(lr){1-7}
            SaTESAC & 908$\pm$65 & 640$\pm$117 & 532$\pm$88  & 
            7430$\pm$1026 & 4058$\pm$890 &  1780$\pm$102 \cr
		SaCCM & 928$\pm$141 & 635$\pm$94 & 555$\pm$88 & 7154$\pm$965 & 3849$\pm$689  &  1926$\pm$218\cr
  		\bottomrule
		\bottomrule
  
    	\multirow{2}{*}{ }& \multicolumn{3}{c}{SlimHumanoid} & \multicolumn{3}{c}{Crippled Half-Cheetah} \cr
		\cmidrule(lr){2-4} \cmidrule(lr){5-7}
		  & Training & Test (moderate) & Test (extreme) & Training & Test (moderate) & Test (extreme) \cr
		\cmidrule(lr){1-7}
            PPO+DOMINO & 7825$\pm$1256 & 5258$\pm$1039 &  & 2503$\pm$658 & 1326$\pm$491 &  \cr
		PPO+CaDM & \textbf{10455.0$\pm$1004.9} & 4975.7$\pm$1305.7 & 3015.1$\pm$1508.3 & 2356.6$\pm$624.3 & 1454.0$\pm$462.6 & 1025.0$\pm$296.2 \cr
		Vanilla+CaDM & 1758.2$\pm$459.1 & 1228.9$\pm$374.0 &  1487.9$\pm$339.0 & 2435.1$\pm$880.4 & 1375.3$\pm$290.6 &  966.9$\pm$89.4  \cr
		PE-TS+CaDM & 1371.9$\pm$400.0 & 903.7$\pm$343.9 &  814.5$\pm$274.8  & 3294.9$\pm$733.9 & 2618.7$\pm$647.1 & 1294.2$\pm$214.9  \cr
            \cmidrule(lr){1-7}
            SaTESAC & 10216$\pm$1620 & \textbf{7886$\pm$2203} &  6123$\pm$1403 &  5169$\pm$730 & 2184$\pm$592 & 1628$\pm$281\cr
		SaCCM & 9312$\pm$705 & 7430$\pm$1587 & \textbf{6473$\pm$2001}  & \textbf{5709$\pm$744} & \textbf{2795$\pm$446} &  \textbf{2115$\pm$466}\cr
  		\bottomrule
		\bottomrule
	\end{tabular}}\vspace{0cm}
	\label{tab:sub-standard-average-return}
\end{table}

\clearpage

\section{A comparison with DOMINO and CaDM} \label{appx:sub-standard-comparison}

In this section, we give a brief comparison between our methods and methods from DOMINO \citep{mu2022decomposed} and CaDM \citep{lee2020context} in the MuJoCo benchmark because we are using the exact same environmental setting (shown in Table \ref{tab:Training_sizes_mujoco_2}).

DOMINO \citep{mu2022decomposed} is based on the InfoNCE $K$-sample estimator. Their implementation, PPO+DOMINO, is a model-free RL algorithm with a pre-trained context encoder. This encoder reduces the demand for large contrastive batch sizes during training by decoupling representation learning for each modality, simplifying tasks while leveraging shared information. However, a pre-trained encoder necessitates a large sample volume, with DOMINO training PPO agents for 5 million timesteps on the MuJoCo benchmark. In contrast, SaTESAC and SaCCM, trained for 1.6 million timesteps without pre-trained encoders, achieve considerably higher average returns across four environments (Table \ref{tab:sub-standard-average-return}). Therefore, it is crucial to focus on extracting MI in contrastive learning that directly optimises downstream tasks, integrating rather than segregating representation learning from task performance.

CaDM \citep{lee2020context} proposes a context-aware dynamics model adaptable to changes in dynamics. Specifically, they utilise contrastive learning to learn context embeddings, and then predict the next state conditioned on them. We copy the numerical results of PPO+CaDM, Vanilla+CaDM, and PE-TS+CaDM from CaDM \citep{lee2020context} as their environmental setting is identical to ours, where PPO+CaDM is a model-free RL algorithm, while Vanilla+CaDM and PE-TS+CaDM are model-based. The model-free RL approach, PPO+CaDM, is trained for 5 million timesteps on the MuJoCo benchmark. As shown in Table \ref{tab:sub-standard-average-return}, SaTESAC and SaCCM significantly outperform PPO+CaDM. The model-based RL algorithms, Vanilla+CaDM and PE-TS+CaDM, require 2 million timesteps for learning in model-based setups, compared to our fewer samples (i.e., million timesteps). In the Ant environment, Vanilla+CaDM and PE-TS+CaDM achieve higher returns than SaTESAC and SaCCM; similarly, in the Half-cheetah environment, PE-TS+CaDM outperforms them. Results in the SlimHumanoid and Crippled Half-cheetah environments show that skill-aware context embeddings are notably effective. An insight here is that our method outperforms the model-free CaCM approach, but not the model-based one. This is consistent with what is empirically found in CaDM \citep{lee2020context}: prediction models are more
effective when the transition function changes across tasks. Therefore, we consider that a model-based approach to SaMI could be an interesting extension for future work.

\clearpage

\section{Statistical hypothesis tests (paried t-tests)} \label{appx:statistical_hypothesis_tests}
We used a t-test \citep{rice2007mathematical} to conduct a statistical hypothesis test to determine whether SaMI brought a statistically significant improvement. we reported the p-value of the t-test in MuJoCo (Table \ref{tab:t_test}) and Panda-gym (Table \ref{tab:t_test_panda}) benchmarks. $*$ next to the number is used to indicate that the algorithm with SaMI has statistically significant improvement over the same algorithm without SaMI at a significance level of 0.05. From Table \ref{tab:t_test} and \ref{tab:t_test_panda}, SaMI brings significant improvement on the extreme test set in which the RL agent needs to execute diverse skills. The statistically significant test aligns with our results in the skill analysis (i.e., video demos, t-SNE and UMAP visualisation). In complex environments that require high skill diversity from the RL agent, the statistically significant improvement and higher returns/success rates brought by SaMI are evident.

\begin{table}
\centering
\fontsize{8}{10}\selectfont  
\caption{The p-value of the statistical hypothesis tests (paried t-tests) for comparing the effectiveness of SaMI in MuJoCo benchmark (over 5 seeds). $*$ next to the number means that the algorithm with SaMI has statistically significant improvement over the same algorithm without SaMI at a significance level of 0.05. The “SaTESAC-TESAC” row indicates the p-value for the return improvement brought by SaMI to TESAC; the “SaCCM-CCM” row indicates the p-value for the return improvement brought by SaMI to CCM. }
	\begin{tabular}{ccccccc}
		\toprule
		\toprule
		\multirow{2}{*}{ }& \multicolumn{3}{c}{Crippled Ant} & \multicolumn{3}{c}{Crippled Half-cheetah}\cr
		\cmidrule(lr){2-4} \cmidrule(lr){5-7}
		  & Training & Test (moderate) & Test (extreme) & Training & Test (moderate) & Test (extreme)\cr
		\cmidrule(lr){1-7}
            SaTESAC-TESAC & 0.121 & 0.109 & 9.54E-07$^*$  & 0.154 & 0.0024$^*$ & 0.0889  \cr
		  SaCCM-CCM & 0.913 & 0.108 & 0.008$^*$ & 0.04$^*$ & 0.307 & 0.106 \cr
		\bottomrule
		\bottomrule
  
    	\multirow{2}{*}{ }& \multicolumn{3}{c}{Ant} & \multicolumn{3}{c}{Half-cheetah}  \cr
		\cmidrule(lr){2-4} \cmidrule(lr){5-7} 
		  & Training & Test (moderate) & Test (extreme) & Training & Test (moderate) & Test (extreme) \cr
		\cmidrule(lr){1-7}
            SaTESAC-TESAC & 0.136 & 0.034$^*$ & 0.021$^*$  & 
            0.346 & 0.224 &  0.004$^*$ \cr
		SaCCM-CCM & 0.138 & 0.275 & 0.163 & 0.73 & 0.791  &  0.005$^*$ \cr
  		\bottomrule
		\bottomrule

  \multirow{2}{*}{ }&  \multicolumn{3}{c}{SlimHumanoid} & \multicolumn{3}{c}{HumanoidStandup}  \cr
		\cmidrule(lr){2-4} \cmidrule(lr){5-7}
		  & Training & Test (moderate) & Test (extreme) & Training & Test (moderate) & Test (extreme)\cr
		\cmidrule(lr){1-7}
        SaTESAC-TESAC & 0.139 & 0.456 &  0.006$^*$ &  0.037$^*$ & 0.129 &  0.003$^*$ \cr
		SaCCM-CCM & 0.113 & 0.059 & 0.008$^*$  & 0.048 & 0.027$^*$ &  0.01 \cr
		\bottomrule
		\bottomrule
  
    \multirow{2}{*}{ }&  \multicolumn{3}{c}{Hopper} & \multicolumn{3}{c}{Crippled Hopper}  \cr
		\cmidrule(lr){2-4} \cmidrule(lr){5-7}
		  & Training & Test (moderate) & Test (extreme) & Training & Test (moderate) & Test (extreme)\cr
		\cmidrule(lr){1-7}
            SaTESAC-TESAC & 0.747 & 0.089 &  0.707  & 0.459 & 0.69 & 0.088  \cr
		SaCCM-CCM & 0.52 & 0.599  &  0.969 & 0.967 & 0.897 & 0.0002$^*$  \cr
		\bottomrule
		\bottomrule
  
    \multirow{2}{*}{ }&  \multicolumn{3}{c}{Walker} & \multicolumn{3}{c}{Crippled Walker}  \cr
		\cmidrule(lr){2-4} \cmidrule(lr){5-7}
		  & Training & Test (moderate) & Test (extreme) & Training & Test (moderate) & Test (extreme)\cr
		\cmidrule(lr){1-7}
            SaTESAC-TESAC & 0.312 & 0.082 & 0.003$^*$ & 0.223 & 0.048$^*$ &  0.028$^*$ \cr
		SaCCM-CCM & 0.55 & 0.541 & 0.022$^*$  & 0.794 & 0.079  & 0.011$^*$ \cr
		\bottomrule
		\bottomrule
	\end{tabular}
	\label{tab:t_test}
\end{table}

\begin{table}
\centering
\fontsize{8}{10}\selectfont  
\caption{The p-value of the statistical hypothesis tests (paried t-tests) for comparing the effectiveness of SaMI in Panda-gym benchmark (over 5 seeds). $*$ next to the number means that the algorithm with SaMI has statistically significant improvement over the same algorithm without SaMI at a significance level of 0.05. The “SaTESAC-TESAC” row indicates the p-value for the return improvement brought by SaMI to TESAC; the “SaCCM-CCM” row indicates the p-value for the return improvement brought by SaMI to CCM.}
    \centering
    \begin{tabular}{cccc}
		\toprule
		\toprule
		  & Training & Test (moderate) & Test (extreme)  \cr
		\cmidrule(lr){1-4}
            SaTESAC-TESAC & 0.000260$^*$ & 0.000160$^*$ & 0.001310$^*$  \cr
		SaCCM-CCM & 0.000230$^*$ & 0.002190$^*$ &  0.001390$^*$  \cr
		\bottomrule
		\bottomrule
	\end{tabular}\vspace{0cm}
\label{tab:t_test_panda}
\end{table}

\end{document}